%% file: main.tex
\documentclass[10pt,twocolumn,letterpaper]{article}

\usepackage[accsupp]{axessibility}

\usepackage{iccv}
\makeatletter
\@namedef{ver@everyshi.sty}{}
\makeatother
\usepackage{tikz}

\usepackage{times}
\usepackage{epsfig}
\usepackage{graphicx}
\usepackage{amsmath}
\usepackage{amssymb}
\usepackage{xcolor}
\usepackage{pifont}

\usepackage[normalem]{ulem}
\usepackage{verbatim}
\usepackage{multirow}
\usepackage{cuted}
\usepackage{array, caption, floatrow, tabularx, makecell, booktabs}
\usepackage{subcaption}
\usepackage{tabu}
\usepackage{tikz}
\usepackage{array}
\usepackage{adjustbox}
\usepackage[font=footnotesize,labelfont=footnotesize]{subcaption}
\usepackage[font=small,labelfont=footnotesize]{caption}

\usepackage[pagebackref=true,breaklinks=true,colorlinks,bookmarks=false]{hyperref}

\definecolor{DarkGreen}{RGB}{21,100,52}
\definecolor{DarkRed}{RGB}{139,0,0}
\definecolor{DarkRed2}{RGB}{180,0,0}
\definecolor{MyGreen}{rgb}{0, 0.55, 0}

\newcommand{\xmark}{{\color{DarkRed2}\ding{55}}}

\newcommand{\ourmethod}[1]{Total-Recon}
\newcolumntype{M}[1]{>{\centering\arraybackslash}m{\#1}}

\newcommand{\captionspace}{\vspace{-6pt}}

\iccvfinalcopy

\begin{document}

\title{\ourmethod{}: Deformable Scene Reconstruction for Embodied View Synthesis}

\author{Chonghyuk Song \; Gengshan Yang \; Kangle Deng \; Jun-Yan Zhu \; Deva Ramanan\\
[2pt]
Carnegie Mellon University\\
}

\maketitle

\begin{strip}\centering
\vspace{-55pt}
\captionsetup{type=figure}
\includegraphics[width=\linewidth, trim={9mm 0cm 0 0cm},clip]{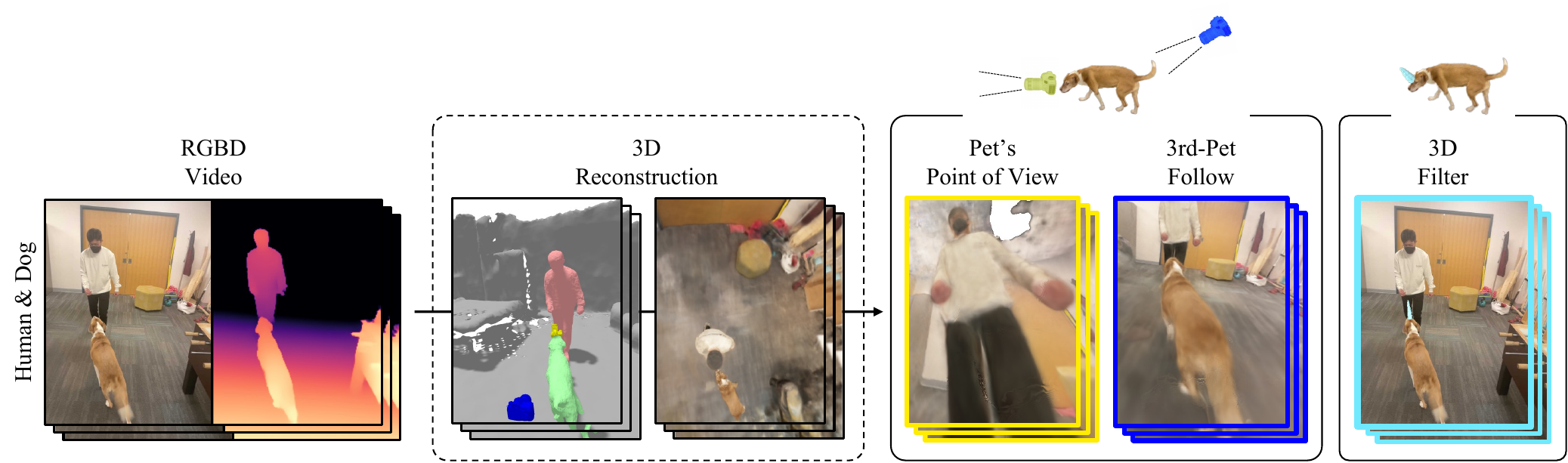}
\captionof{figure}{\textbf{Embodied View Synthesis.} Given a long video of deformable objects captured by a handheld RGBD sensor, \ourmethod{} renders the scene from novel camera trajectories derived from the {\em in}-scene motion of actors: (1) egocentric cameras that simulate the {\em point-of-view of a target actor} (such as the pet) and (2) 3rd-person (or pet) cameras that follow the actor from behind. Our method also enables (3) 3D video filters that attach virtual 3D assets to the actor. \ourmethod{} achieves this by reconstructing the geometry, appearance, root-body- and articulated motion of each deformable object in the scene and the background. \href{https://andrewsonga.github.io/totalrecon}{\textbf{[Videos]}}}
\label{fig:teaser}
\end{strip}

\begin{abstract}
\vspace{-10pt}
We explore the task of embodied view synthesis from monocular videos of deformable scenes. Given a minute-long RGBD video of people interacting with their pets, we render the scene from novel camera trajectories derived from the in-scene motion of actors: (1) egocentric cameras that simulate the point of view of a target actor and (2) 3rd-person cameras that follow the actor. Building such a system requires reconstructing the root-body and articulated motion of every actor, as well as a scene representation that supports free-viewpoint synthesis. Longer videos are more likely to capture the scene from diverse viewpoints (which helps reconstruction) but are also more likely to contain larger motions (which complicates reconstruction). To address these challenges, we present \ourmethod{}, the first method to photorealistically reconstruct deformable scenes from long monocular RGBD videos. Crucially, to scale to long videos, our method hierarchically decomposes the scene into the background and objects, whose motion is decomposed into carefully initialized root-body motion and local articulations. To quantify such ``in-the-wild" reconstruction and view synthesis, we collect ground-truth data from a specialized stereo RGBD capture rig for 11 challenging videos, significantly outperforming prior methods. Our code, models, and data can be found \href{https://andrewsonga.github.io/totalrecon}{here}.
\end{abstract}
\vspace{-10pt}
\input{1_intro}

\input{2_related}

\input{3_method}
\input{4_experiments}

\input{5_conclusion}

{\small
\bibliographystyle{ieee_fullname}
\bibliography{main}
}

\input{Appendix}

\end{document}

%% file: 1_intro.tex
\section{Introduction}
We explore \textit{embodied view synthesis}, a new class of novel-view synthesis tasks that renders deformable scenes from novel 6-DOF trajectories reconstructed from the {\em in}-scene motion of actors: egocentric cameras \cite{perspective, fps_vs_tps} that simulate the point-of-view of moving actors and 3rd-person-follow cameras \cite{3rd_person_follow, fps_vs_tps} that track a moving actor from behind (Figure \ref{fig:teaser}). We focus on everyday scenes of people interacting with their pets, producing renderings from the point-of-view of the person {\em and pet} (Figure~\ref{fig:teaser}). While such camera trajectories could be manually constructed (e.g., by artists via keyframing), building an \textit{automated} system is an interesting problem of its own: spatial cognition theory~\cite{spatial_cognition} suggests that the ability to visualize behavior from another actor's perspective is necessary for action learning and imitation; in the context of gaming and virtual reality~\cite{fps_vs_tps, perspective}, egocentric cameras offer high levels of user immersion, while 3rd-person-follow cameras provide a large field of view that is useful for exploring a user's environment.

\begin{figure*}[t]
\includegraphics[width=.80\linewidth,keepaspectratio=true]{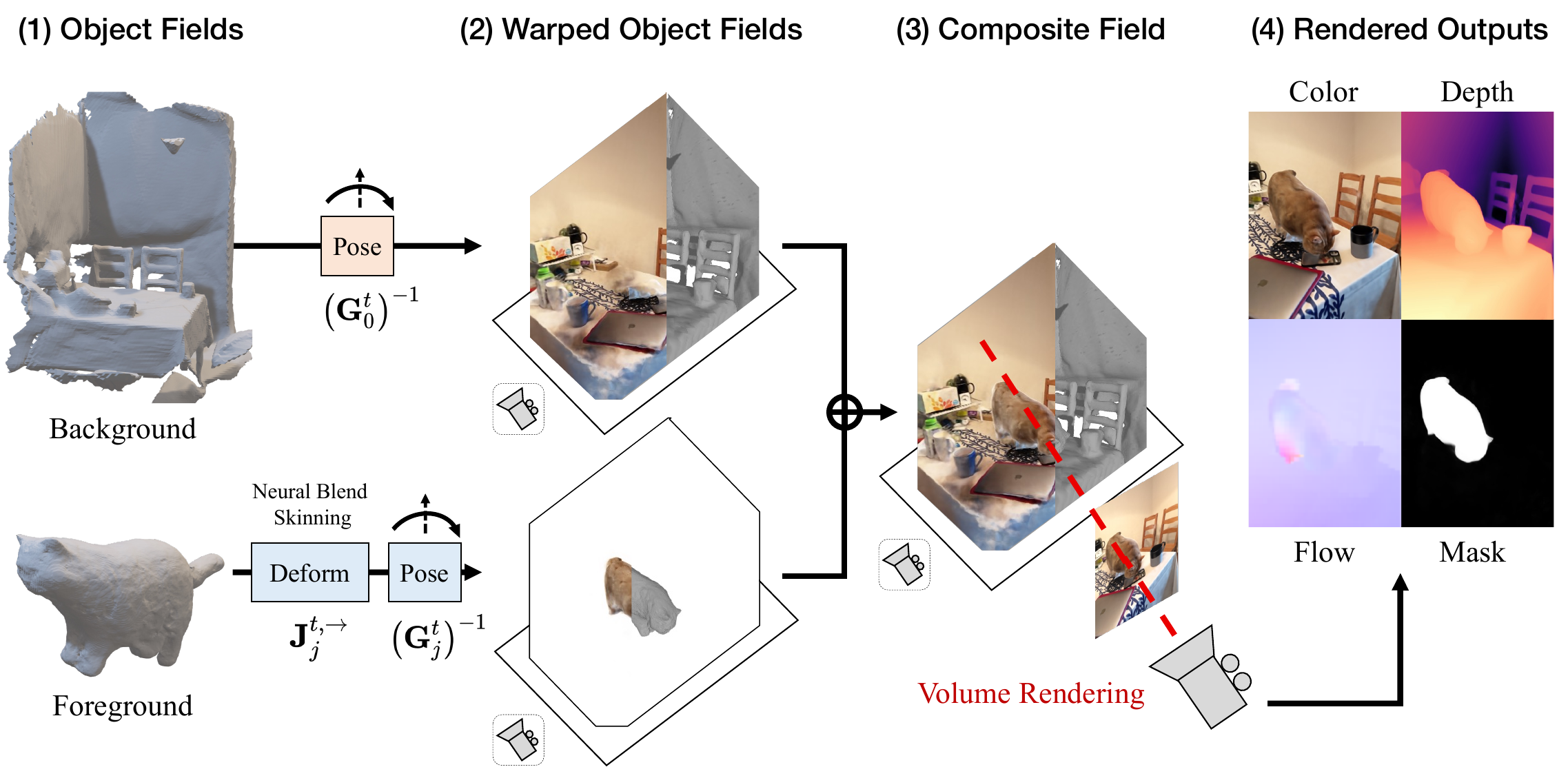}
\caption{\textbf{Method Overview.} \ourmethod{} represents the entire scene as a composition of $M$ object-centric neural fields, one for the rigid background and each of the $M-1$ deformable objects. To render a scene, (1) \textit{each object field} $j$ is transformed into the camera space with a rigid transformation $\left(\mathbf{G}_{j}^{t}\right)^{-1}$ that encodes root-body motion and, for each deformable object, an additional deformation field $\mathbf{J}^{t, \rightarrow}_{j}$ that encodes articulated motion. Next, all (2) \textit{warped object fields} are combined into a (3) \textit{composite field}, which is then volume-rendered into (4) \textit{color, depth, optical flow, and object silhouettes}. Each rendered output defines a reconstruction loss that derives supervision from a monocular RGBD video captured by a moving iPad Pro.
\captionspace
\vspace{-0.3em}
}
\label{fig:method}
\end{figure*}

\paragraph{Challenges.} \looseness=-2
Building a system for embodied view synthesis is challenging for many reasons. First, to reconstruct everyday-but-interesting content, it needs to process long, monocular captures of multiple interacting actors. However, such videos are likely to contain large scene motions, which we demonstrate are difficult to reconstruct with current approaches. Second, it needs to produce a deformable 3D scene representation that supports free-viewpoint synthesis, which also would benefit from long videos likely to capture the scene from diverse viewpoints. Recent approaches have extended Neural Radiance Fields (NeRFs) \cite{mildenhall2020nerf} to deformable scenes, but such work is often limited to rigid-only object motion~\cite{Kundu_2022_CVPR,Ost_2021_CVPR}, short videos with limited scene motion ~\cite{pumarola2020d, park2021nerfies, li2021neural, tretschk2021nonrigid, park2021hypernerf, d2nerf, dynnerf, videonerf, wang2021neural}, or reconstructing single objects as opposed the entire scene~\cite{yang2021lasr, yang2021viser, yang2022banmo, cai2022neural}. Third, it needs to compute global 6-DOF trajectories of root-bodies and articulated body parts (e.g., head) of multiple actors.

\vspace{-1em}

\paragraph{Key Ideas.} \looseness=-2
To address these challenges, we introduce \ourmethod{}, the first monocular NeRF that enables embodied view synthesis for deformable scenes with large motions. Given a monocular RGBD video, \ourmethod{} reconstructs the scene as a composition of object-centric representations, which encode the 3D appearance, geometry, and motion of each deformable object and the background. Crucially, \ourmethod{} hierarchically decomposes scene motion into the motion of individual objects, which itself is decomposed into global root-body movement and the local deformation of articulated body parts. We demonstrate that such decomposition of object motion, along with appropriate initialization of root-body pose, allows reconstruction to scale to longer videos, enabling free-viewpoint synthesis. By reconstructing such motions in a globally consistent coordinate frame, \ourmethod{} can generate renderings from egocentric and 3rd-person-follow cameras, as well as static but extreme viewpoints like bird's-eye-views.

\vspace{-1em}

\paragraph{Evaluation.} \looseness=-2
Due to the difficulty of collecting ground-truth data for embodied view synthesis on in-the-wild videos, we evaluate our method on the proxy task of stereo-view synthesis \cite{park2021nerfies}, which compares rendered views to those captured from a stereo pair. To this end, we build a stereo RGBD sensor capture rig for ground-truthing and collect a dataset of 11 long video sequences in various indoor environments, including people interacting with their pets. \ourmethod{} outperforms the state-of-the-art monocular deformable NeRF methods \cite{park2021hypernerf, d2nerf}, even when modified to use depth sensor measurements.

\paragraph{Contributions.} %
In summary, our contributions are threefold: (1) \ourmethod{}, a hierarchical 3D representation that models deformable scenes as a composition of object-centric representations, each of which decomposes object motion into its global root-body motion and its local articulations; (2) a system based on \ourmethod{} for automated embodied view synthesis from casual, minute-long RGBD videos of deformable scenes; (3) a dataset of stereo RGBD videos containing a variety of highly dynamic and deformable objects, such as humans and pets, in a host of different background environments.

%% file: 2_related.tex
\section{Related Work} 

\begin{table}[t]
\caption{\textbf{Comparison to Related Work.} Unlike prior work, \ourmethod{} targets embodied view synthesis of scenes containing humans and pets, requiring the ability to (1) reconstruct \textit{entire scenes}, (2) model \textit{deformable objects}, (3) extend \textit{beyond humans}, (4) recover \textit{global 6-DOF trajectories} of objects' root-bodies and their articulated parts, (5) process \textit{minute-long videos} of dynamic scenes, and (6) render \textit{extreme views}.\vspace{-0.3em}} %
\label{table:related_works}
\centering
\setlength{\tabcolsep}{2pt}
\input{tables/related_work_capabilities}

\end{table}

\paragraph{Neural Radiance Fields.} Prior works on Neural Radiance Fields (NeRF) optimize a continuous scene function for novel view synthesis given a set of multi-view images, usually under the assumption of a rigid scene and densely sampled views~\cite{mildenhall2020nerf, martinbrualla2020nerfw,meng2021gnerf, lin2021barf, jeong2021self,wang2021nerf}. 
DS-NeRF~\cite{Deng_2022_CVPR} and Dense Depth Priors~\cite{roessle2022depthpriorsnerf} extend NeRFs to the sparse-view setting by introducing depth as additional supervision. \ourmethod{} also operates in the sparse-view regime and uses depth supervision to reduce the ambiguities inherent to monocular, multibody, non-rigid reconstruction \cite{ non_rigid_ambiguity, relative_scale_ambiguity}. Another line of work~\cite{Kundu_2022_CVPR, Ost_2021_CVPR} represents rigidly moving scenes as a composition of multiple object-level NeRFs. \ourmethod{} also leverages such an object-centric scene representation but models scenes containing \textit{non-rigidly} moving objects, such as humans and pets.
\vspace{-0.4em}

\paragraph{Deformable NeRFs.}
Recent approaches extend NeRF to monocular deformable scene reconstruction either by learning an additional function that deforms observed points in the camera space to a time-independent canonical space~\cite{pumarola2020d, park2021nerfies, tretschk2021nonrigid, park2021hypernerf, d2nerf} or explicitly modeling density changes over time~\cite{dynnerf, videonerf, wang2021neural, li2021neural}. Such methods are typically limited to short videos containing little scene and camera motion. They also perform novel-view synthesis only over small baselines. \ourmethod{} belongs to the former category of prior monocular deformable NeRFs, but unlike them, our method hierarchically decomposes scene motion into the motion of each object, which is further decomposed into global root-body motion and local articulations. The proposed motion decomposition is what enables embodied view synthesis: it allows \ourmethod{} to scale to \textit{minute-long} videos and reconstruct a deformable 3D scene representation that supports free-viewpoint synthesis; it also makes it easy to extract an object's root-body motion, the key motion primitive required for 3rd-person-follow view synthesis.
Several works have taken different approaches to making non-rigid reconstruction more tractable. One group of work leverages human-specific priors~\cite{peng2021animatable, su2021anerf, 2021narf, peng2021neural, liu2021neural, neural-human-radiance-field, li2022tava, pavlakos2022sitcoms3D} such as human body models (e.g., SMPL), 3D skeletons, or 2D poses to achieve high reconstruction quality. We achieve similar levels of fidelity \textit{without} relying on such shape priors, allowing \ourmethod{} to generalize to pets and, by extension, reconstruct human-pet interaction videos.  Another body of work~\cite{zhang2021stnerf, shuai2022multinb, Li_2022_CVPR} achieves high-fidelity scene reconstructions by relying on synchronized multi-view video captured from a specialized camera rig ranging from 8 to 18 static cameras. In contrast, \ourmethod{} only requires a single video captured from a moving RGBD camera equipped with inertial sensors, which has now become widely accessible in consumer products with the advent of Apple's iPhone and iPad Pro.

\vspace{-0.4em}
\paragraph{Reconstruction with RGBD Sensors.}
Depth sensors represent the third class of attempts to make non-rigid reconstruction more tractable, reducing the need for a pre-defined shape template.
Kinect-fusion~\cite{newcombe2011kinectfusion} creates a real-time system for indoor scene localization and mapping. Dynamic Fusion~\cite{newcombe2015dynamicfusion} builds a template-free dense SLAM system for dynamic objects. Later works improve RGBD reconstruction to be able to deal with topology changes~\cite{slavcheva2017killingfusion, slavcheva2018sobolevfusion} and use correspondence matching for registration over large motions~\cite{bozic2021neural,bozic2020deepdeform}. Recent works have incorporated neural implicit representations to reconstruct the surface geometry and 3D motion fields for deformable objects~\cite{ren2021class, cai2022neural} or large-scale rigid scenes~\cite{Azinovic_2022_CVPR, Rematas_2022_CVPR} in isolation. Other works have reconstructed humans alongside small-scale objects and furniture~\cite{dou20153d, bozic2021neural} but not the entire background. We aim to go even further by reconstructing the entire scene, which includes the background and multiple deformable targets such as humans and pets; not only do we reconstruct the geometry, but we also recover a radiance field that allows for photorealistic scene rendering from embodied viewpoints and other novel 6-DOF trajectories.

\vspace{-0.4em}
\paragraph{Concurrent Work.} Concurrent work exhibits a subset of the design choices necessary for embodied view synthesis. SLAHMR \cite{ye2023slahmr} reconstructs the geometry and in-scene motion of human actors but not the scene appearance. Nerflets \cite{nerflets} models the appearance, geometry, and motion of each scene element but is limited to rigidly moving objects. RoDynRF \cite{liu2023robust}, NeRF-DS \cite{nerf-ds}, HexPlane \cite{Cao2022FWD}, and K-planes \cite{kplanes_2023} reconstruct other types of dynamic scene elements, such as deformable or specular objects, but these methods have been demonstrated on only short videos and/or videos containing limited object root-body motion \cite{davis, nvidia_nvs_dataset, nerf-ds, gao2022dynamic}. DynIBaR \cite{li2023dynibar} scales dynamic view synthesis to longer videos with complex camera and scene motion, and SUDS \cite{turki2023suds} scales reconstruction to urban-scale dynamic scenes captured from 1.2 million frames. However, neither demonstrates extreme-view synthesis, a prerequisite for rendering embodied views. We summarize and compare prior work to \ourmethod{} in Table \ref{table:related_works}.

%% file: tables/related_work_capabilities.tex
\centering
\footnotesize{
\setlength{\tabcolsep}{1pt}
\begin{tabularx}{\textwidth}{l||X|X|X|X|X|X}
    \hline
    \makecell{Method} 
    & \makecell{Entire\\Scenes}
    & \makecell{Deform.\\Objects}
    & \makecell{Beyond\\Humans}
    & \makecell{Global\\6-DOF\\Traj.}
    & \makecell{Long\\Videos}
    & \makecell{Extreme\\Views} \\
    \hline
    BANMo \cite{yang2022banmo}
    & \makecell{\xmark} 
    & \makecell{\textcolor{MyGreen}{\checkmark}} 
    & \makecell{\textcolor{MyGreen}{\checkmark}}
    & \makecell{\xmark} 
    & \makecell{\textcolor{MyGreen}{\checkmark}}
    & \makecell{\textcolor{MyGreen}{\checkmark}}
    \\
    PNF \cite{Kundu_2022_CVPR}
    & \makecell{\textcolor{MyGreen}{\checkmark}} 
    & \makecell{\xmark}
    & \makecell{\textcolor{MyGreen}{\checkmark}} 
    & \makecell{\textcolor{MyGreen}{\checkmark}}
    & \makecell{\xmark}
    & \makecell{\xmark}
    \\
    NeuMan \cite{neural-human-radiance-field}
    & \makecell{\textcolor{MyGreen}{\checkmark}} 
    & \makecell{\textcolor{MyGreen}{\checkmark}} 
    & \makecell{\xmark}  
    & \makecell{\textcolor{MyGreen}{\checkmark}} 
    & \makecell{\xmark}
    & \makecell{\xmark}
    \\
    SLAHMR \cite{ye2023slahmr}
    & \makecell{\xmark} 
    & \makecell{\textcolor{MyGreen}{\checkmark}} 
    & \makecell{\xmark}  
    & \makecell{\textcolor{MyGreen}{\checkmark}} 
    & \makecell{\xmark}
    & \makecell{\xmark}
    \\
    HyperNeRF \cite{park2021hypernerf}
    & \makecell{\textcolor{MyGreen}{\checkmark}}
    & \makecell{\textcolor{MyGreen}{\checkmark}}
    & \makecell{\textcolor{MyGreen}{\checkmark}} 
    & \makecell{\xmark}
    & \makecell{\xmark}
    & \makecell{\xmark}
    \\
    D$^2$NeRF \cite{d2nerf}
    & \makecell{\textcolor{MyGreen}{\checkmark}}
    & \makecell{\textcolor{MyGreen}{\checkmark}} 
    & \makecell{\textcolor{MyGreen}{\checkmark}} 
    & \makecell{\xmark}
    & \makecell{\xmark}
    & \makecell{\xmark}
    \\
    DynIBaR \cite{li2023dynibar}
    & \makecell{\textcolor{MyGreen}{\checkmark}} 
    & \makecell{\textcolor{MyGreen}{\checkmark}} 
    & \makecell{\textcolor{MyGreen}{\checkmark}} 
    & \makecell{\xmark} 
    & \makecell{\textcolor{MyGreen}{\checkmark}}
    & \makecell{\xmark}\\
    SUDS \cite{turki2023suds}
    & \makecell{\textcolor{MyGreen}{\checkmark}} 
    & \makecell{\textcolor{MyGreen}{\checkmark}} 
    & \makecell{\textcolor{MyGreen}{\checkmark}} 
    & \makecell{\xmark} 
    & \makecell{\textcolor{MyGreen}{\checkmark}}
    & \makecell{\xmark}\\
    \hline
    Ours
    & \makecell{\textcolor{MyGreen}{\checkmark}} 
    & \makecell{\textcolor{MyGreen}{\checkmark}} 
    & \makecell{\textcolor{MyGreen}{\checkmark}} 
    & \makecell{\textcolor{MyGreen}{\checkmark}} 
    & \makecell{\textcolor{MyGreen}{\checkmark}}
    & \makecell{\textcolor{MyGreen}{\checkmark}}\\
    \hline
\end{tabularx}
}

%% file: 3_method.tex
\section{Method}
\label{sec:method}

\subsection{Limitations of Prior Art}
The state-of-the-art monocular deformable NeRFs \cite{park2021hypernerf, d2nerf} decompose a deformable scene into a rigid, canonical template model and a deformation field $\mathbf{J}^{t, \leftarrow}$ that maps the world space $\mathbf{G}_\textrm{0}^{t} \mathbf{X}^{t}$ to the canonical space $\mathbf{X}^*$, where ${\bf G}^t_0$ is the \textit{known} camera pose at time $t$, and $\mathbf{X}^t$ is a camera space point at time $t$:
\begin{align}
\label{eq:obj_motion_modeling_baselines}
&\mathbf{X}^*=\mathcal{W}^{t, \leftarrow}\left(\mathbf{X}^{t}\right)=\mathbf{J}^{t, \leftarrow} ( \mathbf{G}_0^{t} \mathbf{X}^{t}).
\end{align}

\noindent In theory, this formulation is sufficient to represent all continuous motion; it performs well on short videos containing near-rigid scenes, as the deformation field only has to learn minute deviations from the template model. However, this motion model is difficult to scale to minute-long videos, which are more likely to contain deformable objects undergoing large translations (e.g., a person walking into another room) and pose changes (e.g., a person sitting down). Here, the deformation field must learn large deviations from the canonical model, significantly complicating optimization. 

Another critical limitation of HyperNeRF and D$^2$NeRF is that they cannot track separate deformable objects and therefore cannot perform 3rd-person-follow view synthesis for scenes with \textit{multiple} actors.

\subsection{Component Radiance Fields}
\label{sec:method_representation}

To address the limitations of existing monocular deformable NeRFs, we propose \ourmethod{}, a novel 3D representation that models a deformable scene as a composition of $M$ object-centric neural fields, one for the rigid background and each of the $M-1$ deformable objects (Figure~\ref{fig:method}). Crucially, \ourmethod{} hierarchically decomposes scene motion into the motion of each object, which itself is decomposed into global root-body motion and local articulations. This key design choice scales our method to minute-long videos containing highly dynamic and deformable objects.

\paragraph{Background Radiance Field.} We begin by modeling the background environment as a Neural Radiance Field (NeRF) \cite{mildenhall2020nerf}. For a 3D point ${\bf X}^* \in \mathbb{R}^3$ and a viewing direction ${\bf v}^*$ in the canonical world space, NeRF defines a color $\mathbf{c}$ and density $\mathbf{\sigma}$ represented by an MLP. %
We follow contemporary variants~\cite{martinbrualla2020nerfw} that include a time-specific embedding code $\omega_e^t$ to model illumination changes over time and model density with as a function of a neural signed distance function (SDF) $\mathbf{MLP}_{\mathbf{\sigma}}(\cdot) = \alpha\Gamma_{\beta}(\mathbf{MLP}_\textrm{SDF}(\cdot))$~\cite{yariv2021volume} to encourage the reconstruction of a valid surface:
\begin{align}
    \label{eq:nerf}
    \sigma = \mathbf{MLP}_{\mathbf{\sigma}}(\mathbf{X}^*), \qquad
    \mathbf{c}^{t} = \mathbf{MLP}_{\mathbf{c}}(\mathbf{X}^*, \mathbf{v}^*, \mathbf{\omega}_{e}^{t}).
\end{align}
The pixel color can then be computed with differentiable volume rendering equations (Section~\ref{sec:method_rendering}).

Most NeRF methods, including HyperNeRF~\cite{park2021hypernerf} and D$^2$NeRF~\cite{d2nerf}, assume images with known cameras. While our capture devices are equipped with inertial sensors, we find their self-reported camera poses have room for improvement. As such, we also model camera pose as an \textit{optimizable} rigid-body transformation ${\bf G}_0^t \in SE(3)$ that maps points in a time-specific camera space ${\bf X}^t \in \mathbb{R}^3$ to the world space (where we assume homogenous notation):
\begin{align}
    {\bf X}^* = {\bf G}^t_0 {\bf X}^t. \label{eq:camera}
\end{align}

\paragraph{Deformable Field (for Object $j$).}
We model the deformable radiance field of object $j \in \{1, \cdots, M-1\}$ with BANMo~\cite{yang2022banmo}, which consists of a canonical rest shape and time-\textit{dependent} deformation field. %
The canonical rest shape is represented by the same formulation described by Equation~\ref{eq:nerf}, but now defined in a local \textit{object-centric canonical space} rather than the world space. BANMo represents object motion with a warping function $\mathcal{W}_{j}^{t, \leftarrow} : \mathbf{X}^{t} \rightarrow \mathbf{X}^{*}_j$ that maps the camera space points $\mathbf{X}^{t}$ to canonical space points $\mathbf{X}^{*}_j$ with a rigid-body transformation $\mathbf{G}_j^{t} \in SE(3)$ and a deformation field $\mathbf{J}_{j}^{t, \leftarrow}$ modeled by linear blend skinning~\cite{lbs}:

\begin{align}
\label{eq:backward_warp}
&\mathbf{X}_{j}^*=\mathcal{W}_j^{t, \leftarrow}\left(\mathbf{X}^{t}\right)=\mathbf{J}^{t, \leftarrow}_j \left( \mathbf{G}_j^{t} \mathbf{X}^{t}\right).
\end{align}

Note that our choice of deformation field differs from the $SE(3)$-field used in HyperNeRF and D$^2$NeRF, which has been shown to produce irregular deformation in the presence of complex scene motion \cite{yang2022banmo}. Intuitively, rigid-body transformation $\mathbf{G}_j^{t}$ captures the global root-body pose of object $j$ relative to the camera at time $t$, while deformation field $\mathbf{J}^{t, \leftarrow}_j$ aligns more fine-grained articulations relative to its local canonical space (Figure~\ref{fig:method}). Explicitly disentangling these two sources of object motion (as opposed to conflating them) enables easier optimization of the deformation field, because local articulations are significantly easier to learn than those modeled relative to the world space (Equation~\ref{eq:obj_motion_modeling_baselines}). Furthermore, this motion decomposition makes the deformation field invariant to rigid-body transformations of the object. A motion model similar to ours was proposed by ST-NeRF \cite{zhang2021stnerf}, but their model encodes an object’s global root-body motion with a 3D axis-aligned bounding box that does not explicitly represent object orientation, a prerequisite for embodied view synthesis from 3rd-person-follow cameras.

As did BANMo, \ourmethod{} also models a forward warp $\mathbf{X}_{j}^t=\mathcal{W}_j^{t, \rightarrow}\left(\mathbf{X}^{*}\right)= \left(\mathbf{G}_j^{t}\right)^{-1}\mathbf{J}^{t, \rightarrow}_j\left(\mathbf{X}^{*}\right)$ that maps the canonical space to the camera space, which is used to establish the surface correspondences required for egocentric view synthesis and 3D video filters.

\subsection{Composite Rendering of Multiple Objects}
\label{sec:method_rendering}
Given a set of $M$ object representations (the background is treated as an object as well), we use the composite rendering scheme from prior work~\cite{Niemeyer2020GIRAFFE, obsurf} to combine the outputs of all object representations and volume-render the entire scene. To volumetrically render the image at frame $t$, we sample multiple points along each camera ray $\mathbf{v}^t$. Denoting the $i^{th}$ sample as  ${\bf X}^t_{i}$, we write the density and color observed at sample $i$ due to object $j$ as:
\begin{align}
    \sigma_{ij}=\mathbf{MLP}_{\sigma, j}\left(\mathbf{X}_{ij}^{*} \right), \qquad
    \mathbf{c}_{ij}=\mathbf{MLP}_{\mathbf{c}, j}\left(\mathbf{X}_{ij}^{*}, \mathbf{v}^{*}_{j},  \mathbf{\omega}_{e}^{t}\right), \nonumber
\end{align} 
\noindent where $\mathbf{X}_{ij}^{*} = \mathcal{W}_{j}^{t, \leftarrow}\left(\mathbf{X}_{i}^{t}\right)$ and $\mathbf{v}^{*}_{j} = \mathcal{W}_{j}^{t, \leftarrow}(\mathbf{v}^{t}$) are sample $i$ and camera ray $\mathbf{v}^{t}$ backward-warped into object $j$'s canonical space, respectively. The composite density $\sigma_{i}$ at sample $i$ along the ray is then computed as the sum of each object's density $\sigma_{ij}$; the composite color $\mathbf{c}_{i}$ is computed as the weighted sum of each object's color $\mathbf{c}_{ij}$, where the weights are the normalized object densities $\sigma_{ij} / \sigma_{i}$:
\begin{align}
    \label{eq:composite}
    &\sigma_{i} = \sum_{j=0}^{M-1}\sigma_{ij},\quad \mathbf{c}_{i} = \frac{1}{\sigma_{i}}\sum_{j=0}^{M-1}\sigma_{ij}\mathbf{c}_{ij}.
\end{align} 
\noindent We can then use the standard volume rendering equations to generate an RGB image of the scene, where $N$ is the number of sampled points along camera ray $\mathbf{v}^{t}$, $\tau_{i}$ is the transmittance, $\alpha_{i}$ is the alpha value for sample point $i$ and $\delta_{i}$ is the distance between sample point $i$ and the $(i+1)$: %
\begin{align}
\mathbf{\hat{c}} = \sum_{i=1}^{N}\tau_{i}\alpha_{i}\mathbf{c}_{i}, \quad  \tau_{i} = \prod_{k=1}^{i-1}(1-\alpha_k), \quad \alpha_i = 1 - e^{-\sigma_i \delta_i}. \nonumber
\end{align}
\begin{table}[t]
\centering 
\resizebox{\linewidth}{!}{
\begin{tabular}{l | l}
\toprule
Rendered features $\hat{\mathbf{f}}$ at pixel $\mathbf{x}^{t}$ & Corresponding 3D features $\mathbf{f}_{ij}(\mathbf{X}_{i}^{t})$\\[0.5ex] 
\midrule 
color $\mathbf{\hat{c}}(\mathbf{x}^{t})$ & $\mathbf{c}_{i}^{t}\left(\mathcal{W}_{j}^{t, \ \leftarrow}(\mathbf{X}_{i}^{t})\right)$\\ 
flow $\hat{\mathcal{F}}(\mathbf{x}^{t}, t\rightarrow t')$  & $\Pi^{t'}\left(\mathcal{W}_{j}^{t', \ \rightarrow}\left(\mathcal{W}_{j}^{t, \ \leftarrow}(\mathbf{X}_{i}^{t})\right)\right) - \mathbf{x}^{t}$ \\
depth $\mathbf{\hat{d}}(\mathbf{x}^{t})$ & $[0, \ 0, \ 1]\cdot \mathbf{X}_{i}^{t}$ \\    \bottomrule
\end{tabular}
\caption{Rendered 2D features $\hat{\mathbf{f}}$ and their corresponding 3D features $\mathbf{f}_{ij}$. $\Pi^{t'}$ denotes the camera intrinsics at time $t'$. \captionspace}
\vspace{-1em}
\label{table:rendered_features}
}
\end{table}
\begin{figure}[t]
    \captionsetup[subfigure]{labelformat=empty}
    \centering
    \subfloat{{\includegraphics[width=\textwidth,clip=true,trim = 9mm 0mm 0mm 0mm]{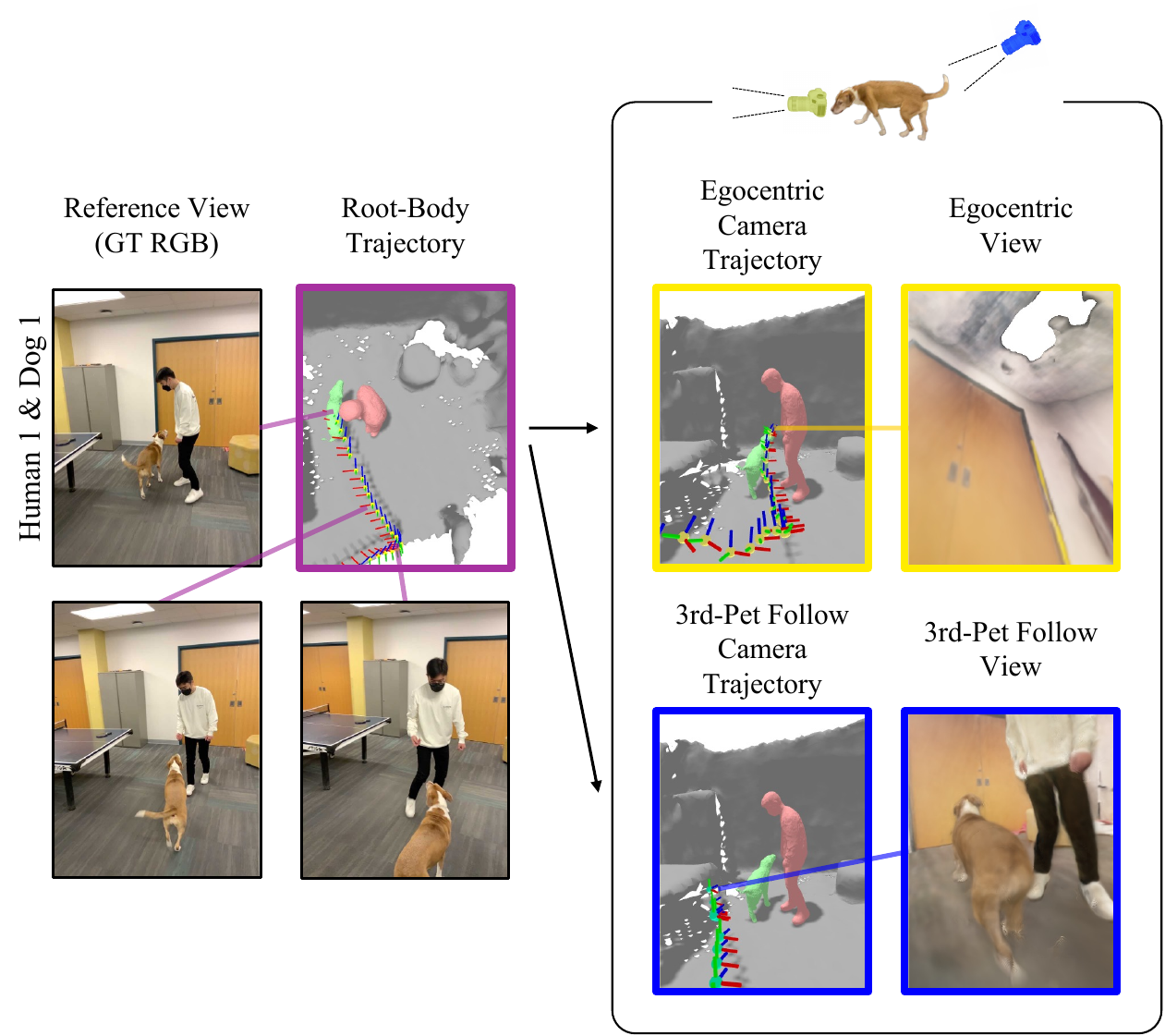}}}%
    \caption{\textbf{6-DOF Trajectories for Embodied View Synthesis.} To synthesize embodied views from egocentric or actor-following cameras, \ourmethod{} reconstructs the entire background, every individual actor in the scene, as well as global 6-DOF trajectories of its root-body and its articulated body parts (e.g. head). \href{https://andrewsonga.github.io/totalrecon/\#trajectories}{\textbf{[Videos]}} \captionspace}%
    \label{fig:render_traj}
\end{figure}

\paragraph{Rendering Flow, Depth, and Silhouettes.}
Our composite rendering scheme can be used to render different quantities by replacing the object color $\mathbf{c}_{ij}$ in Equation~\ref{eq:composite} with the appropriately defined 3D \textit{feature} $\mathbf{f}_{ij}$ (Table \ref{table:rendered_features}) and rendering the resulting composite feature $\mathbf{f}_{i}$. %
To render occlusion-aware object silhouettes, we follow ObSURF \cite{obsurf} to produce a categorical distribution over the $M$ objects:
\begin{align}
    &\mathbf{\hat o}_{j} = \sum_{i=1}^{N}\tau_{i}\alpha_{ij}, \quad \textrm{where} \quad \tau_{i} = \prod_{k=1}^{i-1}(1-\alpha_k), \\ &\alpha_i = 1 - e^{-\sigma_i \delta_i}, \qquad \alpha_{ij} = 1 - e^{-\sigma_{ij} \delta_i}.
\end{align}
\paragraph{Losses.}
\label{sec:method_optimization}
Given a monocular RGBD video, we optimize all parameters in our composite scene representation, which for each of the $M$ objects includes the appearance and shape MLPs ($\mathbf{MLP}_{c,j}$, $\mathbf{MLP}_{\sigma,j}$), rigid-body transformations ${\mathbf{G}}^t_j$, and forward, backward deformation fields $\mathbf{J}_j^{\leftarrow}$, $\mathbf{J}_j^{\rightarrow}$. We optimize three reconstruction losses: a color loss $\mathcal{L}_\textrm{rgb}$, a flow loss $\mathcal{L}_\textrm{flow}$, and a depth loss $\mathcal{L}_\textrm{depth}$, where the ground truth color $\mathbf{c}$ and depth $\mathbf{d}$ are provided by the RGBD video, and the ``ground truth'' flow $\mathcal{F}$ is computed by an off-the-shelf network \cite{yang2019volumetric}. The model also optimizes a 3D-cycle consistency loss $\mathcal{L}_{\textrm{cyc,}j}$ \cite{yang2022banmo} for each deformable object to encourage their forward and backward warps to be consistent, where ${\bf x}^t \in \mathbb{R}^2$ denotes pixel location at time $t$:
\begin{align}
    \label{eq:rgb_loss}
    \mathcal{L}_\textrm{rgb} &= \sum_{\mathbf{x}^t}||\mathbf{c}(\mathbf{x}^{t})-\hat{\mathbf{c}}(\mathbf{x}^{t})||^{2},\\
    \label{eq:flow_loss}
    \mathcal{L}_\textrm{flow} &= \sum_{\mathbf{x}^t}||\mathcal{F}(\mathbf{x}^{t})-\hat{\mathcal{F}}(\mathbf{x}^{t})||^{2}, \\
    \label{eq:depth_loss}
    \mathcal{L}_\textrm{depth} &= \sum_{\mathbf{x}^t}||\mathbf{d}(\mathbf{x}^{t})-\hat{\mathbf{d}}(\mathbf{x}^{t})||^{2}, \\
    \mathcal{L}_{\textrm{cyc,}j} &= \sum_{i}\tau_{i}\alpha_{ij}\left\lVert \mathcal{W}_{j}^{t', \ \rightarrow}\left(\mathcal{W}_{j}^{t, \ \leftarrow}(\mathbf{X}_{i}^{t})\right) - \mathbf{X}_{i}^{t}\right\rVert^{2}.
    \label{eq:cycle_loss}
\end{align}
\paragraph{Initialization.} We initialize the rigid-body transformations of each deformable object $\mathbf{G}^{t}_{j}$ using a pre-trained category-specific PoseNet \cite{yang2022banmo}; we initialize the rigid-body transformation of the background $\mathbf{G}^{t}_{0}$ with the camera poses provided by the iPad Pro. 
\begin{figure*}[htbp]
    \captionsetup[subfigure]{labelformat=empty}
    \centering
    \subfloat{{\includegraphics[width=0.95\linewidth]{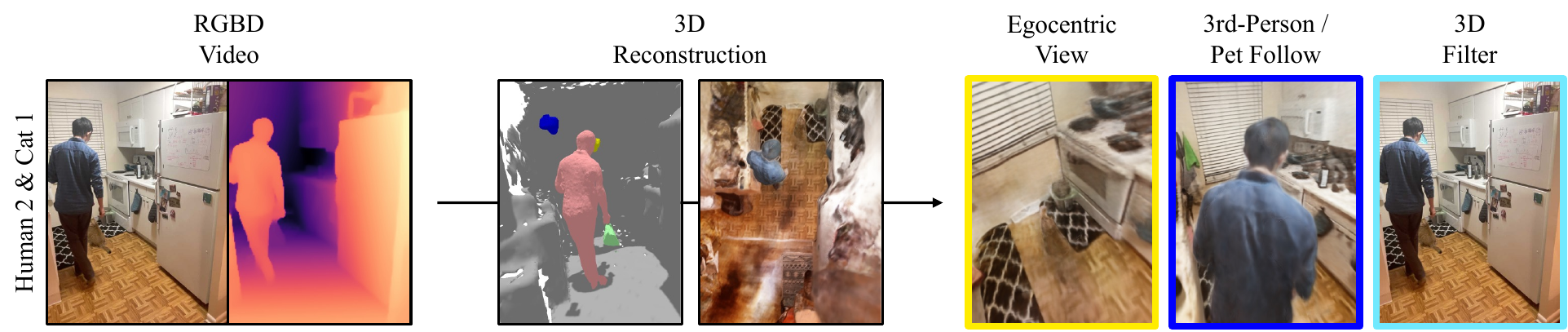}}}\\%
    \subfloat{{\includegraphics[width=0.95\linewidth]{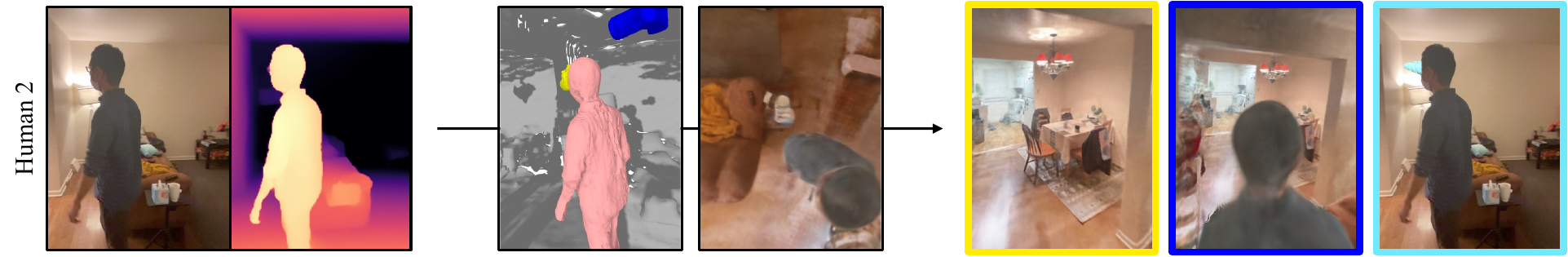}}}\\%
    \subfloat{{\includegraphics[width=0.95\linewidth]{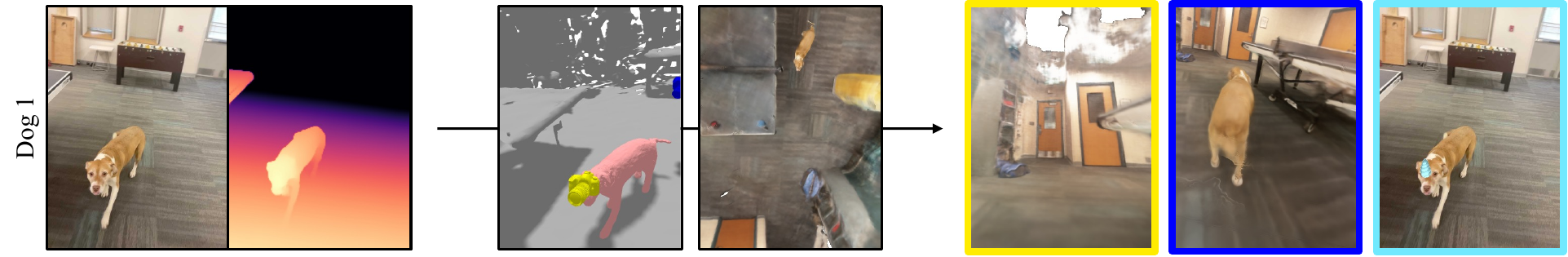}}}\\%
    \subfloat{{\includegraphics[width=0.95\linewidth]{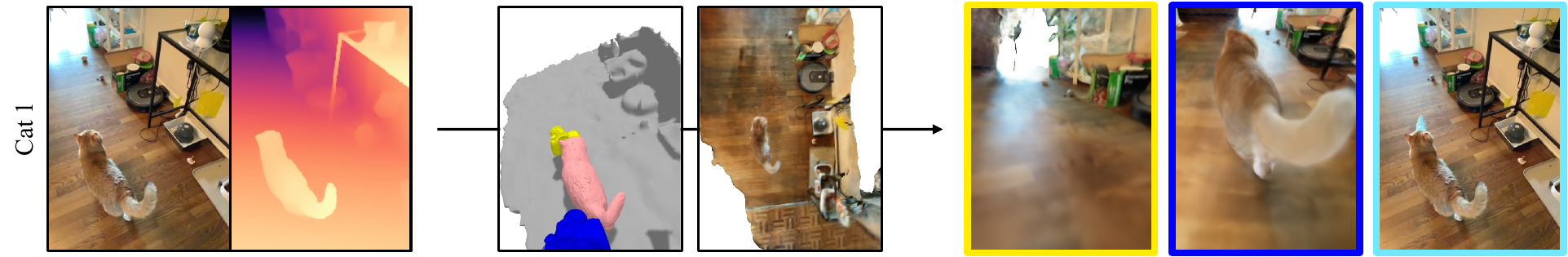}}}\\%
    \subfloat{{\includegraphics[width=0.95\linewidth]{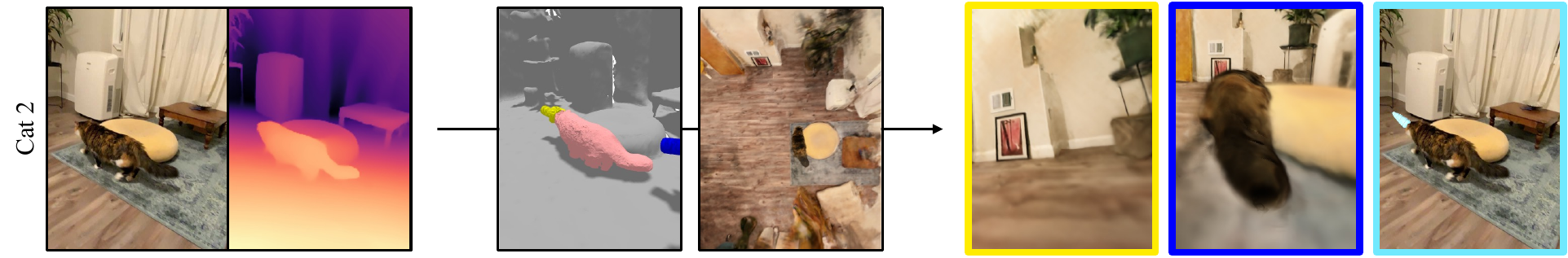}}}\\%
    \caption{\textbf{Embodied View Synthesis and 3D Filters.} For select sequences of our RGBD dataset, we visualize the scene geometry and appearance reconstructed by our method (3D reconstruction) and the resulting downstream applications. The yellow and blue camera meshes in the mesh renderings represent the egocentric and 3rd-person-follow cameras, respectively. To showcase the 3D video filter we attach a sky-blue unicorn horn to the forehead of the target object, which is then automatically propagated across all frames. \href{https://andrewsonga.github.io/totalrecon/applications.html}{\textbf{[Videos]}}\captionspace
    \vspace{-1 em}
    }%
    \label{fig:extremeview}
\end{figure*}
\paragraph{Embodied View Synthesis and 3D Filters.}
To enable embodied view synthesis and 3D video filters (Figure~\ref{fig:render_traj}), we design a simple interface that allows a user to select a point on a target object's surface in its reconstructed canonical mesh, and use its forward warping function $\mathcal{W}_j^{t, \rightarrow}: \mathbf{X}^{*} \rightarrow \mathbf{X}^{t}$ followed by the rigid-body transformation ${\bf G}^t_0$ to place the egocentric camera (or virtual 3D asset) in the world space. The surface normal to the object's mesh at the user-defined point provides a reference frame to align the egocentric camera's viewing direction and place the 3D asset. To implement a 3rd-person-follow camera, we add a user-defined offset to the object's local reference frame, which is defined by its root-body pose.

%% file: 4_experiments.tex
\section{Experiments}

\paragraph{Implementation Details.}
\label{sec:implementation details}

In practice, we train our composite scene representation by first pre-training each object field separately. 
For deformable objects, we pre-train using a depth loss (Equation \ref{eq:depth_loss}) combined with the losses optimized by BANMo \cite{yang2022banmo}. This includes a silhouette loss $\mathcal{L}_\textrm{mask} = \sum_{\mathbf{x}^t}||\mathbf{o}_j(\mathbf{x}^t) - \hat{\mathbf{o}}_j(\mathbf{x}^t)||^2$, where the ``ground truth" object silhouette $\mathbf{o}_j$ is computed by an off-the-shelf instance segmentation engine \cite{kirillov2020pointrend}. For pre-training the background, we optimize color, flow, and depth losses (Equations~\ref{eq:rgb_loss}, \ref{eq:flow_loss}, \ref{eq:depth_loss}) on pixels outside the ground truth object silhouettes. Importantly, we don't supervise the object fields on frames that are not provided an object silhouette since it cannot be determined whether the absence of detection is a true or false negative. 

After pre-training, we composite-render the pre-trained object fields and jointly finetune them using only the color, depth, flow, and object-specific 3D-cycle consistency losses. Since the silhouette loss is no longer used, the scene representation is supervised on \textit{all} frames of the training sequence during joint-finetuning. We provide the complete description of implementation details in Appendix \ref{sec:implementation details_supp}.

\paragraph{Dataset.}
\label{sec:dataset}

We evaluate \ourmethod{} on novel-view synthesis for deformable scenes. To enable quantitative evaluation, we built a stereo rig comprised of two iPad-Pros rigidly attached to a camera mount, a setup similar to that of Nerfies \cite{park2021nerfies}.
Using the stereo rig, we captured 11 RGBD sequences containing 3 different cats, 1 dog, and 2 human subjects in 4 different indoor environments. The RGBD videos were captured using the Record3D iOS App \cite{record3d}, which also automatically registers the frames captured by each camera. These video sequences, which were subsampled at 10 fps, range from 392 to 901 frames, amount to, on average minute-long videos that are significantly longer and contain more dynamic motion than the datasets introduced by \cite{park2021nerfies, park2021hypernerf, d2nerf, gao2022dynamic}. The left and right cameras were registered by solving a Perspective-n-Point (PnP) problem using manually annotated correspondences, and their videos were synchronized based on audio. We provide a complete description of our dataset in Appendix \ref{sec:dataset_details}. 

\begin{figure}[!t]
    \captionsetup[subfigure]{labelformat=empty}
    \centering
    \subfloat{{\includegraphics[width=\linewidth, clip=true,trim = 12mm 0mm 0mm 0mm]{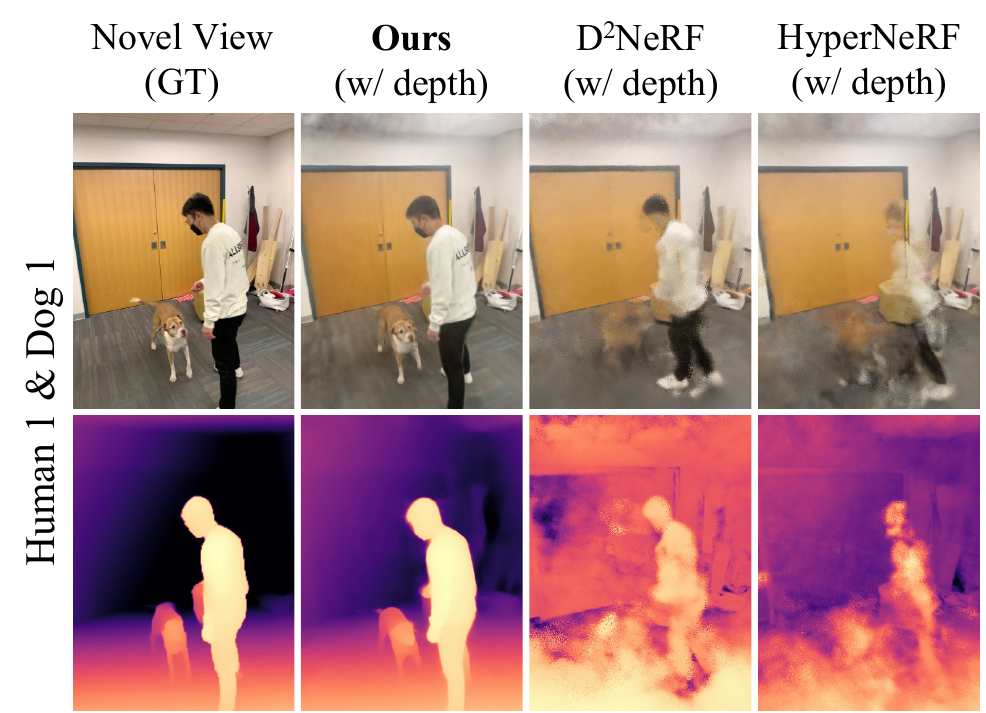}}}%
    \caption{\textbf{Baseline Comparisons.} We compare \ourmethod{} to depth-supervised variants of HyperNeRF \cite{park2021hypernerf} and D$^2$NeRF \cite{d2nerf} on the task of stereo-view synthesis (the left camera is used for training and the right is used for testing). While the baselines are able to reconstruct only the background, \ourmethod{} can reconstruct \textit{both} the background and the moving deformable object(s), demonstrating holistic scene reconstruction. \href{https://andrewsonga.github.io/totalrecon/nvs.html}{\textbf{[Videos]}}\captionspace}%
    \label{fig:stereoview}
\end{figure}

\paragraph{Reconstruction and Applications.}
By hierarchically decomposing scene motion into the motion of each object, which itself is decomposed into root-body motion and local articulations, \ourmethod{} \textit{automatically} computes novel 6-DoF trajectories such as those traversed by egocentric cameras and 3rd-person follow cameras (Figure \ref{fig:render_traj}). In turn, these trajectories enable automated embodied view synthesis and 3D occlusion-aware video filters (Figure \ref{fig:extremeview}). These tasks are also enabled by \ourmethod{}'s ability to recover an accurate deformable 3D scene representation, which is currently out of reach for the best of related methods (Figure \ref{fig:stereoview}). As shown in the bird's eye view, each reconstructed object is properly situated with respect to the background and other objects, a direct consequence of our use of depth supervision. Furthermore, even though the iPad Pro can only measure depth up to 4m, \ourmethod{} can \textit{render} depth \textit{beyond} this sensor limit by pooling the measurements from other frames into a single metric scene reconstruction. We provide results on additional sequences in Appendix \ref{sec:reconstruction and applications_supp}.

\paragraph{Baselines and Evaluation.}
\label{sec:nvs_comparisons}

In Figure \ref{fig:stereoview} and Table \ref{table:nvs_comparisons_entireimg}, we compare \ourmethod{} to D$^{2}$NeRF \cite{d2nerf} and HyperNeRF \cite{park2021hypernerf}, and their depth-supervised equivalents on the proxy task of stereo-view synthesis, a prerequisite for \textit{embodied} view synthesis: we train each method on the RGBD frames captured from the left camera of our dataset and evaluate the images rendered from the viewpoint of the right camera. The depth-supervised versions of the baselines contain the same depth loss used in \ourmethod{}. We report LPIPS \cite{lpips} and the average (depth) accuracy at 0.1m \cite{Rematas_2022_CVPR} in all subsequent experiments, and we include a more complete set of metrics (PSNR, SSIM, RMS depth error) in Appendix \ref{sec:baseline_setup}. Because D$^{2}$NeRF and HyperNeRF were not designed to recover a \textit{metric} scene representation, we replaced their COLMAP \cite{Schonberger_2016_colmap} camera poses with those provided by the iPad Pro (which are metric measurements) for the sake of fair comparison.
\paragraph{Comparisons.}

\ourmethod{} qualitatively and quantitatively outperforms all of the baselines. As shown in Figure \ref{fig:stereoview}, \ourmethod{} successfully reconstructs the entire scene, whereas the baselines are only able to reconstruct the rigid background at best. As shown in Table \ref{table:nvs_comparisons_entireimg}, \ourmethod{} significantly outperforms all baselines in terms of LPIPS and the average accuracy at 0.1m (Acc@0.1m). We attribute this huge gap to the baselines' inability to reconstruct highly dynamic objects. We provide more details regarding the baselines and additional visualizations in Appendix \ref{sec:baseline_setup}.

\begin{table*}[t]
\caption{\textbf{Baseline Comparisons}. We train \ourmethod{}, HyperNeRF \cite{park2021hypernerf}, D$^2$NeRF \cite{d2nerf}, and their depth-supervised variants on the \emph{left} video captured with our stereo rig, and evaluate the novel view synthesis results on the \emph{held-out} right video. \ourmethod{} significantly outperforms all of the baselines for all 11 sequences. These sequences are sampled at 10 fps, amounting to minute-long videos, on average.
}
    \label{table:nvs_comparisons_entireimg}
\centering
\resizebox{\linewidth}{!}{
\setlength{\tabcolsep}{1pt}
\input{tables/comparisons_entireimg_allseqs_with_acc.tex}
}
\end{table*}

\subsection{Ablation Studies}
\label{sec:ablations}

\begin{figure}[t]
    \captionsetup[subfigure]{labelformat=empty}
    \centering
    \subfloat{{\includegraphics[width=\linewidth,clip=true,trim=15mm 0mm 0mm 0mm]{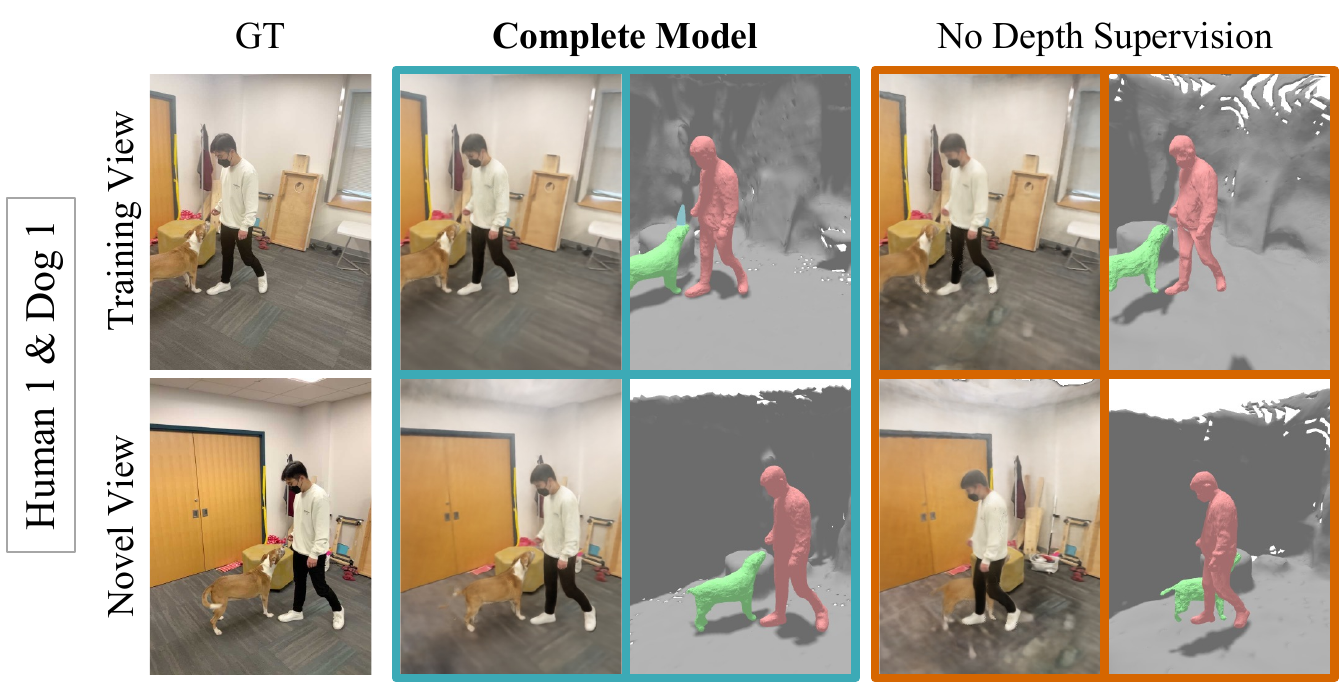}}}
    \caption{\textbf{Ablation Study on Depth Supervision.} 
    While removing depth supervision does not significantly deterioriate the training-view RGB renderings, it does hurt the \textit{novel-view} 3D reconstructions, as characterized by the following:
    (a) floating foreground objects (as evidenced by their shadows) and (b) the human incorrectly occluding the dog. These failure modes indicate that without depth supervision, \ourmethod{} overfits the training view, and the reconstructed objects fail to converge to the same scale. \href{https://andrewsonga.github.io/totalrecon/ablation_depth.html}{\textbf{[Videos]}}\captionspace
    }%
    \label{fig:ablation_depth_humandog}
\end{figure}

Table \ref{table:ablation_avg} (and Figures~\ref{fig:ablation_depth_humandog} and \ref{fig:ablation_humancat}) analyzes the importance of \ourmethod~'s design choices (see Section \ref{sec:method}) by ablating its key components: the depth loss $\mathcal{L}_\textrm{depth}$ (row 2), the deformation field $\mathbf{J}_{j}^{t}$ (row 3), PoseNet-initialization of the root-body pose (row 4), and the root-body pose $\mathbf{G}_{j}^{t}$ itself (row 5), where $j$ denotes a deformable actor. For all ablations, we use the same set of training losses used in \ourmethod~ and initialize camera pose $\mathbf{G}^t_0$ with those reported by ARKit. For ablations that model root-body motions, we initialize each deformable actor's root-body pose $\mathbf{G}^t_j$ with predictions made by PoseNet \cite{yang2022banmo} and optimize them during reconstruction; for row 4, we replace the PoseNet predictions with identity rotations. We report the novel-view metrics averaged over 6 select sequences of our dataset: \textsc{dog 1} (v1), \textsc{cat 1} (v1), \textsc{cat 2} (v1), \textsc{human 1}, \textsc{human 1 \& dog 1}, and \textsc{human 2 \& cat 1}.

\begin{table}[t]
\caption{\textbf{Ablation Study.} Removing depth supervision (2) significantly hurts performance, while removing the deformation field (3) and PoseNet-initialization of root-body poses (4) hurts moderately. Most importantly, removing root-body poses entirely (5) prevents convergence (N/A) as the deformation field alone has to explain \textit{global} object motion (see Figure~\ref{fig:method}). These experiments justify our hierarchical modeling of motion, as even root-bodies without a deformation field (3) or poorly initialized root-bodies (4) are better than no root-bodies (5). We visualize these ablations in Figure~\ref{fig:ablation_humancat} and explore other ablations in Appendix \ref{sec:ablation_supp}.}
\label{table:ablation_avg}
\centering
\resizebox{\linewidth}{!}{
\setlength{\tabcolsep}{2pt}
\input{tables/ablation_avg_wo_freezecam_col_with_acc.tex}
}
\end{table}

\paragraph{Depth Supervision.}

Table \ref{table:ablation_avg} shows that removing depth supervision (row 2) significantly reduces the average accuracy at 0.1m (Acc). Figure \ref{fig:ablation_depth_humandog} indicates that this reflects the incorrect arrangement of objects stemming from their scale inconsistency - while removing depth supervision does not significantly deteriorate the training-view RGB renderings, it induces critical failure modes as shown in the \textit{novel-view} 3D reconstructions: (a) floating foreground objects, as evidenced by their shadows, and (b) the human incorrectly occluding the dog. In other words, without depth supervision, \ourmethod{} overfits the training view and learns a degenerate scene representation where the reconstructed objects fail to converge to the same scale. We show results on additional RGBD sequences in Appendix \ref{sec:ablation_depth_supp}.

\paragraph{Motion Modeling.} Table \ref{table:ablation_avg} shows that removing the deformation field (row 3) also hurts performance. This is because, without the deformation field, our method has to explain an object's non-rigid motion solely with its rigid, root-body poses. As a result, this ablation can only recover coarse object reconstructions that fail to model moving body parts such as limbs. Removing PoseNet-initialization of root-body poses (row 4) is just as detrimental, resulting in noisy appearance and geometry artifacts; see Figure \ref{fig:ablation_humancat} and additional visualizations in Figure \ref{fig:ablation_objmotion_full}. Most notably, Table \ref{table:ablation_avg} shows that removing object root-bodies entirely (row 5) causes the optimization to fail to converge (N/A), even though the deformation field can (in theory) represent all continuous motion.
It appears difficult for deformation fields alone to explain \textit{global} root-body motion because such motions can deviate significantly from a canonical model,  complicating optimization.

These diagnostics justify \ourmethod{}'s hierarchical motion representation, which explicitly models objects' root-body poses; root-bodies without articulated deformations (row 3) or poorly initialized root-bodies (row 4) are better than no root-bodies at all (row 5). Our ablations also suggest that the poor performance of the baseline methods (on our challenging dataset) may be attributed to the lack of object-centric motion modeling. We provide a more detailed analysis with additional experiments and RGBD sequences in Appendix \ref{sec:ablation_objmotion_supp}.

%% file: tables/comparisons_entireimg_allseqs_with_acc.tex
\small{
\begin{tabular}{l||cc|cc|cc|cc|cc|cc|cc|cc|cc|cc|cc|cc}

\toprule

& \multicolumn{ 2 }{c}{
  \makecell{
  \textsc{\small Dog 1}
\\\small(626 images)
  }
}
& \multicolumn{ 2 }{c}{
  \makecell{
  \textsc{\small Dog 1 (v2)}
\\\small(531 images)
  }
}
& \multicolumn{ 2 }{c}{
  \makecell{
  \textsc{\small Cat 1}
\\\small(641 images)
  }
}
& \multicolumn{ 2 }{c}{
  \makecell{
  \textsc{\small Cat 1 (v2) }
\\\small(632 images)
  }
}
& \multicolumn{ 2 }{c}{
  \makecell{
  \textsc{\small Cat 2}
\\\small(834 images)
  }
}
& \multicolumn{ 2 }{c}{
  \makecell{
  \textsc{\small Cat 2 (v2) }
\\\small(901 images)
\\
  }
}
& \multicolumn{ 2 }{c}{
  \makecell{
  \textsc{\small Cat 3}
\\\small(767 images)
  }
}
& \multicolumn{ 2 }{c}{
  \makecell{
  \textsc{\small Human 1 }
\\\small(550 images)
  }
}
& \multicolumn{ 2 }{c}{
  \makecell{
  \textsc{\small Human 2 }
\\\small(483 images)
  }
}
& \multicolumn{ 2 }{c}{
  \makecell{
  \textsc{\small Human - Dog}
\\\small(392 images)
  }
}
& \multicolumn{ 2 }{c|}{
  \makecell{
  \textsc{\small Human - Cat}
\\\small(431 images)
  }
}
& \multicolumn{ 2 }{c}{
  \makecell{
  \textsc{\small Mean }
  }
}
\\

& \multicolumn{1}{c}{ \footnotesize LPIPS$\downarrow$}
& \multicolumn{1}{c}{ \footnotesize Acc$\uparrow$}
& \multicolumn{1}{c}{ \footnotesize LPIPS$\downarrow$}
& \multicolumn{1}{c}{ \footnotesize Acc$\uparrow$}
& \multicolumn{1}{c}{ \footnotesize LPIPS$\downarrow$}
& \multicolumn{1}{c}{ \footnotesize Acc$\uparrow$}
& \multicolumn{1}{c}{ \footnotesize LPIPS$\downarrow$}
& \multicolumn{1}{c}{ \footnotesize Acc$\uparrow$}
& \multicolumn{1}{c}{ \footnotesize LPIPS$\downarrow$}
& \multicolumn{1}{c}{ \footnotesize Acc$\uparrow$}
& \multicolumn{1}{c}{ \footnotesize LPIPS$\downarrow$}
& \multicolumn{1}{c}{ \footnotesize Acc$\uparrow$}
& \multicolumn{1}{c}{ \footnotesize LPIPS$\downarrow$}
& \multicolumn{1}{c}{ \footnotesize Acc$\uparrow$}
& \multicolumn{1}{c}{ \footnotesize LPIPS$\downarrow$}
& \multicolumn{1}{c}{ \footnotesize Acc$\uparrow$}
& \multicolumn{1}{c}{ \footnotesize LPIPS$\downarrow$}
& \multicolumn{1}{c}{ \footnotesize Acc$\uparrow$}
& \multicolumn{1}{c}{ \footnotesize LPIPS$\downarrow$}
& \multicolumn{1}{c}{ \footnotesize Acc$\uparrow$}
& \multicolumn{1}{c}{ \footnotesize LPIPS$\downarrow$}
& \multicolumn{1}{c|}{ \footnotesize Acc$\uparrow$}
& \multicolumn{1}{c}{ \footnotesize LPIPS$\downarrow$}
& \multicolumn{1}{c}{ \footnotesize Acc$\uparrow$}
\\
\hline

  HyperNeRF
  &$.634$
  &$.107$

  &$.432$
  &$.176$
  
  &$.521$
  &$.316$

  &$.438$
  &$.314$  
  
  &$.641$
  &$.277$

  &$.397$
  &$.252$

  &$.592$
  &$.213$
  
  &$.632$
  &$.053$

  &$.585$
  &$.067$
  
  &$.487$
  &$.072$
  
  &$.462$
  &$.162$
  
  &$.531$
  &$.198$
  
  \\   D$^{2}$NeRF
  &$.540$
  &$.219$

  &$.546$
  &$.220$
  
  &$.687$
  &$.346$

  &$.588$
  &$.403$
  
  &$.556$
  &$.333$

  &$.595$
  &$.339$

  &$.759$
  &$.231$
  
  &$.588$
  &$.066$

  &$.630$
  &$.128$
  
  &$.576$
  &$.078$
  
  &$.628$
  &$.126$
  
  &$.611$
  &$.247$

  \\ \hline  HyperNeRF (+depth)
  & $.373$ & $.352$
  & $.425$ & $.357$
  & $.532$ & $.552$
  & $.371$ & $.596$
  & $.330$ & $.605$
  & $.376$ & $.612$
  & $.514$ & $.451$
  & $.501$ & $.211$
  & $.445$ & $.249$
  & $.450$ & $.283$
  & $.456$ & $.214$
  & $.428$ & $.439$
  
  \\  D$^{2}$NeRF (+depth)
  &$.507$
  &$.338$

  &$.532$
  &$.270$  
  
  &$.685$
  &$.510$

  &$.580$
  &$.362$
  
  &$.561$
  &$.438$

  &$.553$
  &$.376$

  &$.730$
  &$.243$

  &$.585$
  &$.086$

  &$.609$
  &$.131$
  
  &$.608$
  &$.154$
  
  &$.645$
  &$.176$
  
  &$.599$
  &$.302$
  
  \\ \hline   \textbf{Total-Recon}
  &$\textbf{.271}$
  &$\textbf{.841}$

  &$\textbf{.313}$
  &$\textbf{.790}$
  
  &$\textbf{.382}$
  &$\textbf{.889}$

  &$\textbf{.333}$
  &$\textbf{.894}$
  
  &$\textbf{.237}$
  &$\textbf{.967}$

  &$\textbf{.281}$
  &$\textbf{.925}$

  &$\textbf{.261}$
  &$\textbf{.949}$
  
  &$\textbf{.213}$
  &$\textbf{.909}$

  &$\textbf{.264}$
  &$\textbf{.849}$
  
  &$\textbf{.256}$
  &$\textbf{.827}$
  
  &$\textbf{.233}$
  &$\textbf{.914}$
  
  &$\textbf{.278}$
  &$\textbf{.895}$
  \\ \bottomrule

\end{tabular}
}

%% file: tables/ablation_avg_wo_freezecam_col_with_acc.tex
\small{
\begin{tabular}{l||cccc|cc}
    \toprule
    \multicolumn{1}{c||}{Methods}
    & \multicolumn{1}{c}{ \footnotesize \makecell{Depth\\Loss}}
    & \multicolumn{1}{c}{ \footnotesize \makecell{Deform.\\Obj.}}
    & \multicolumn{1}{c}{ \footnotesize \makecell{Root\\Init.}}
    & \multicolumn{1}{c|}{ \footnotesize \makecell{Root\\Motion}}
    & \multicolumn{1}{c}{ \footnotesize LPIPS$\downarrow$ }
    & \multicolumn{1}{c}{ \footnotesize Acc@0.1m$\uparrow$ }\\

    \specialrule{0.1pt}{2pt}{2pt} 
    (1) \textbf{Full model} 
    & \makecell{\textcolor{MyGreen}{\checkmark}}
    & \makecell{\textcolor{MyGreen}{\checkmark}}
    & \makecell{\textcolor{MyGreen}{\checkmark}}
    & \makecell{\textcolor{MyGreen}{\checkmark}}
    &$\textbf{.268}$
    &$\textbf{.898}$ \\
    
    (2) w/o loss $\mathcal{L}_\textrm{depth}$
    & \makecell{\xmark}
    & \makecell{\textcolor{MyGreen}{\checkmark}}
    & \makecell{\textcolor{MyGreen}{\checkmark}}
    & \makecell{\textcolor{MyGreen}{\checkmark}}
    &$.372$
    &$.154$\\

    (3) w/o deform. $\mathbf{J}_j$
    & \makecell{\textcolor{MyGreen}{\checkmark}}
    & \makecell{\xmark}
    & \makecell{\textcolor{MyGreen}{\checkmark}}
    & \makecell{\textcolor{MyGreen}{\checkmark}}
    &$.296$
    &$.867$ \\

    (4) w/o root-body init.
    & \makecell{\textcolor{MyGreen}{\checkmark}}
    & \makecell{\textcolor{MyGreen}{\checkmark}}
    & \makecell{\xmark}
    & \makecell{\textcolor{MyGreen}{\checkmark}}
    &$.293$
    &$.870$ \\
    
    (5) w/o root-body $\mathbf{G}_j$
    & \makecell{\textcolor{MyGreen}{\checkmark}}
    & \makecell{\textcolor{MyGreen}{\checkmark}}
    & \makecell{\xmark}
    & \makecell{\xmark}
    &\small{N/A}
    &\small{N/A}\\

    \bottomrule
    
\end{tabular}
}

%% file: 5_conclusion.tex
\begin{figure}[t]
    \captionsetup[subfigure]{labelformat=empty}
    \centering
    \subfloat{{\includegraphics[width=\linewidth, clip=true,trim=0mm 0mm 0mm 0mm]{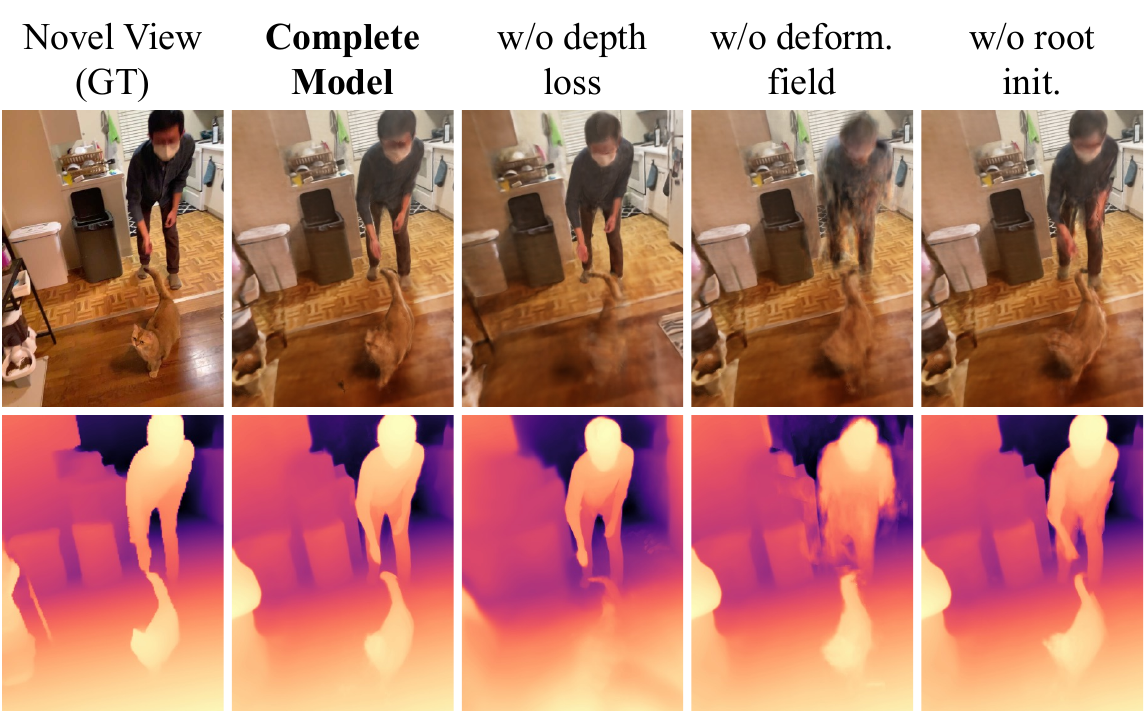}}}\\%
    \caption{\textbf{Ablation Study.} We visualize the ablations from Table \ref{table:ablation_avg}. Removing the depth loss causes the cat to sink into the ground due to inconsistent object scales. Removing the deformation field produces coarse object reconstructions that fail to capture articulated body parts such as moving limbs. Removing PoseNet-initialization of root-body poses results in noisier appearance and geometry, as shown by the human actor's left hand. We omit the ablation without root-body poses as it does not converge. We present additional visualizations in Appendix \ref{sec:ablation_objmotion_supp}. \href{https://andrewsonga.github.io/totalrecon/ablation_objmotion.html}{\textbf{[Videos]}} \captionspace\vspace{-1 em}
    }%
    \label{fig:ablation_humancat}
\end{figure}

\section{Discussion and Limitations}
We have presented a new system for automated embodied view synthesis from monocular RGBD videos, focusing on videos of people interacting with their pets. Our main technical contribution is \ourmethod{}, a 3D representation for deformable scenes that hierarchically decomposes scene motion into the motion of each object, which is further decomposed into its root-body motion and local articulations; this key design choice enables appropriate initialization of root-body poses and hence easier optimization over long videos containing large motions. %
By explicitly reconstructing the geometry, appearance, root-body- and articulated motion of each object, \ourmethod{} enables seeing through the eyes of people and pets and generating game-like traversals of deformable scenes from behind a target actor. 

\paragraph{Limitations.}
In \ourmethod{}, scene decomposition is primarily supervised by object silhouettes computed by an off-the-shelf segmentation model \cite{kirillov2020pointrend}, which may be inaccurate, especially in partial occlusion scenarios. This may damage the resulting reconstructions and embodied-view renderings. We believe that incorporating the latest advances in video instance segmentation \cite{huang2022minvis} will enable \ourmethod{} to be applied to more challenging scenarios. Second, \ourmethod{} initializes the root-body pose of each deformable object using a PoseNet \cite{yang2022banmo} trained for humans and quadruped animals, which does not generalize to other object categories (e.g., birds, fish). We reserve the reconstruction of generic scenes for future work. Finally, our model needs to be optimized on a per-sequence basis for roughly 15 hours with 4 NVIDIA RTX A5000 GPUs and is therefore not suitable for real-time applications. Incorporating recent advances in fast neural field training methods is an interesting avenue for future research.

\paragraph{Acknowledgments.} We thank Nathaniel Chodosh, Jeff Tan, George Cazenavette, and Jason Zhang for proofreading and Songwei Ge for reviewing the codebase. We also thank Sheng-Yu Wang, Daohan (Fred) Lu, Tamaki Kojima, Krishna Wadhwani, Takuya Narihira, and Tatsuo Fujiwara for providing valuable feedback. This work is supported in part by the Sony Corporation, Cisco Research, and the CMU Argo AI Center for Autonomous Vehicle Research.

%% file: Appendix.tex
\clearpage
\appendix

\section*{\Large Appendix}

The appendix is comprised of the following: additional details of \ourmethod{}'s implementation (Section~\ref{sec:implementation details_supp}), our dataset (Section~\ref{sec:dataset_details}), and the baselines (Section~ \ref{sec:baseline_setup}), additional metrics and results for the baseline comparisons (Tables \ref{table:nvs_comparisons_entireimg_visual} and \ref{table:nvs_comparisons_entireimg_depth}, Figure \ref{fig:nvs_comparisons_moreseqs}), 
reconstructions and embodied view synthesis results on additional sequences and object removal results (Section~\ref{sec:reconstruction and applications_supp}), additional ablation studies (Section~\ref{sec:ablation_supp}), and a societal impact statement (Section~\ref{sec:societal_impact}).

\section{Implementation Details}
\label{sec:implementation details_supp}

\paragraph{Data Preprocessing.} Before training our composite scene representation, we follow BANMo \cite{yang2022banmo} by resizing the raw RGB images and the ground truth depth maps from 960 $\times$ 720 and 256 $\times$ 192 resolution, respectively, to a resolution of 512 $\times$ 512, which is the resolution used during training. We also scale the ground-truth depth measurements and the translation component of the ARKit camera poses (used to initialize the background's root-body poses $\mathbf{G}^{t}_{0}$) by a scaling factor of 0.2, which we empirically found to improve the pre-training of the deformable objects. After training, we scale the ground-truth and rendered depth back to the original metric space and compute the evaluation metrics at 480 $\times$ 360 resolution.

\paragraph{Optimization.} We optimize our composite scene representation by first pre-training each object field separately and then jointly finetuning them. We use the same batch size, sampled rays per batch, and sampled points per ray as BANMo \cite{yang2022banmo} for both the pre-training and joint-finetuning stages. pre-training a deformable object takes 8.5 hours with 4 NVIDIA RTX A5000 GPUs, and pre-training the background takes 4.5 hours. Jointly finetuning one deformable object and the background takes an additional 1.5 hours with 4 NVIDIA RTX A5000 GPUs, and jointly finetuning two deformable objects and the background takes an additional 2.5 hours with 4 NVIDIA RTX A6000 GPUs.

\paragraph{Pre-training.}
For pre-training deformable objects, we follow the training procedure of and use the same hyper-parameters as BANMo \cite{yang2022banmo}, which we augment with a depth reconstruction loss weighted by a default value of $\lambda_\textrm{depth} = 5$ (for the \textsc{Human 1} sequence, we use a loss weight of $\lambda_\textrm{depth} = 1.5$ for pre-training the deformable object). Following BANMo, we pre-train each deformable object in three training stages, each for 24k, 6k, and 24k iterations.

For pre-training the background model, we optimize color, flow, and depth reconstruction losses $\mathcal{L}_\textrm{rgb}$, $\mathcal{L}_\textrm{flow}$, $\mathcal{L}_\textrm{depth}$ on pixels outside the ground-truth object silhouettes, each with a default weight of $\lambda_\textrm{rgb} = 0.1$, $\lambda_\textrm{flow} = 1$, $\lambda_\textrm{depth} = 1$, respectively. We also optimize an eikonal loss $\mathcal{L}_\textrm{SDF}$ \cite{yariv2021volume} with a weight of $\lambda_\textrm{SDF} = 0.001$ to encourage the reconstruction of a valid signed distance function (SDF):

\begin{align}
    \label{eq:eikonal_loss}
    \mathcal{L}_\textrm{SDF} = \sum_{\mathbf{x}^{t}}\sum_{\mathbf{X}_{i}^{t}}(||\nabla_{\mathbf{X}_{i}^{*}}\mathbf{MLP}_\textrm{SDF}(\mathbf{X}_{i}^{*})||_{2} - 1)^{2},
\end{align}

\noindent where $\mathbf{x}^{t} \in \mathbb{R}^{2}$ denotes the pixel location at time $t$, $\mathbf{X}_{i}^{*} \in \mathbb{R}^{3}$ is the 3D point in the canonical world space corresponding to $\mathbf{X}^{t}_{i} \in \mathbb{R}^{3}$, the $i^{th}$ sample in the camera space. To compute this eikonal loss, we sample 17 uniformly spaced points $\mathbf{X}^{t}_{i}$ along each camera ray $\mathbf{v}^{t}$ from a truncated region that is 0.2m long and centered at the surface point computed by backprojecting the ground-truth depth.

We pre-train the background model in two stages: in the first stage, we optimize the color, flow, depth, and eikonal losses with their respective default loss weights for 24k iterations. In the second stage, we optimize the same set of losses for another 24k iterations while fixing the background model's root-body poses $\mathbf{G}^{t}_{0}$, increasing the weight of the color loss from $\lambda_\textrm{rgb} = 0.1$ to $\lambda_\textrm{rgb} = 1$, and performing active sampling of pixels $\mathbf{x}^{t}$ to improve the background model's appearance, as was done in \cite{yang2022banmo}.

\paragraph{Joint Finetuning.}
For the joint-finetuning of all of the object models, we optimize the color, flow, depth, and per-object 3D-cycle consistency losses for another 6k iterations, each with a default weight of $\lambda_\textrm{rgb} = 1$, $\lambda_\textrm{flow} = 1$, $\lambda_\textrm{depth} = 5$, and $\lambda_{\textrm{cyc, }j} = 1$, respectively. Importantly, we freeze the background's appearance and shape models by default and only allow its root-body poses $\mathbf{G}^{t}_{0}$, the foregrounds' root-body poses $\mathbf{G}^{t}_{j}$, and the foregrounds' appearance and shape models to be optimized (for the \textsc{Human 1} sequence, we use a loss weight of $\lambda_\textrm{depth} = 1.5$, and for the \textsc{Cat 1} and \textsc{Cat 1 (v2)} sequences, we allow the background's appearance and shape models to be optimized during joint-finetuning). We also perform active sampling of pixels $\mathbf{x}^{t}$ over all deformable foreground objects. Intuitively, the joint-finetuning stage improves the appearance of the foreground objects and helps the model learn correct object-to-object interactions.

\begin{figure}[t]
    \captionsetup[subfigure]{labelformat=empty}
    \centering
    \subfloat{{\includegraphics[width=\textwidth]{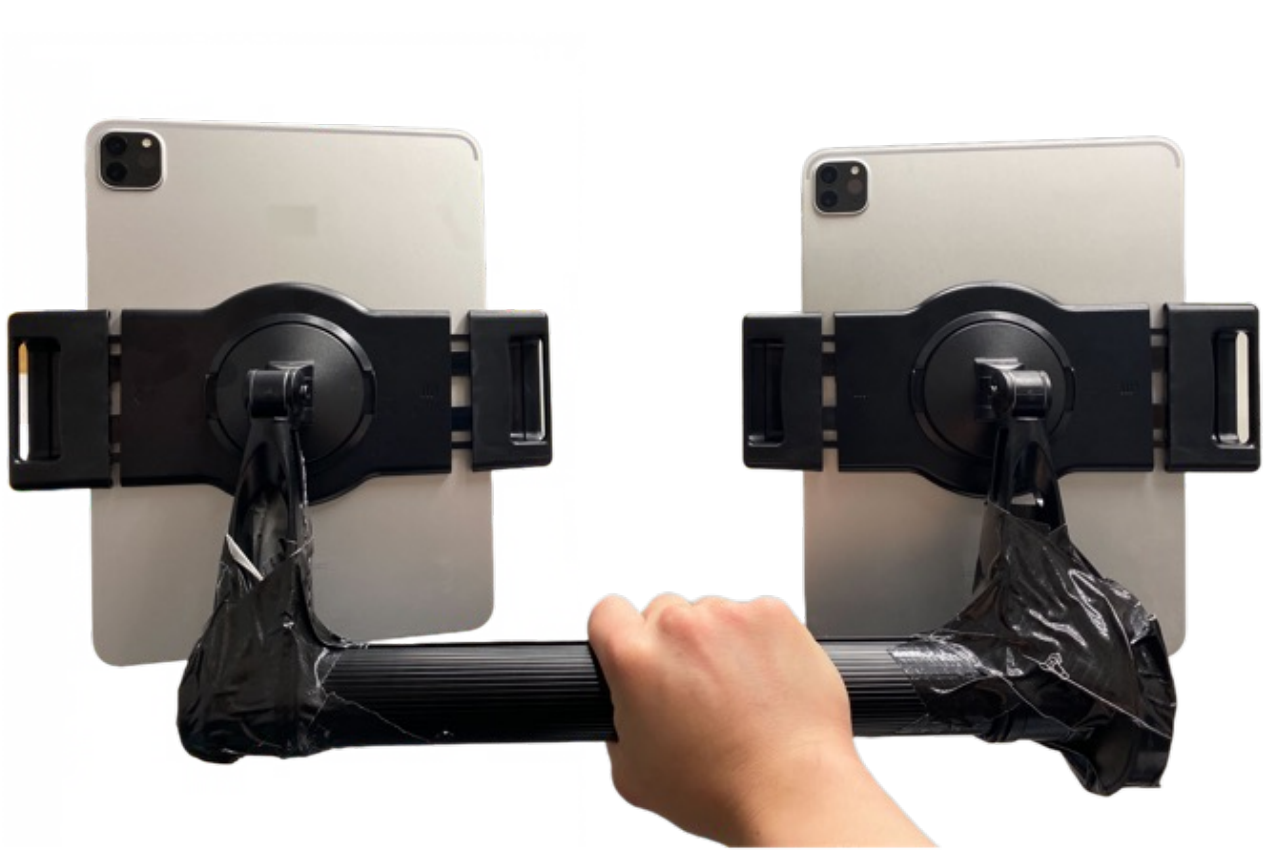}}}\\
    \caption{\textbf{Stereo Validation Rig Used Only for Evaluation}. To enable quantitative evaluation, we built a stereo rig comprised of two iPad Pros rigidly attached to a camera mount and captured 11 pairs of RGBD sequences. \textbf{We train each method \textit{only} on the sequences captured from the left camera} and evaluate the images rendered from the viewpoint of the right camera.\captionspace}%
    \label{fig:stereo_rig}
\end{figure}

\section{Dataset Details}
\label{sec:dataset_details}

In this section, we describe the stereo validation rig we built for evaluation and elaborate on how the validation rig is used to evaluate novel-view synthesis.

As shown in Figure \ref{fig:stereo_rig}, our stereo validation rig is comprised of two iPad Pro's rigidly attached to a camera mount. Importantly, \textbf{we train each method \textit{only} on the sequences captured from the left camera} and evaluate the images rendered from the viewpoint of the right camera \textit{i.e.,} the ``novel-view''. To compute the pose of the ``novel-view'' camera, we compute the rigid transform between the left and right cameras and use this transform to map the optimized training-view cameras of our method to the novel-view cameras. For each sequence, we register the two cameras by solving a Perspective-n-Point (PnP) problem using manually annotated 2D-2D correspondences.

The PnP problem aims to estimate the pose of a \textit{calibrated camera} given $n$ 3D-2D correspondences \textit{i.e.,} a set of $n$ 3D points defined in some \textit{world frame} and their corresponding 2D image projections. We formulate the problem of estimating the left-to-right camera transform as a PnP problem where the left camera of our validation rig corresponds to the \textit{world frame}, and the right camera of our validation rig corresponds to the \textit{calibrated camera}.

To obtain 3D-2D correspondences, we first manually annotate at least 20 2D-2D correspondences for each sequence. Next, we obtain the 3D points defined in the frame of the left camera by backprojecting its ground-truth depth using the provided intrinsics. Finally, we feed 1) the 3D points in the left-camera frame, 2) the 2D annotations, and 3) the intrinsics of the right camera to a generic PnP solver to compute the desired left-to-right camera transform.

Using the stereo validation rig, we captured a dataset containing 11 pairs of RGBD sequences featuring 3 different cats, 1 dog, and 2 human subjects in 4 different indoor environments. For each sequence, we provide 1) the RGBD frames (and object masks) captured from both cameras of our validation rig, 2) their camera pose trajectories, 3) their camera intrinsics, and 4) the left-to-right camera transform.

\section{Details of Baseline Comparisons}
\label{sec:baseline_setup}

\begin{table*}[ht]
\caption{\textbf{Quantitative Comparisons on Novel View Synthesis (Visual Metrics)}. We compare our method to HyperNeRF \cite{park2021hypernerf}, D$^2$NeRF \cite{d2nerf}, and their depth-supervised variants on the 11 sequences of our stereo RGBD dataset, in terms of LPIPS, PSNR, and SSIM. Our method significantly outperforms all baselines for all sequences.
\captionspace
}
\label{table:nvs_comparisons_entireimg_visual}
\centering
\resizebox{\linewidth}{!}{
\setlength{\tabcolsep}{2pt}
\input{tables/comparisons_entireimg_visual_allseqs_part1.tex}
}\\
\resizebox{\linewidth}{!}{
\setlength{\tabcolsep}{2pt}
\input{tables/comparisons_entireimg_visual_allseqs_part2.tex}
}
\vspace{-0.40em}
\end{table*}

\begin{table*}[ht]
\caption{\textbf{Quantitative Comparisons on Novel View Synthesis (Depth Metrics)}. We compare our method to HyperNeRF \cite{park2021hypernerf}, D$^2$NeRF \cite{d2nerf}, and their depth-supervised variants on the 11 sequences of our stereo RGBD dataset, in terms of the average accuracy at 0.1m (Acc@0.1m) and the RMS depth error $\epsilon_\textrm{depth}$ (units: meters). Our method significantly outperforms all baselines for all sequences. \captionspace}
    \label{table:nvs_comparisons_entireimg_depth}
\centering
\resizebox{\linewidth}{!}{
\setlength{\tabcolsep}{2pt}
\input{tables/comparisons_entireimg_depth_allseqs_part1.tex}
}\\
\resizebox{\linewidth}{!}{
\setlength{\tabcolsep}{2pt}
\input{tables/comparisons_entireimg_depth_allseqs_part2.tex}
}
\vspace{-0.40em}
\end{table*}

\paragraph{Baseline Experiment Details.} We provide additional details of the experiment settings of the baselines. For the sake of fair comparison, we set up augmented versions of the baselines D$^2$NeRF \cite{d2nerf} and HyperNeRF \cite{park2021hypernerf}, whereby we replace their COLMAP \cite{Schonberger_2016_colmap} camera poses with the iPad Pro's camera poses provided by ARKit - the same camera poses used to initialize the root-body transforms $\mathbf{G}^{t}_{0}$ of our method's background model. We also compare our method to depth-supervised variants of HyperNeRF and D$^2$NeRF, which uses the same losses and hyperparameters as the raw baselines, with the exception of an additional depth loss with weight $\lambda_\textrm{depth} = 0.1$. We empirically observe that using a higher depth-loss weight significantly deteriorates the baseline methods' rendered appearance. As was done for our method, before training, we scale the ground-truth depth measurements and the translation component of the ARKit camera poses by a scaling factor of 0.2; after training, we scale both the ground-truth and rendered depth back to the original metric space and compute the evaluation metrics at 480 $\times$ 360 resolution.

\paragraph{Additional Qualitative Results.} In Figure \ref{fig:nvs_comparisons_moreseqs}, we show qualitative comparisons on the remaining 10 sequences of our RGBD dataset that were not shown in the main paper, namely sequences \textsc{human 2 \& cat1}, \textsc{human 2}, \textsc{dog 1} (v1), \textsc{dog 1} (v2), \textsc{cat 1} (v1), \textsc{cat 1} (v2), \textsc{cat 2} (v1), \textsc{cat 2} (v2), \textsc{cat 3}, \textsc{human 1}. Due to space limits, we only display the visualizations for the depth-supervised variants of the baselines, but we showcase the complete set of baselines on our \href{https://andrewsonga.github.io/totalrecon/nvs.html}{project page}. \ourmethod{} outperforms all of the baselines, which can only reconstruct the rigid background. On the other hand, \ourmethod{} reconstructs the \textit{entire} scene, including all dynamic objects.

\paragraph{Additional Quantitative Metrics.}
In Tables \ref{table:nvs_comparisons_entireimg_visual} and \ref{table:nvs_comparisons_entireimg_depth}, we display the full set of quantitative metrics for our method and all of the baselines. In addition to the LPIPS and the average depth accuracy at 0.1m reported in the main paper, we also report the PSNR, SSIM, and RMS depth error. Our method significantly outperforms all of the baselines in terms of LPIPS, PSNR, SSIM, the average depth accuracy at 0.1m, and the RMS depth error for all sequences.

\begin{figure}[ht]
    \vspace{-5pt}
    \captionsetup[subfigure]{labelformat=empty}
    \centering
    \subfloat{{\includegraphics[width=\linewidth]{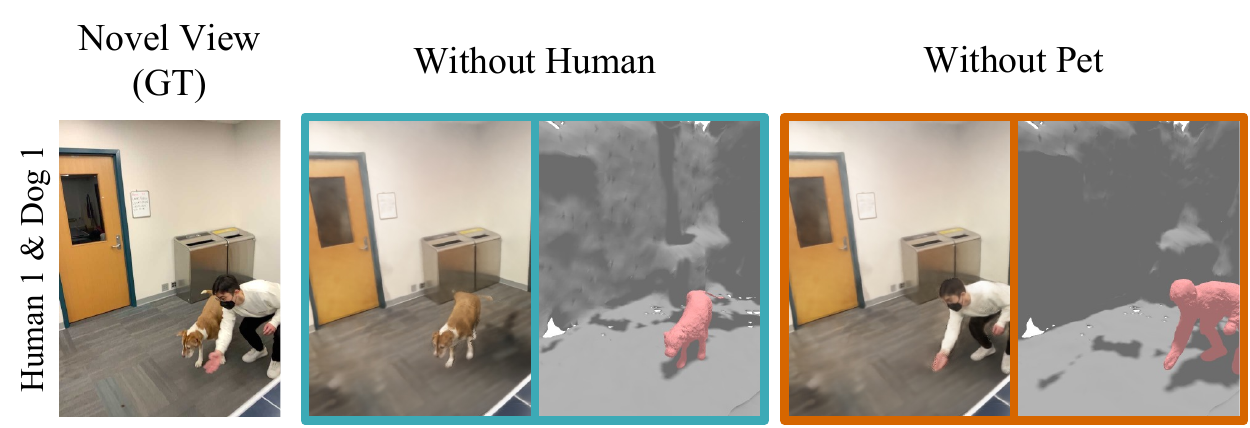}}}\\%
    \subfloat{{\includegraphics[width=\linewidth]{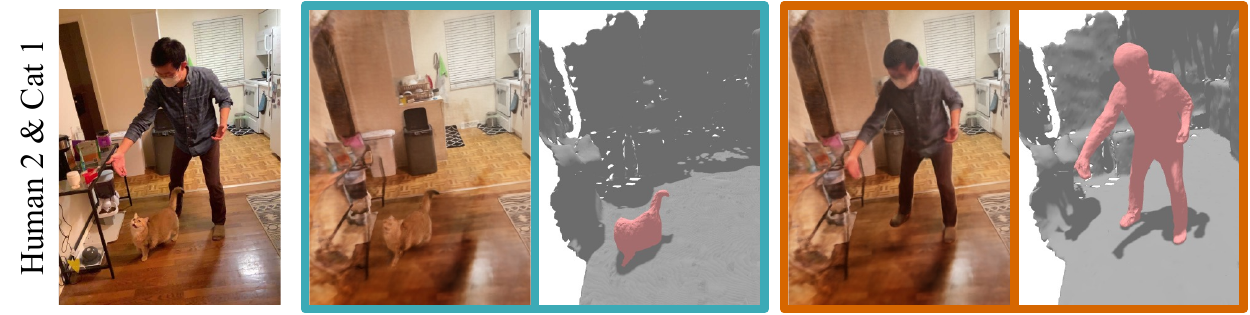}}}%
    \caption{\textbf{Object Removal.} Our compositional scene representation enables object removal. We remove the \textsc{human} and then the \textsc{pet} object from our composite rendering process and display the resulting renderings. \href{https://andrewsonga.github.io/totalrecon/\#object_removal}{\textbf{[Videos]}}\captionspace}%
    \label{fig:object_removal}
\end{figure}

\section{Reconstruction and Applications}
\label{sec:reconstruction and applications_supp}

\paragraph{Geometry and Embodied View Synthesis.}

\begin{figure*}[htbp]
    \captionsetup[subfigure]{labelformat=empty}
    \centering
    \subfloat{{\includegraphics[width=0.95\textwidth]
    {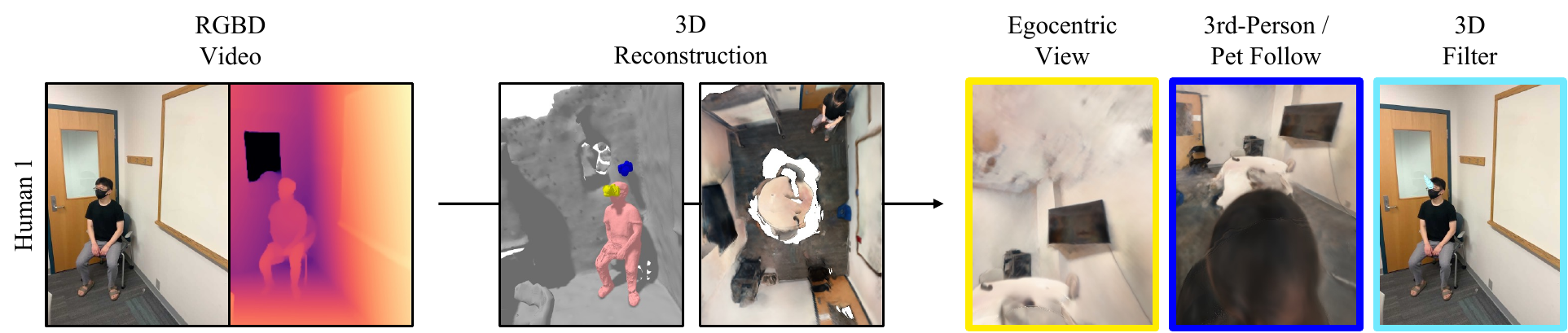}}}\\%
    \subfloat{{\includegraphics[width=0.95\textwidth]{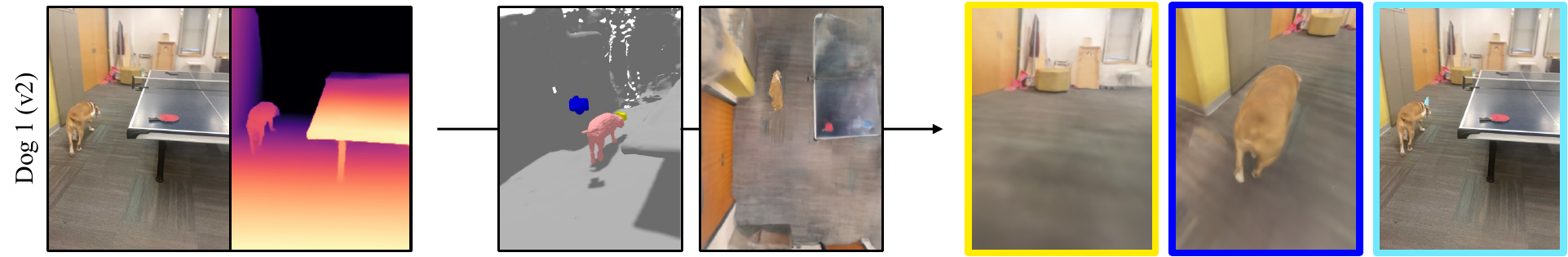}}}\\%
    \subfloat{{\includegraphics[width=0.95\textwidth]{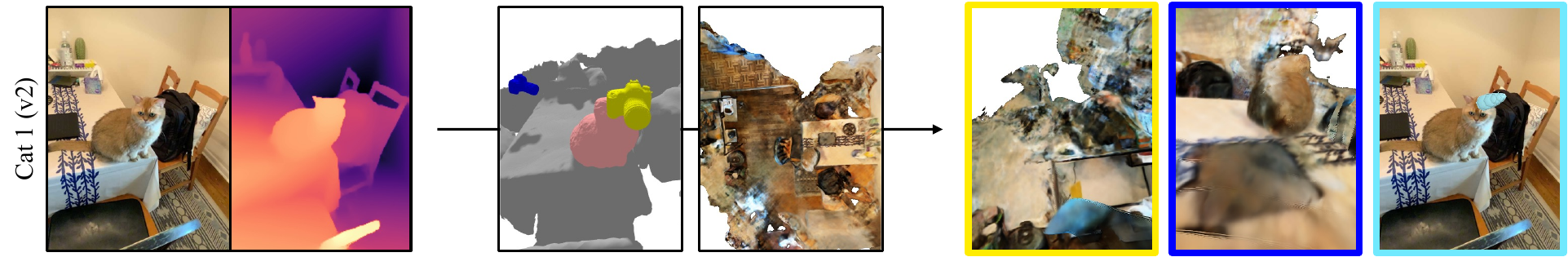}}}\\%
    \subfloat{{\includegraphics[width=0.95\textwidth]{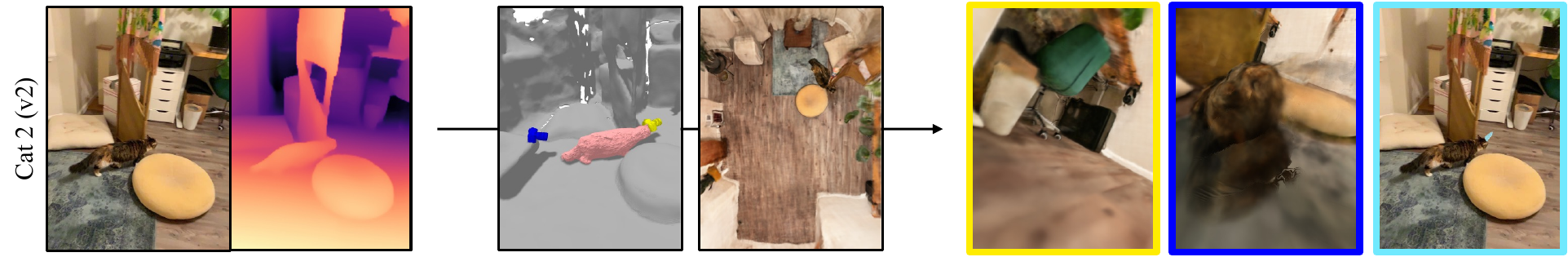}}}\\%
    \subfloat{{\includegraphics[width=0.95\textwidth]{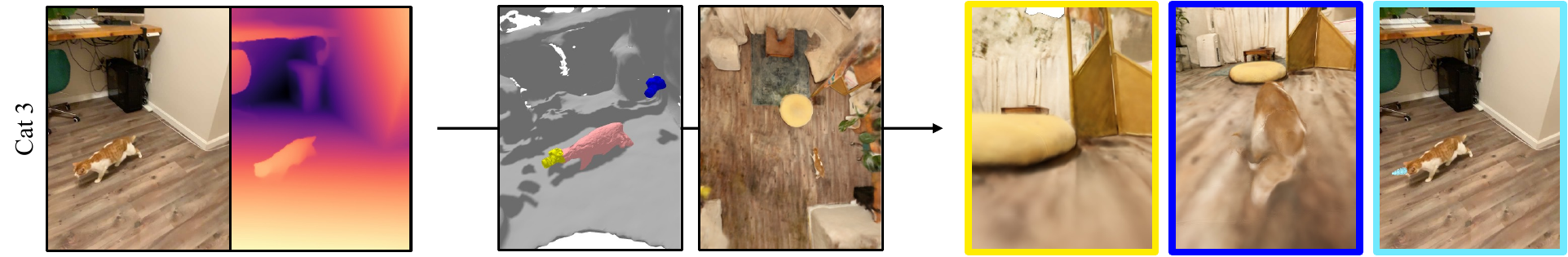}}}\\%
    \caption{\textbf{Embodied View Synthesis and 3D Filters (Additional Sequences).} We visualize the 3D reconstructions (rendered from the novel view) and the applications enabled by \ourmethod{} for the remaining 5 sequences of our RGBD dataset that were not shown in the main paper. The yellow and blue camera meshes in the mesh renderings represent the egocentric and 3rd-person-follow cameras, respectively. To showcase the 3D filter, we attach a sky-blue unicorn horn to the forehead of the foreground object, which is automatically propagated across all frames. \href{https://andrewsonga.github.io/totalrecon/applications.html}{\textbf{[Videos]}}\captionspace}%
    \label{fig:extremeview_moreseqs}
\end{figure*}

In Figure \ref{fig:extremeview_moreseqs}, we display the novel-view reconstructions and the corresponding embodied view synthesis results for the remaining 5 sequences of our RGBD dataset that were not shown in the main paper: sequences \textsc{human 1}, \textsc{dog 1} (v2), \textsc{cat 1} (v2), \textsc{cat 2} (v2), \textsc{cat 3}. 

\begin{table*}[ht]
\caption{\textbf{Ablation Study on Depth Supervision.} Depth supervision improves our model both in terms of the visual (LPIPS) and depth (Acc@0.1m) metrics, an observation that is consistent with the qualitative results displayed in Figure \ref{fig:ablation_depth_full}. \captionspace}
    \label{table:ablation_depth}
\centering
\resizebox{\linewidth}{!}{
\setlength{\tabcolsep}{2pt}
\input{tables/ablation_depth_with_acc}

}
\end{table*}

\begin{table*}[ht]
\caption{\textbf{Ablation Study on Motion Modeling.} Ablating camera-pose optimization (row 2), changing the deformation field (row 3), removing deformation modeling (row 4), or removing PoseNet initialization of object root-body poses (row 5) moderately hurts the visual and depth metrics. Importantly, removing root-body modeling entirely (row 6) prevents our method from converging (N/A), as the deformation field alone has to explain global object motion (see Figure \ref{fig:method}). We perform another ablation (row 7) that replaces \ourmethod{}'s neural blend skinning (NBS) function with the more flexible SE(3)-field \cite{park2021nerfies}, which does converge but still performs worse than other converging ablations. These experiments justify \ourmethod{}'s hierarchical motion representation, which decomposes object motion into global root-body motion and local articulations. $^\dagger$When ablating root-body poses, we freeze the camera poses to prevent the object fields, which are now all defined in a shared world space, from learning different camera poses during their separate pre-training processes.\captionspace}
\label{table:ablation_objmotion_extended_perseq}
\centering
\resizebox{\linewidth}{!}{
\setlength{\tabcolsep}{2pt}
\input{tables/ablation_objmotion_extended_perseq_with_acc}
}
\end{table*}

\paragraph{Object Removal.} Our compositional scene representation allows for easy object removal. To remove object $k$ from our trained scene representation, one skips object $j = k$ in the summation operation of  Equation \ref{eq:composite}. We showcase object removal in Figure \ref{fig:object_removal}. %

\section{Additional Ablation Studies}
\label{sec:ablation_supp}

\subsection{Ablation Study on Depth Supervision}
\label{sec:ablation_depth_supp}

\begin{figure*}[!t]
    \captionsetup[subfigure]{labelformat=empty}
    \caption{\textbf{Ablation Study on Depth Supervision.} While removing depth supervision from \ourmethod{} (\textsc{Complete Model}) doesn't significantly hamper the training-view RGB renderings, it induces the following failure modes in the \textit{novel-view} 3D reconstructions. (a) \textit{Floating objects}: for the \textsc{Human 1 \& Dog 1}, \textsc{Dog 1}, \textsc{Human 1}, and \textsc{Cat 2} sequences, the foreground objects float above the ground, as evidenced by their shadows. (b) \textit{Objects that sink into the background}: for the \textsc{Human 2 \& Cat 1} sequence, the reconstructed cat is halfway sunk into the ground. (c) \textit{Incorrect occlusions}: for the \textsc{Human 1 \& Dog 1} sequence, the human is incorrectly occluding the dog. (d) \textit{Lower reconstruction quality}: for the \textsc{Human 2 \& Cat 1} sequences, we observe that the cat has lower reconstruction quality, and, for all sequences except \textsc{Human 1 \& Dog 1} and \textsc{Cat 2}, we observe that the background object has lower reconstruction quality. \href{https://andrewsonga.github.io/totalrecon/ablation_depth.html}{\textbf{[Videos]}}\captionspace}%
    \label{fig:ablation_depth_full}
    \centering
    \subfloat{{\includegraphics[width=0.48\textwidth]{figures/ablation_depth_humandog.pdf}}}\quad%
    \subfloat{{\includegraphics[width=0.48\textwidth]{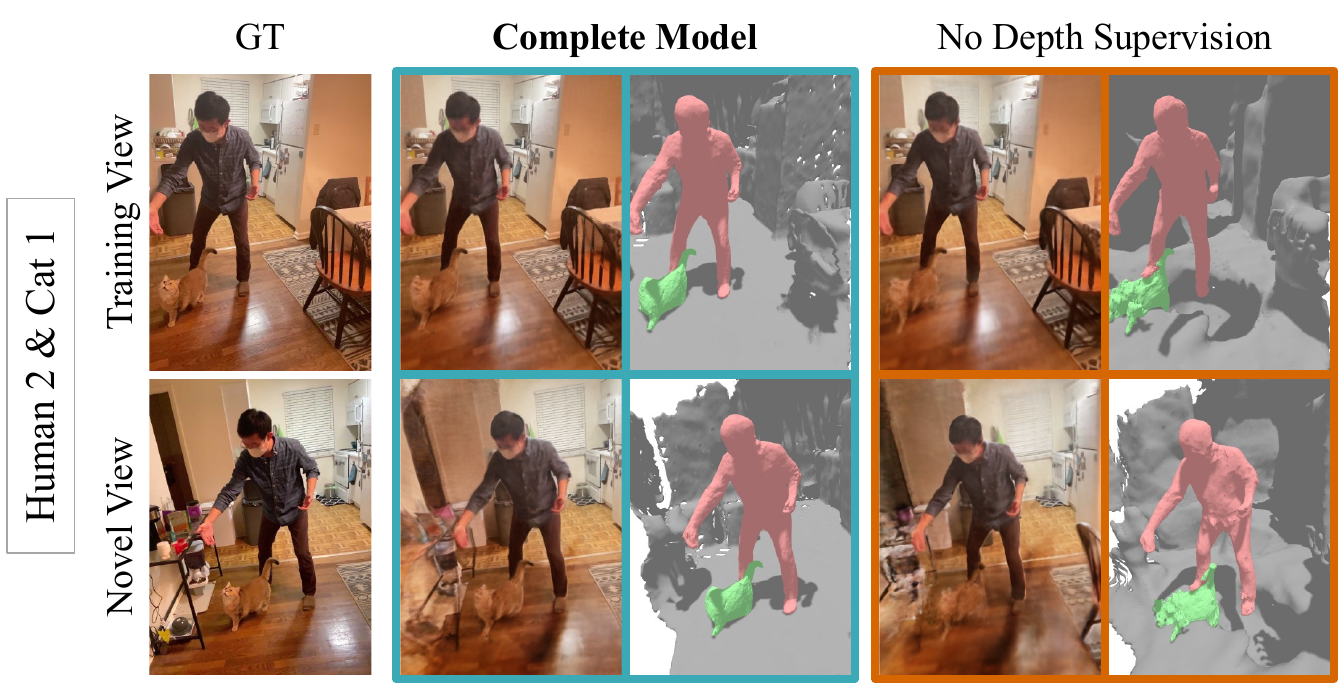}}}\\%
    \subfloat{{\includegraphics[width=0.48\textwidth]{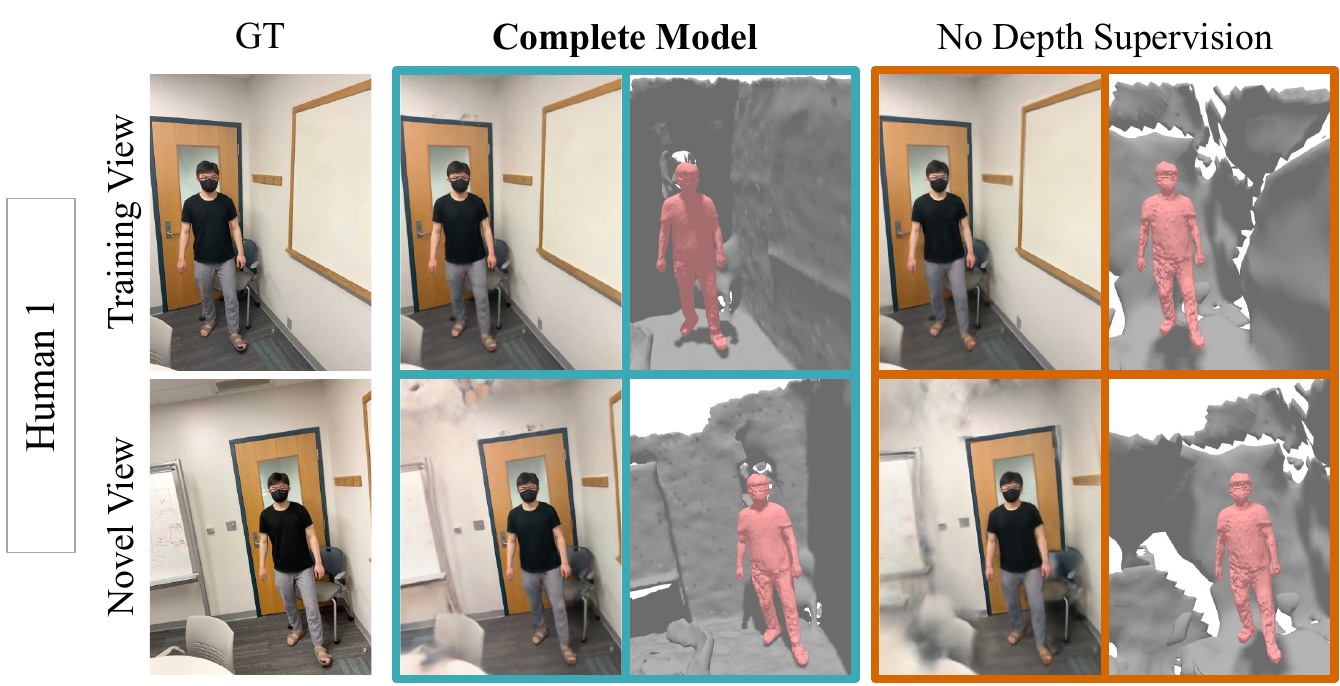}}}\quad%
    \subfloat{{\includegraphics[width=0.48\textwidth]{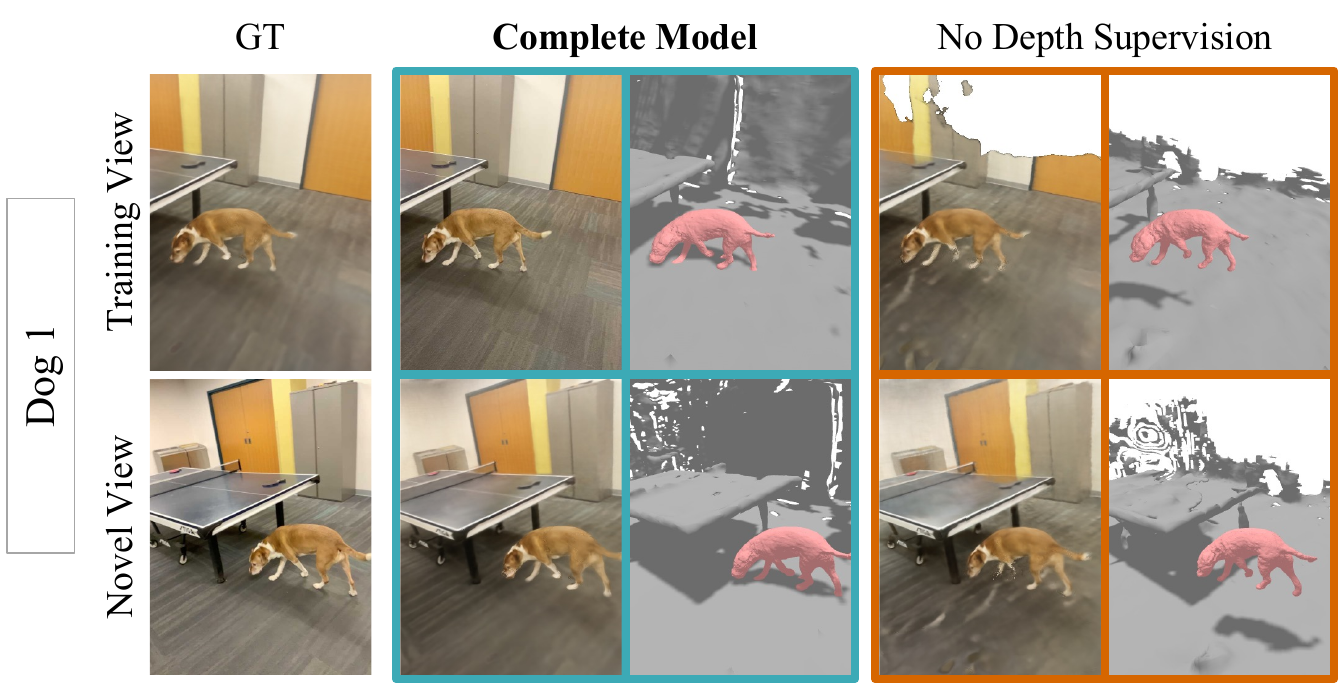}}}\\%
    \subfloat{{\includegraphics[width=0.48\textwidth]{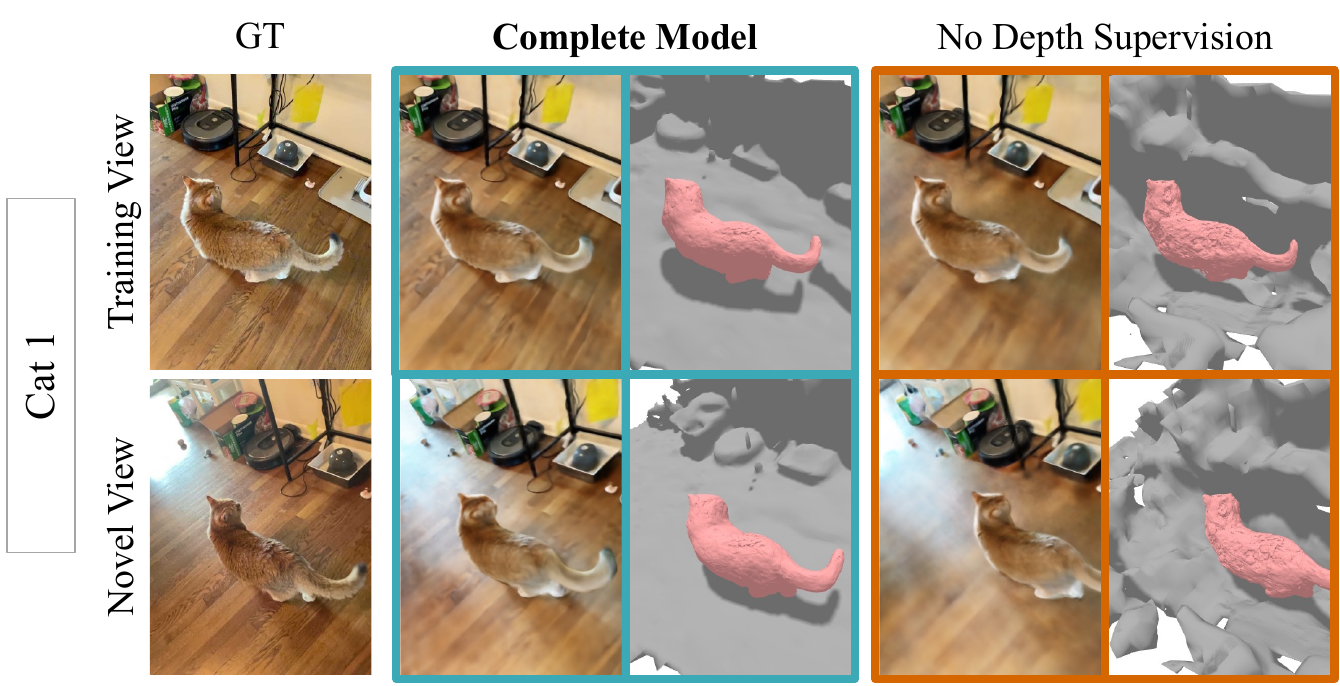}}}\quad%
    \subfloat{{\includegraphics[width=0.48\textwidth]{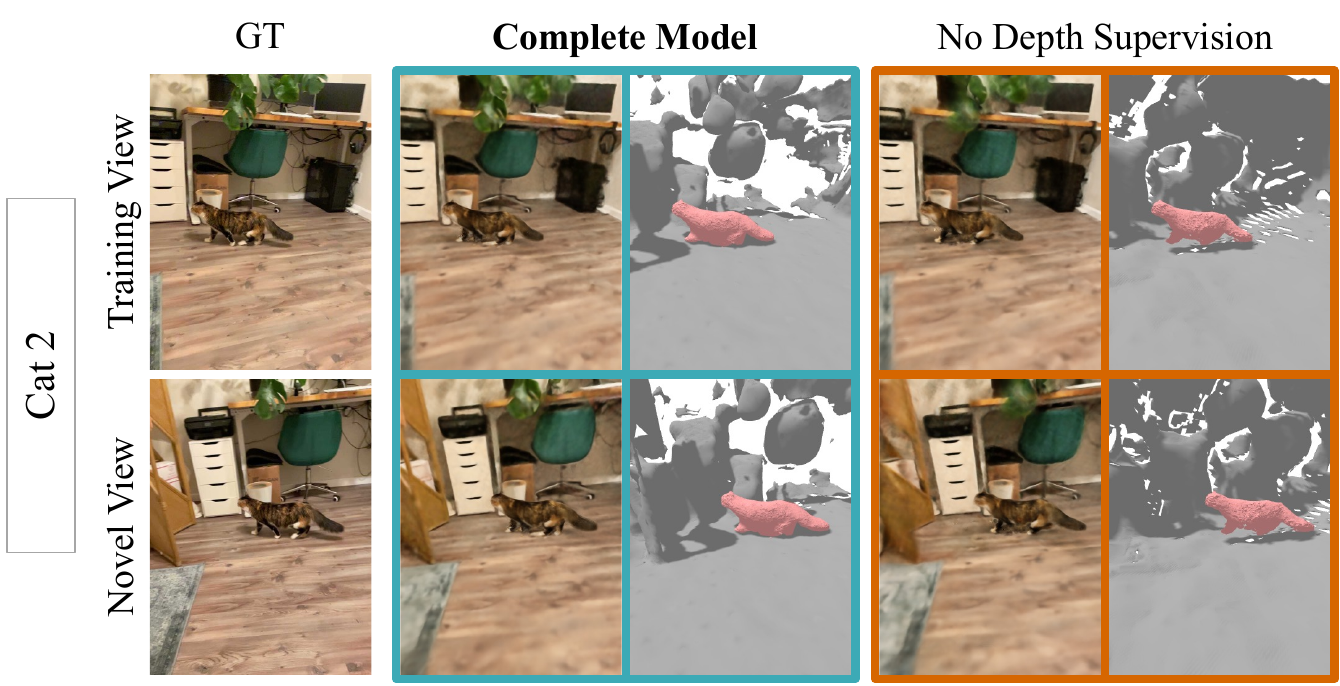}}}\\%
\end{figure*}

In this section, we perform an ablation study on depth supervision for additional sequences in our dataset. Figure \ref{fig:ablation_depth_full} shows that while removing depth supervision from \ourmethod{} does not significantly deteriorate the training-view RGB renderings, it induces critical failure modes as shown in the \textit{novel-view} 3D reconstructions: (a) \textit{Floating objects}: for the \textsc{Human 1 \& Dog 1}, \textsc{Dog 1}, \textsc{Human 1}, and \textsc{Cat 2} sequences, the foreground objects float above the ground, as evidenced by their shadows. (b) \textit{Objects that sink into the background}: for the \textsc{Human 2 \& Cat 1} sequence, the reconstructed cat is halfway sunk into the ground. (c) \textit{Incorrect occlusions}: for the \textsc{Human 1 \& Dog 1} sequence, the human is incorrectly occluding the dog. (d) \textit{Lower reconstruction quality}: for the \textsc{Human 2 \& Cat 1} sequence, the cat displays lower reconstruction quality, and for all sequences except \textsc{Human 1 \& Dog 1} and \textsc{Cat 2}, the background exhibits lower reconstruction quality.

These observations are corroborated by Table \ref{table:ablation_depth}, which shows that depth supervision significantly improves our method's visual and depth metrics over all sequences. Another reason for the large difference in metrics is that the novel-view cameras computed for the non-depth-supervised version may not be entirely accurate. This is because our method optimizes the camera poses during training, meaning that in the absence of the depth loss, the training-view camera poses may converge to a different scale to the ground-truth left-to-right camera transform from Section \ref{sec:dataset_details}, resulting in slightly misaligned novel-view cameras.

\subsection{Ablation Study on Motion Modeling}
\label{sec:ablation_objmotion_supp}
In this section, we perform ablation studies on \ourmethod{}'s motion model for a more comprehensive set of design choices than those presented in the main paper. Table \ref{table:ablation_objmotion_extended_perseq} and Figure \ref{fig:ablation_objmotion_full} show that ablating camera-pose optimization (row 2) worsens the metrics but does not result in qualitative deterioration of the scene reconstruction. This suggests that the ARKit camera poses used to initialize $\mathbf{G}^{t}_{0}$ (Equation \ref{eq:camera}) are already reasonably accurate. Changing the deformation field from \ourmethod{}'s neural blend skinning (NBS) function to the SE(3)-field used in HyperNeRF \cite{park2021hypernerf} (row 3) further worsens the metrics, which are reflected in the minor artifacts that appear in the foreground reconstructions. Removing the deformation field entirely (row 4) also worsens the results, as our method now has to explain each object's (non-rigid) motion solely via its rigid, root-body poses. As a result, this ablation can only recover coarse object reconstructions that fail to model moving body parts such as limbs. Ablating PoseNet-initialization of root-body poses (row 5) is just as detrimental, resulting in noisy appearance and geometry and sometimes even failed object reconstructions (see \textsc{Dog 1} sequence in Figure \ref{fig:ablation_objmotion_full}).

Most notably, Table \ref{table:ablation_objmotion_extended_perseq} shows that removing object root-body poses entirely (row 6) prevents our method from converging, even though the deformation field should be sufficient (in theory) to represent all continuous motion. However, when root-body poses are removed from our method, each object's canonical model is defined in the \textit{static, world} coordinate frame (Equation \ref{eq:obj_motion_modeling_baselines}) as opposed to the moving, object-centric coordinate frame (Equation \ref{eq:backward_warp}). Therefore, the deformation field alone has to explain \textit{global} object motion by learning potentially large deviations from the canonical model, significantly complicating optimization. We posit that \ourmethod{}'s neural blend skinning function is too constraining of a deformation field to model global object motion, so we perform another ablation with the more flexible SE(3)-field (row 7). This ablation does converge but still performs worse than other converging ablations.

These diagnostics justify \ourmethod{}'s hierarchical motion representation, which explicitly models objects' root-body motion; even root-bodies without a deformation field (row 4) or poorly initialized root-bodies (row 5) outperform no root-bodies (row 6). Our ablations also suggest that the poor performance of the baseline methods may be attributed to the lack of object-centric motion modeling, especially since the baseline method \textsc{D$^2$NeRF (w/ depth)} and the ablation \textsc{w/o root-body (SE3)} both exhibit the ghosting artifacts that indicate failed foreground reconstruction (see Figures \ref{fig:nvs_comparisons_moreseqs} and \ref{fig:ablation_objmotion_full}, respectively). Note that these two methods are not strictly equivalent.

\begin{table*}[t]
\caption{\textbf{Ablation Study on Joint-Finetuning.} Joint-finetuning improves LPIPS across most sequences but does not always improve the average depth accuracy at 0.1m. We posit that using a depth signal during pre-training of the individual objects is sufficient for learning a metric model, such that the qualitative improvements induced by joint-finetuning (Figure \ref{fig:ablation_jointfinetuning_supp}) are not always reflected in the metrics. \captionspace}
\label{table:ablation_jointfinetuning_supp}
\centering
\resizebox{\linewidth}{!}{
\setlength{\tabcolsep}{2pt}
\input{tables/ablation_jointfinetuning_entireimg_with_acc.tex}

}
\end{table*}

\begin{figure}[!t]
    \captionsetup[subfigure]{labelformat=empty} 
    \centering
    \subfloat{{\includegraphics[width=\textwidth]{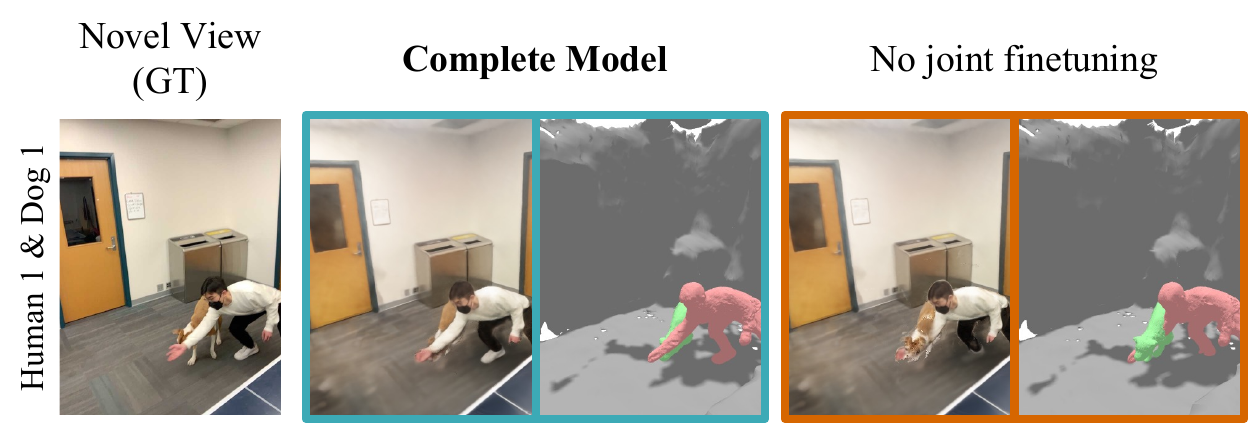}}}\\%
    \subfloat{{\includegraphics[width=\textwidth]{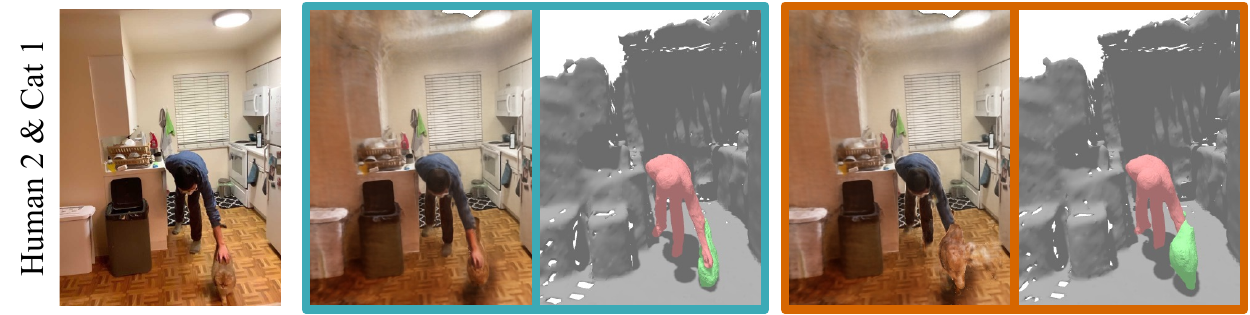}}}%
    \caption{\textbf{Ablation Study on Joint-Finetuning.} 
    Joint-finetuning enables \ourmethod{} to learn the correct human-pet interactions, particularly for frames without any detected object masks.\captionspace}
    \label{fig:ablation_jointfinetuning_supp}
\end{figure}

\subsection{Ablation Study on Joint Finetuning}
\label{sec:ablation_jointfinetuning_supp}

In Figure \ref{fig:ablation_jointfinetuning_supp}, we show that joint-finetuning is indispensable by visualizing its effects on frames without any detected object segmentation masks, which often exist in human-pet interaction videos due to partial occlusions. Since our method does not supervise on frames without segmentation masks during pre-training, the appearance, deformation, and root-body pose of the deformable foreground objects remain uncertain for such frames.

For the \textsc{human1 \& dog1} sequence, the dog ends up penetrating the human arm; for the \textsc{human2 \& cat1} sequence, the cat lies in front of the human hand rather than behind it. Joint-finetuning resolves these issues as it does not optimize a silhouette loss, enabling our scene representation to be supervised on all frames of the training sequence. By jointly optimizing all objects in the scene, our scene representation learns the correct human-pet interactions by reasoning about occlusions, resulting in a general improvement of the visual metrics, as shown in Table \ref{table:ablation_jointfinetuning_supp}. Note that joint-finetuning doesn't always improve the depth metrics. We posit that the depth supervision during pre-training was sufficient in learning a metric model that the qualitative improvements brought by joint-finetuning are not always reflected in the metrics.

\begin{figure*}[!t]
    \captionsetup[subfigure]{labelformat=empty}
    \centering
    \subfloat{{\includegraphics[width=0.48\textwidth]{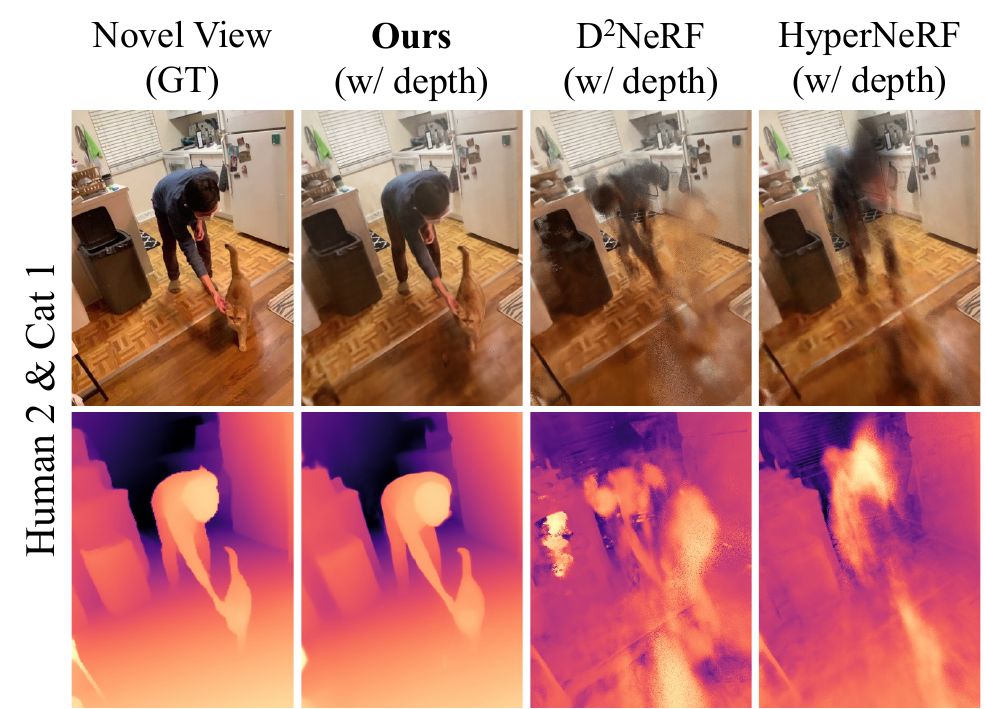}}}\quad%
    \subfloat{{\includegraphics[width=0.48\textwidth]{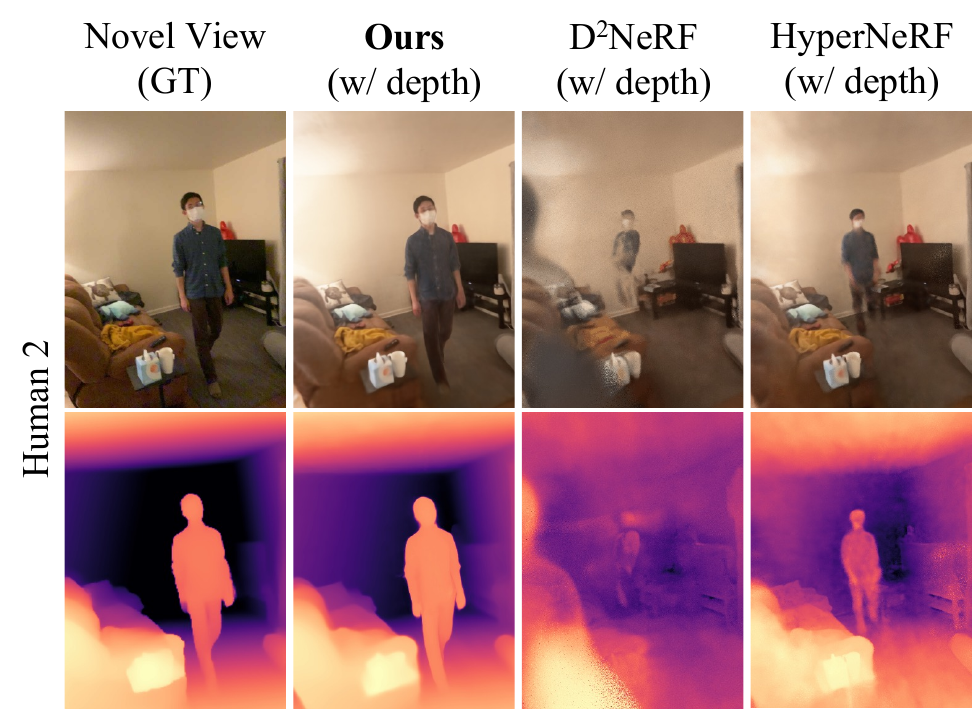}}}\\%
    \subfloat{{\includegraphics[width=0.48\textwidth]{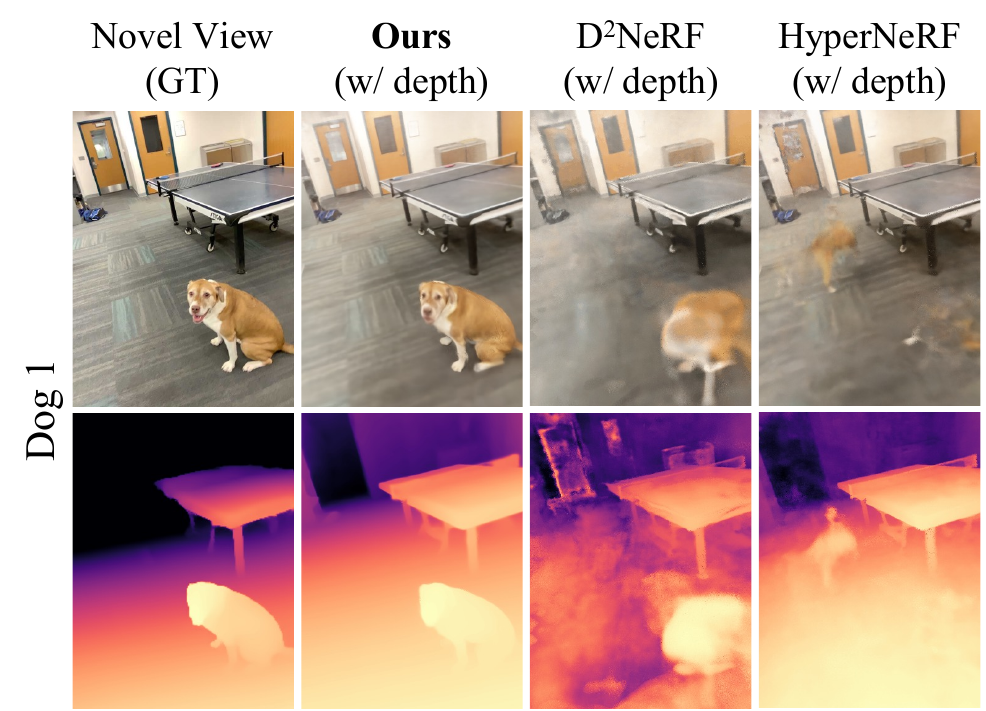}}}\quad%
    \subfloat{{\includegraphics[width=0.48\textwidth]{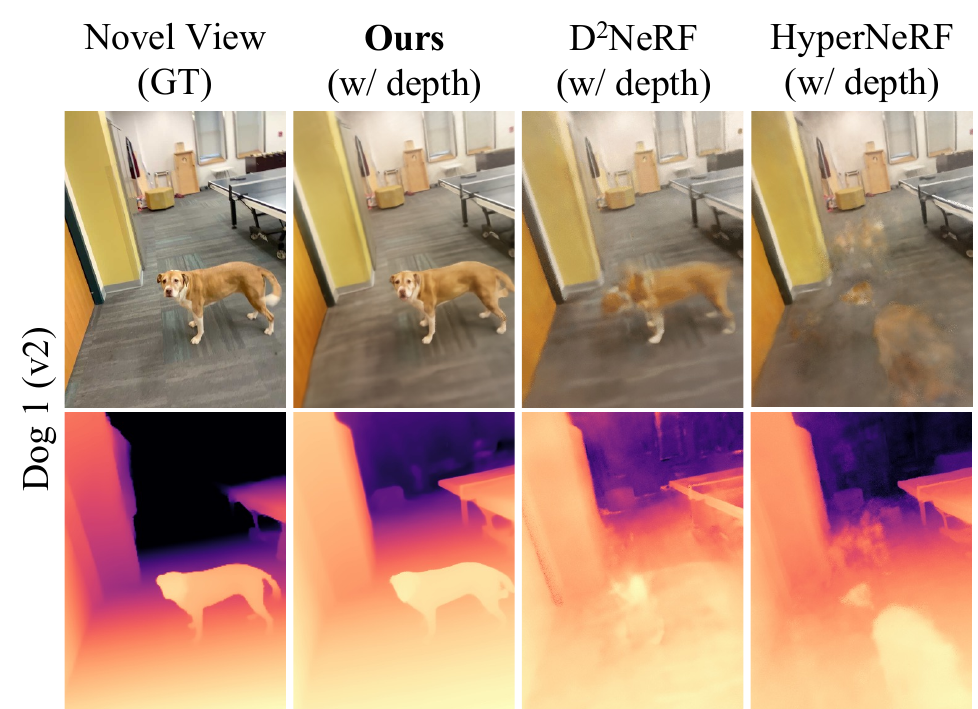}}}\\%
    \subfloat{{\includegraphics[width=0.48\textwidth]{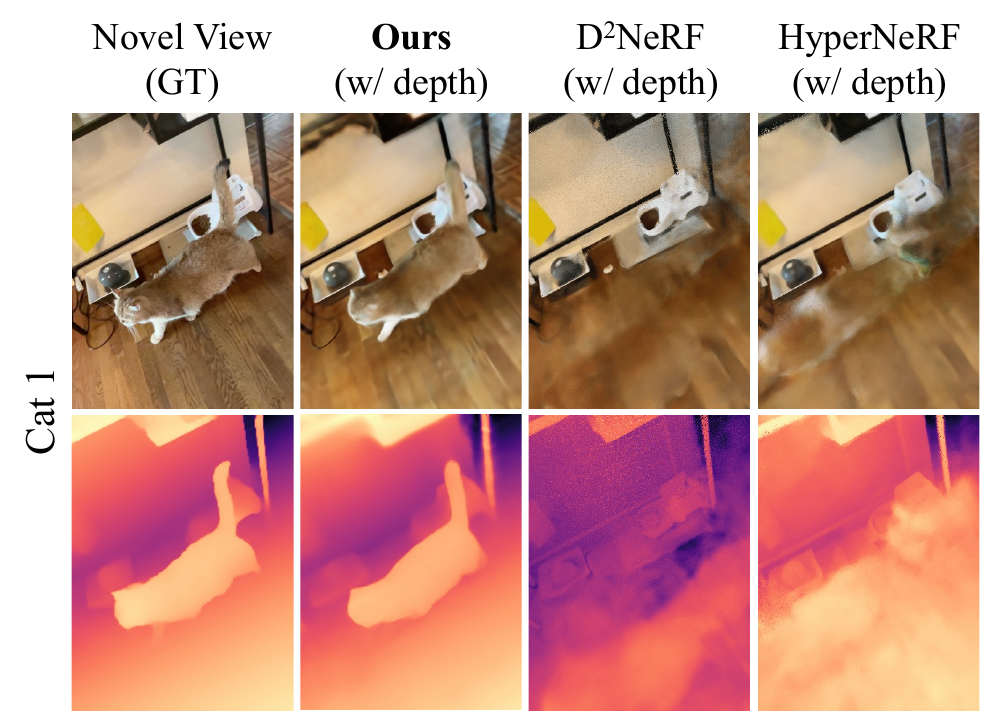}}}\quad%
    \subfloat{{\includegraphics[width=0.48\textwidth]{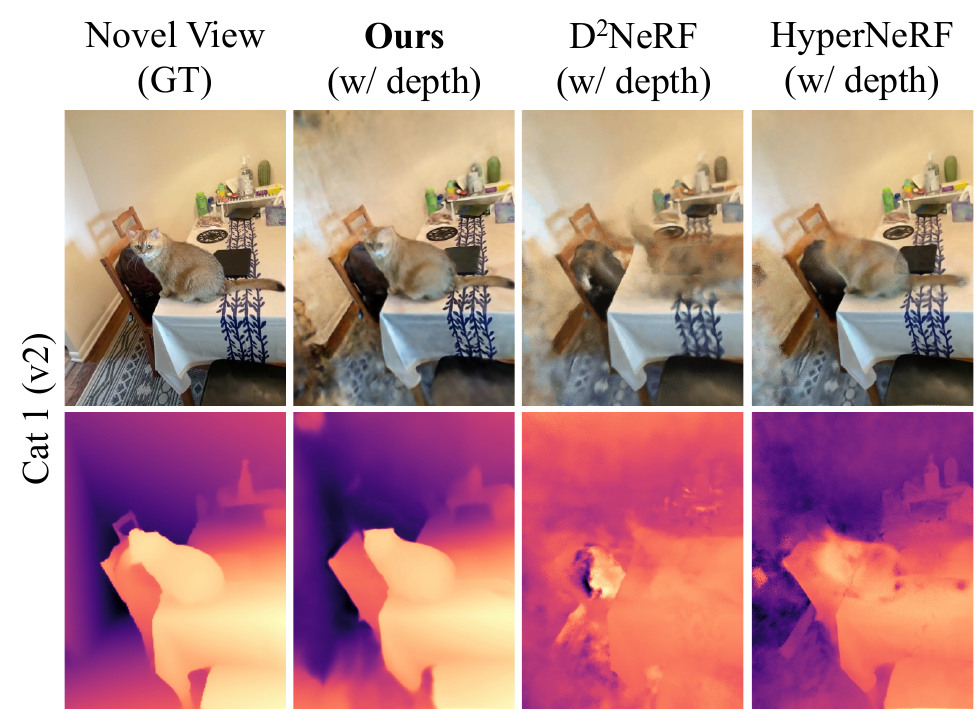}}}\\%
    \caption{\textbf{Qualitative Comparisons on Novel View Synthesis (Additional Sequences).} We compare \ourmethod{} to depth-supervised variants of HyperNeRF \cite{park2021hypernerf} and D$^2$NeRF \cite{d2nerf} on the task of stereo-view synthesis (the left camera is used for training and the images are rendered to the right camera). While the baselines are only able to reconstruct the background at best, \ourmethod{} is able to reconstruct \textit{both} the background and the moving deformable object(s), demonstrating holistic scene reconstruction. \href{https://andrewsonga.github.io/totalrecon/nvs.html}{\textbf{[Videos]}}\captionspace}%
    \label{fig:nvs_comparisons_moreseqs}
\end{figure*}

\begin{figure*}[!t]
    \captionsetup[subfigure]{labelformat=empty}
    \ContinuedFloat
    \centering
    \subfloat{{\includegraphics[width=0.48\textwidth]{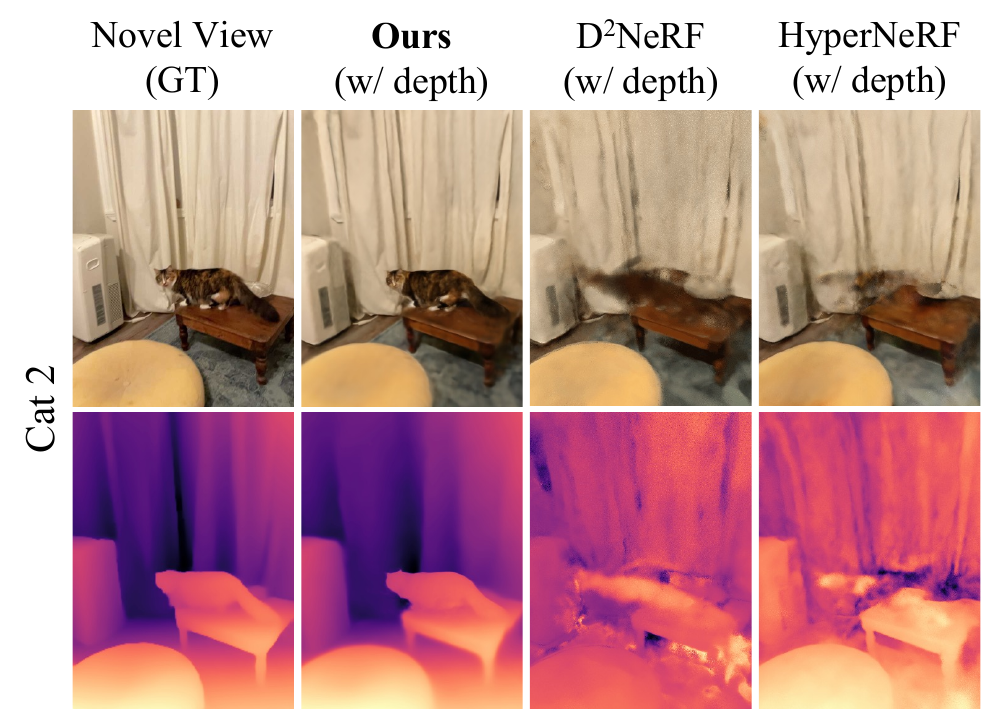}}}\quad%
    \subfloat{{\includegraphics[width=0.48\textwidth]{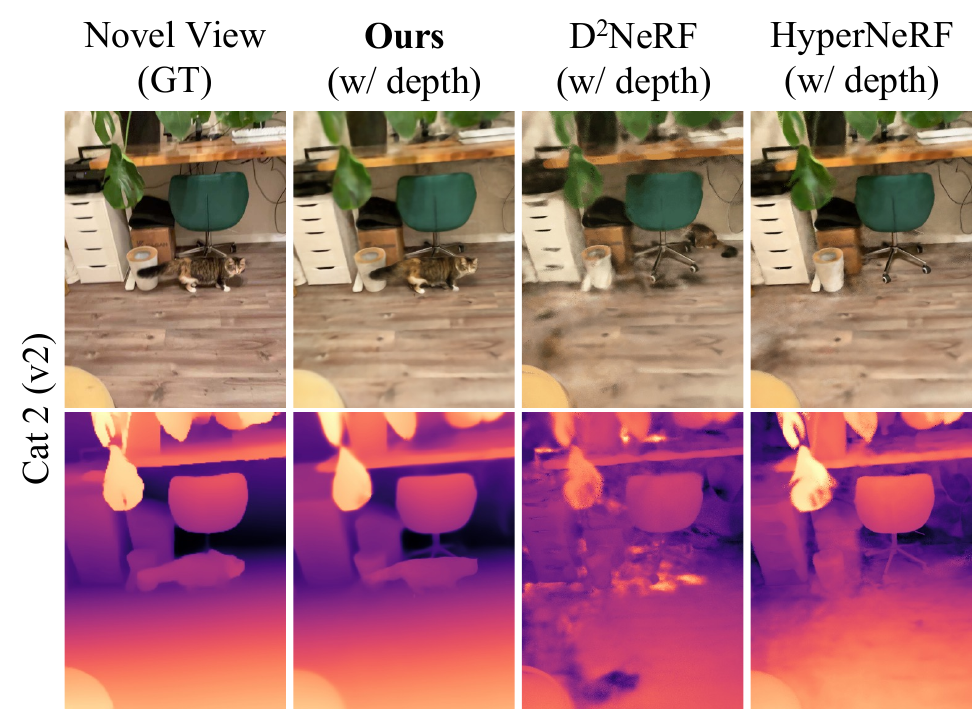}}}\\%
    \subfloat{{\includegraphics[width=0.48\textwidth]{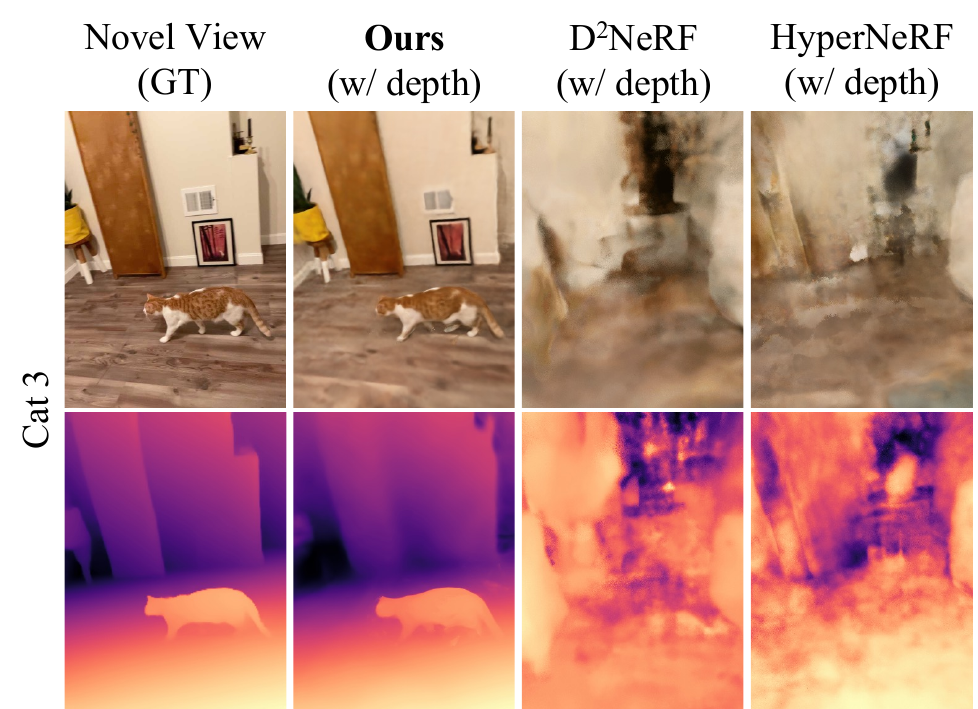}}}\quad%
    \subfloat{{\includegraphics[width=0.48\textwidth]{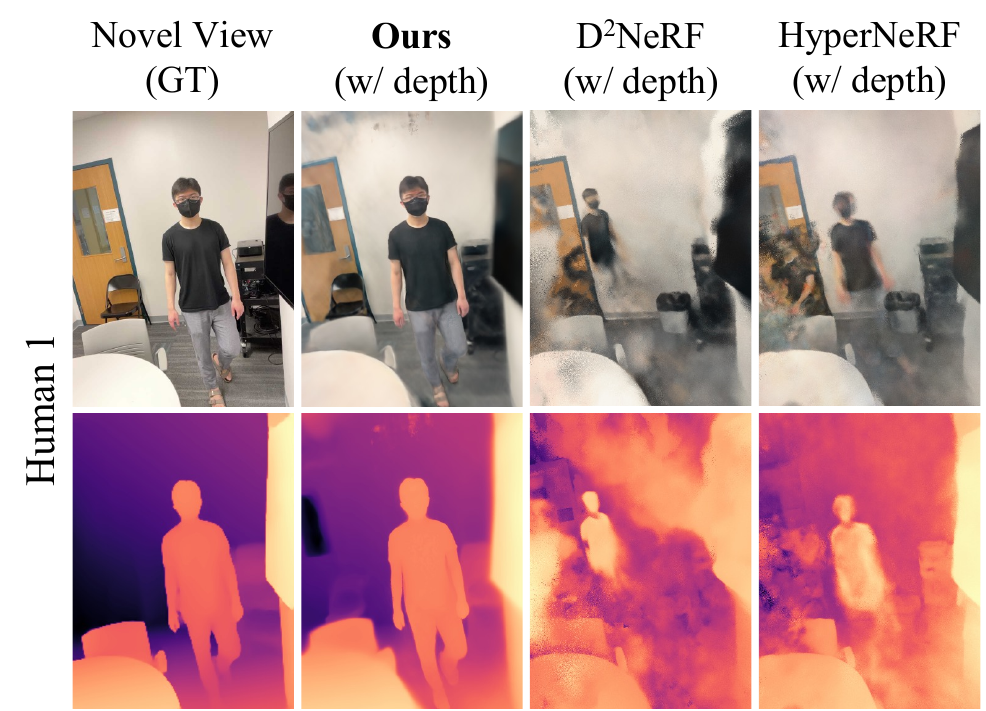}}}%
    \caption{\textbf{[Continued] Qualitative Comparisons on Novel View Synthesis (Additional Sequences).} We compare \ourmethod{} to depth-supervised variants of HyperNeRF \cite{park2021hypernerf} and D$^2$NeRF \cite{d2nerf} on the task of stereo-view synthesis (the left camera is used for training and the images are rendered to the right camera). While the baselines are only able to reconstruct the background at best, \ourmethod{} is able to reconstruct \textit{both} the background and the moving deformable object(s), demonstrating holistic scene reconstruction. \href{https://andrewsonga.github.io/totalrecon/nvs.html}{\textbf{[Videos]}}\captionspace}%
\label{fig:nvs_comparisons_moreseqs_cont}
\end{figure*}

\begin{figure*}[!t]
    \captionsetup[subfigure]{labelformat=empty}
    \centering
    \subfloat{{\includegraphics[width=0.85\textwidth]{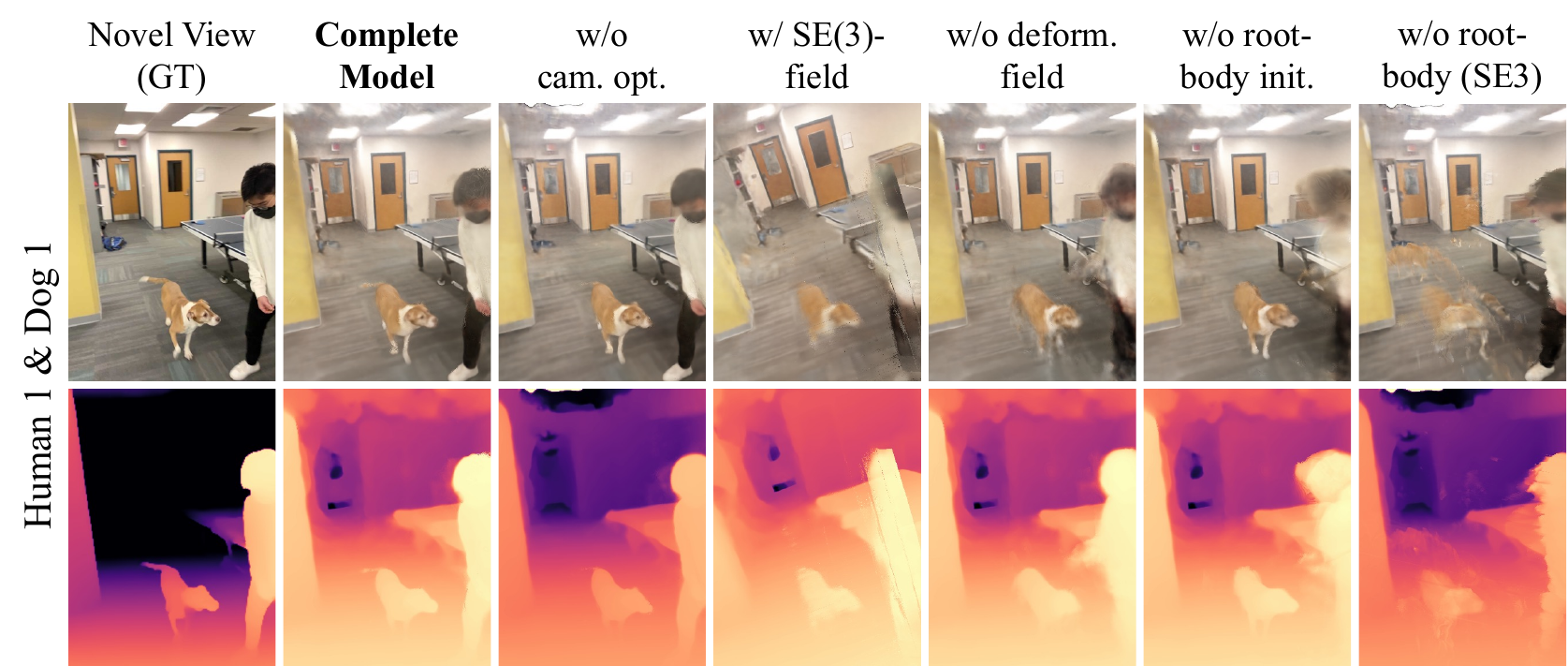}}}\\
    \vspace{2.4mm}
    \subfloat{{\includegraphics[width=0.85\textwidth]{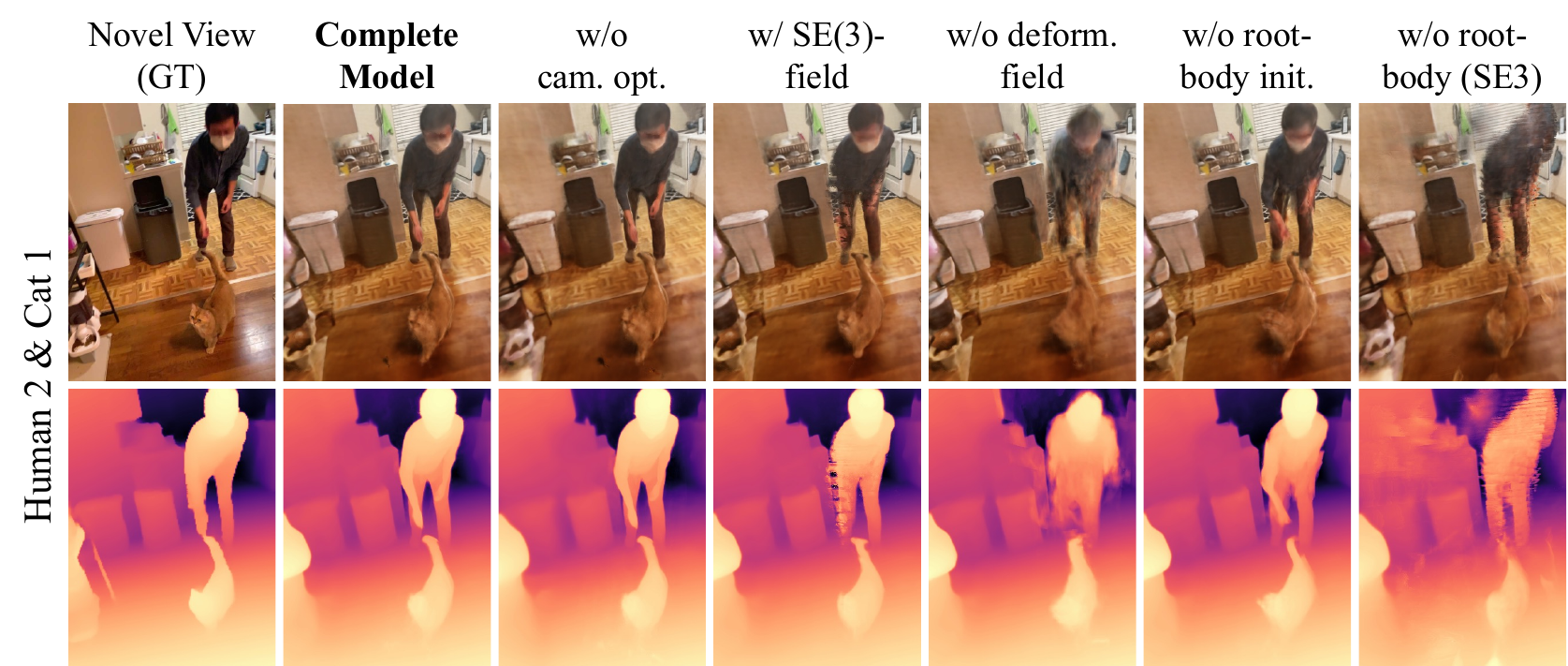}}}\\
    \vspace{2.4mm}
    \subfloat{{\includegraphics[width=0.85\textwidth]{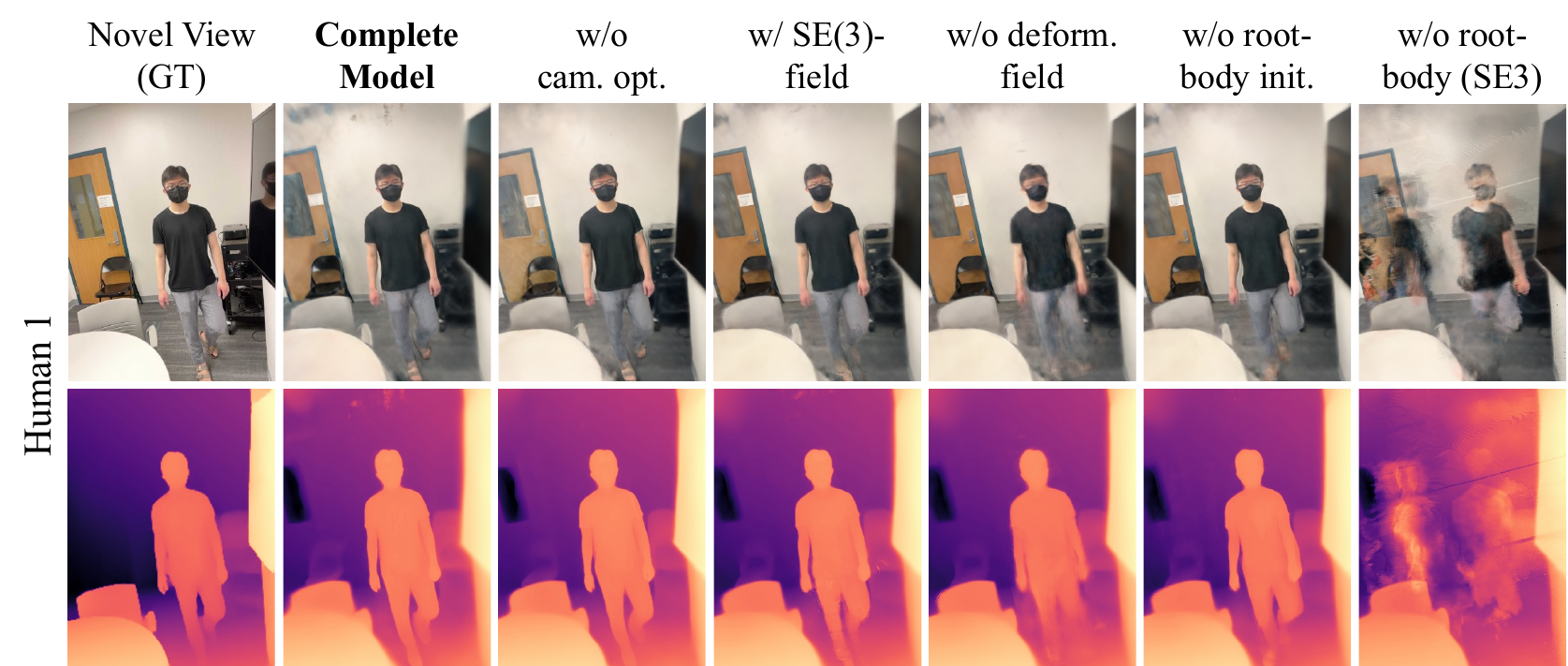}}}\\
    \caption{\textbf{Ablation Study on Motion Modeling.} We render novel views of the ablations in Table \ref{table:ablation_objmotion_extended_perseq}. Ablating camera-pose optimization (\textit{w/o cam. opt.}) does not qualitatively change the scene reconstruction. Changing the deformation field from \ourmethod{}'s neural blend skinning function to an SE(3)-field (\textit{w/ SE(3)-field}) results in minor artifacts in the foreground reconstruction. Removing the deformation field entirely (\textit{w/o deform. field}) produces coarse object reconstructions that fail to model moving body parts such as limbs. Removing PoseNet-initialization of object root-body poses (\textit{w/o root-body init.}) results in noisy and sometimes even failed object reconstructions. We omit the ablation without root-body poses (\textit{w/o root-body}) as it does not converge, and instead present a version that does converge (\textit{w/o root-body (SE3)}). However, this ablation also performs significantly worse than previous ablations, as evidenced by the ghosting artifacts indicative of failed foreground reconstruction. These experiments justify \ourmethod{}'s hierarchical motion representation, which explicitly models objects' root-body motion. \href{https://andrewsonga.github.io/totalrecon/ablation_objmotion.html}{\textbf{[Videos]}}\captionspace}%
    \label{fig:ablation_objmotion_full}
\end{figure*}

\begin{figure*}[!t]
    \captionsetup[subfigure]{labelformat=empty}
    \ContinuedFloat
    \centering
    \subfloat{{\includegraphics[width=0.85\textwidth]{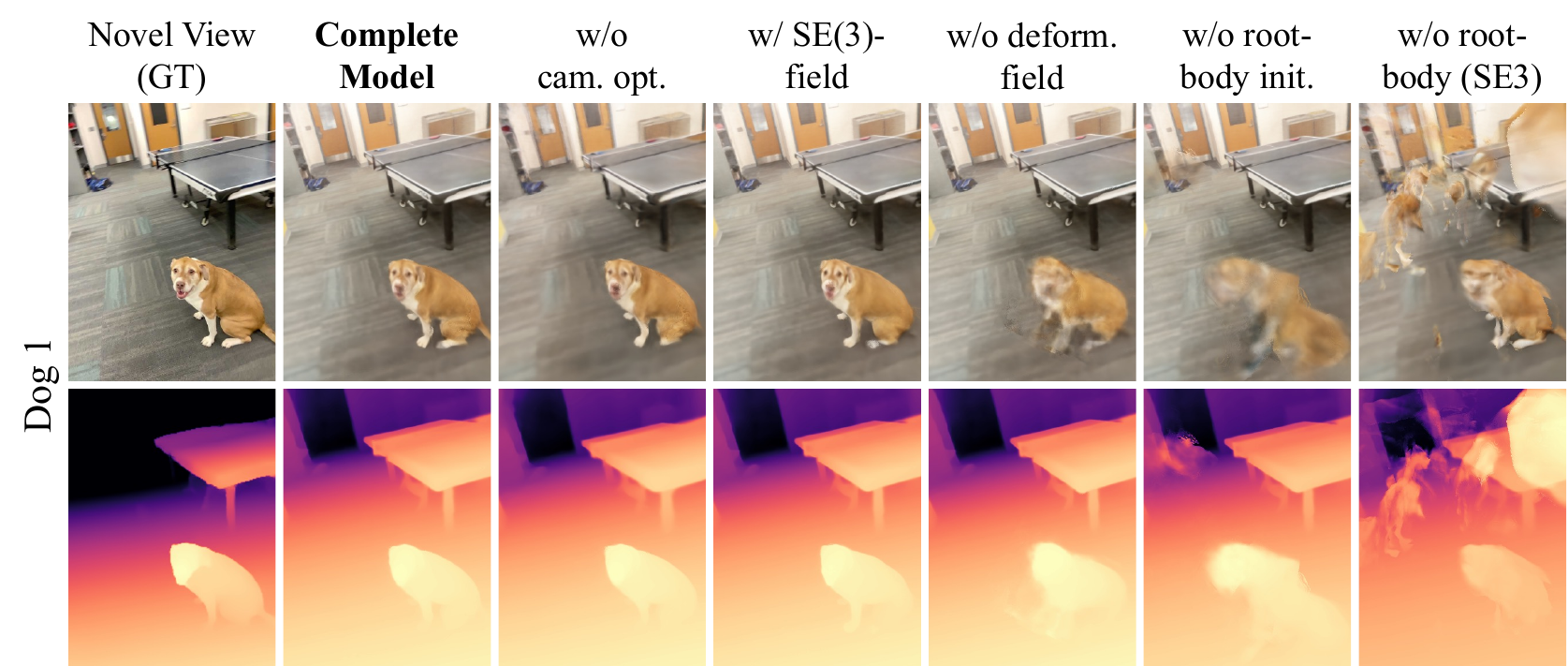}}}\\
    \vspace{2.4mm}
    \subfloat{{\includegraphics[width=0.85\textwidth]{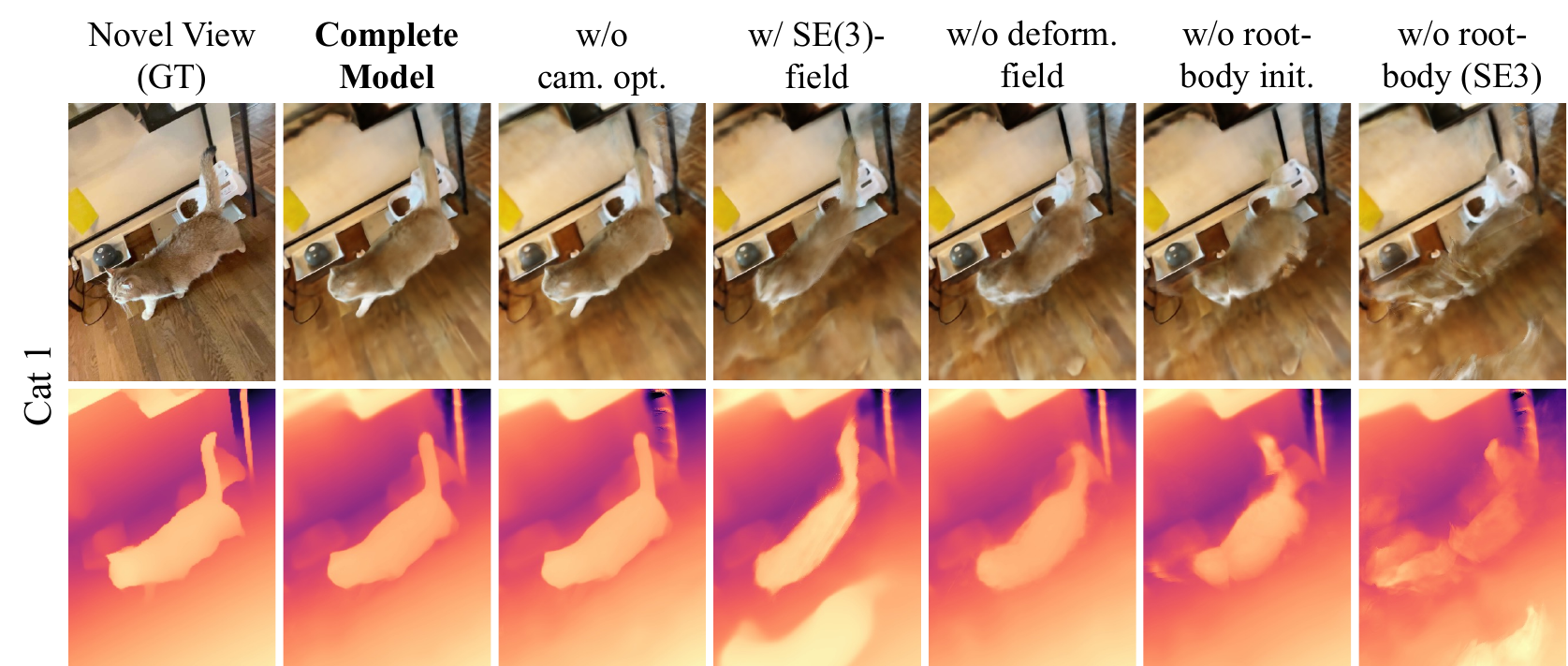}}}\\
    \vspace{2.4mm}
    \subfloat{{\includegraphics[width=0.85\textwidth]{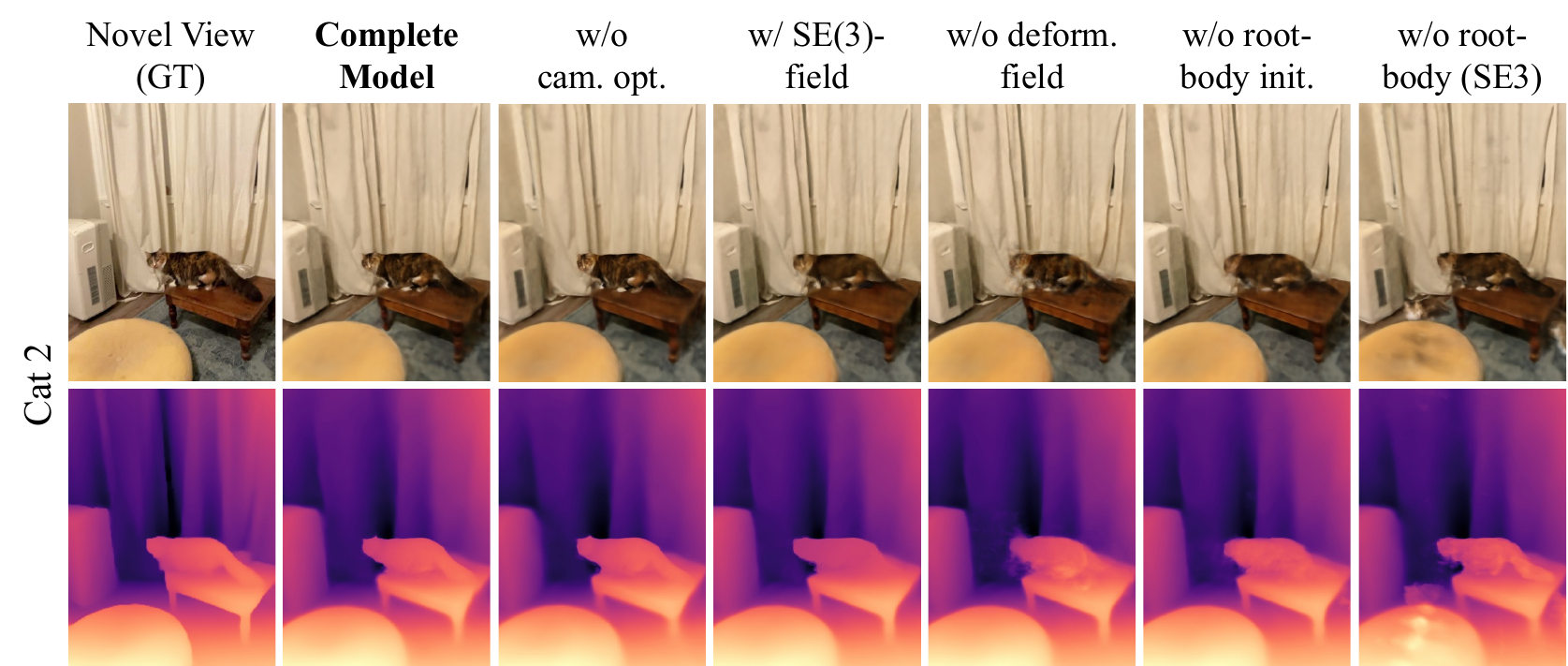}}}\\
    \caption{\textbf{[Continued] Ablation Study on Motion Modeling.} We render novel views of the ablations in Table \ref{table:ablation_objmotion_extended_perseq}. Ablating camera-pose optimization (\textit{w/o cam. opt.}) does not qualitatively change the scene reconstruction. Changing the deformation field from \ourmethod{}'s neural blend skinning function to an SE(3)-field (\textit{w/ SE(3)-field}) results in minor artifacts in the foreground reconstruction. Removing the deformation field entirely (\textit{w/o deform. field}) produces coarse object reconstructions that fail to model moving body parts such as limbs. Removing PoseNet-initialization of object root-body poses (\textit{w/o root-body init.}) results in noisy and sometimes even failed object reconstructions. We omit the ablation without root-body poses (\textit{w/o root-body}) as it does not converge, and instead present a version that does converge (\textit{w/o root-body (SE3)}). However, this ablation also performs significantly worse than previous ablations, as evidenced by the ghosting artifacts indicative of failed foreground reconstruction. These experiments justify \ourmethod{}'s hierarchical motion representation, which explicitly models objects' root-body motion. \href{https://andrewsonga.github.io/totalrecon/ablation_objmotion.html}{\textbf{[Videos]}}\captionspace}%
\end{figure*}

\section{Societal Impact}
\label{sec:societal_impact}

We have demonstrated that our method can holistically reconstruct a dynamic scene containing multiple deformable objects, such as humans and pets - all from a single RGBD video captured from a commodity consumer device. We believe that a truly holistic reconstruction of the background geometry, each moving object, its own deformation, and camera pose would enable a number of new applications ranging from augmented reality to asset generation for virtual worlds, especially given the ubiquity of consumer-grade RGBD sensors.

However, the reconstruction capabilities of our method could be a double-edged sword; the very ease with which one could reconstruct a realistic 3D human model from nothing but a casually captured RGBD video poses potential privacy concerns. For instance, one could extract sensitive personal information such as height and other body measurements from a metric human model reconstructed with our method. In terms of appearance synthesis, our method poses similar types of risks as Deepfakes pose to society, especially given that the deformable object model used in our method is animatable (\textit{i.e.}, user-drivable) \cite{yang2022banmo}. An important future direction of research that needs to accompany 3D reconstruction research would therefore be methods of distinguishing photorealistic rendered videos from genuine content.

%% file: tables/comparisons_entireimg_visual_allseqs_part1.tex
\small{
\begin{tabu}{l||ccc|ccc|ccc|ccc|ccc|ccc}

\toprule

& \multicolumn{ 3 }{c}{
  \makecell{
  \textsc{\small Dog 1 (v1)}
\\\small(626 images)
  }
}
& \multicolumn{ 3 }{c}{
  \makecell{
  \textsc{\small Dog 1 (v2)}
\\\small(531 images)
  }
}
& \multicolumn{ 3 }{c}{
  \makecell{
  \textsc{\small Cat 1 (v1)}
\\\small(641 images)
  }
}
& \multicolumn{ 3 }{c}{
  \makecell{
  \textsc{\small Cat 1 (v2)}
\\\small(632 images)
  }
}
& \multicolumn{ 3 }{c}{
  \makecell{
  \textsc{\small Cat 2 (v1)}
\\\small(834 images)
  }
}
& \multicolumn{ 3 }{c}{
  \makecell{
  \textsc{\small Cat 2 (v2)}
\\\small(901 images)
  }
}
\\

& \multicolumn{1}{c}{ \footnotesize LPIPS$\downarrow$ }
& \multicolumn{1}{c}{ \footnotesize PSNR$\uparrow$ }
& \multicolumn{1}{c}{ \footnotesize SSIM$\uparrow$ }
& \multicolumn{1}{c}{ \footnotesize LPIPS$\downarrow$ }
& \multicolumn{1}{c}{ \footnotesize PSNR$\uparrow$ }
& \multicolumn{1}{c}{ \footnotesize SSIM$\uparrow$ }
& \multicolumn{1}{c}{ \footnotesize LPIPS$\downarrow$ }
& \multicolumn{1}{c}{ \footnotesize PSNR$\uparrow$ }
& \multicolumn{1}{c}{ \footnotesize SSIM$\uparrow$ }
& \multicolumn{1}{c}{ \footnotesize LPIPS$\downarrow$ }
& \multicolumn{1}{c}{ \footnotesize PSNR$\uparrow$ }
& \multicolumn{1}{c}{ \footnotesize SSIM$\uparrow$ }
& \multicolumn{1}{c}{ \footnotesize LPIPS$\downarrow$ }
& \multicolumn{1}{c}{ \footnotesize PSNR$\uparrow$ }
& \multicolumn{1}{c}{ \footnotesize SSIM$\uparrow$ }
& \multicolumn{1}{c}{ \footnotesize LPIPS$\downarrow$ }
& \multicolumn{1}{c}{ \footnotesize PSNR$\uparrow$ }
& \multicolumn{1}{c}{ \footnotesize SSIM$\uparrow$ }
\\
\hline

  HyperNeRF~\cite{park2021hypernerf}
  &$.634$
  &$12.84$
  &$.673$
  
  &$.432$
  &$14.27$
  &$.721$
  
  &$.521$
  &$14.86$
  &$.632$
  
  &$.438$
  &$14.87$
  &$.597$
  
  &$.641$
  &$12.32$
  &$.632$
  
  &$.397$
  &$15.68$
  &$.657$
  
  \\   D$^{2}$NeRF~\cite{d2nerf}
  &$.540$
  &$13.37$
  &$.694$

  &$.546$
  &$11.74$
  &$.685$
  
  &$.687$
  &$10.92$
  &$.545$

  &$.588$
  &$11.88$
  &$.548$
  
  &$.556$
  &$12.55$
  &$.664$

  &$.595$
  &$12.71$
  &$.604$
  \\
  \hline
  HyperNeRF (w/ depth)
  & $.373$ & $16.86$ & $.730$
  & $.425$ & $16.95$ & $.740$
  & $.532$ & $14.37$ & $.621$
  & $.371$ & $15.65$ & $.617$
  & $.330$ & $18.47$ & $.728$
  & $.376$ & $16.56$ & $.670$
  \\  D$^{2}$NeRF (w/ depth)
  &$.507$
  &$13.44$
  &$.698$

  &$.532$
  &$11.88$
  &$.690$
  
  &$.685$
  &$10.81$
  &$.534$
 
  &$.580$
  &$12.00$
  &$.563$
  
  &$.561$
  &$12.59$
  &$.656$

  &$.553$
  &$12.76$
  &$.629$
  
  \\\hline   \textbf{Ours} (w/ depth)
  &$\textbf{.271}$
  &$\textbf{17.60}$
  &$\textbf{.745}$
  
  &$\textbf{.313}$
  &$\textbf{17.78}$
  &$\textbf{.768}$
  
  &$\textbf{.382}$
  &$\textbf{15.77}$
  &$\textbf{.657}$
  
  &$\textbf{.333}$
  &$\textbf{16.44}$
  &$\textbf{.652}$
  
  &$\textbf{.237}$
  &$\textbf{21.22}$
  &$\textbf{.793}$

  &$\textbf{.281}$
  &$\textbf{18.52}$
  &$\textbf{.713}$
  \\ \bottomrule

\end{tabu}
}

%% file: tables/comparisons_entireimg_visual_allseqs_part2.tex
\small{
\begin{tabu}{l||ccc|ccc|ccc|ccc|ccc|ccc}

\specialrule{0pt}{1pt}{1pt}

& \multicolumn{ 3 }{c}{
  \makecell{
  \textsc{\small Cat 3}
\\\small(767 images)
  }
}
& \multicolumn{ 3 }{c}{
  \makecell{
  \textsc{\small Human 1}
\\\small(550 images)
  }
}
& \multicolumn{ 3 }{c}{
  \makecell{
  \textsc{\small Human 2}
\\\small(483 images)
  }
}
& \multicolumn{ 3 }{c}{
  \makecell{
  \textsc{\small Human - Dog}
\\\small(392 images)
  }
}
& \multicolumn{ 3 }{c|}{
  \makecell{
  \textsc{\small Human - Cat}
\\\small(431 images)
  }
}
& \multicolumn{ 3 }{c}{
  \makecell{
  \textsc{\small Mean }
  }
}
\\

& \multicolumn{1}{c}{ \footnotesize LPIPS$\downarrow$ }
& \multicolumn{1}{c}{ \footnotesize PSNR$\uparrow$ }
& \multicolumn{1}{c}{ \footnotesize SSIM$\uparrow$ }
& \multicolumn{1}{c}{ \footnotesize LPIPS$\downarrow$ }
& \multicolumn{1}{c}{ \footnotesize PSNR$\uparrow$ }
& \multicolumn{1}{c}{ \footnotesize SSIM$\uparrow$ }
& \multicolumn{1}{c}{ \footnotesize LPIPS$\downarrow$ }
& \multicolumn{1}{c}{ \footnotesize PSNR$\uparrow$ }
& \multicolumn{1}{c}{ \footnotesize SSIM$\uparrow$ }
& \multicolumn{1}{c}{ \footnotesize LPIPS$\downarrow$ }
& \multicolumn{1}{c}{ \footnotesize PSNR$\uparrow$ }
& \multicolumn{1}{c}{ \footnotesize SSIM$\uparrow$ }
& \multicolumn{1}{c}{ \footnotesize LPIPS$\downarrow$ }
& \multicolumn{1}{c}{ \footnotesize PSNR$\uparrow$ }
& \multicolumn{1}{c|}{ \footnotesize SSIM$\uparrow$ }
& \multicolumn{1}{c}{ \footnotesize LPIPS$\downarrow$ }
& \multicolumn{1}{c}{ \footnotesize PSNR$\uparrow$ }
& \multicolumn{1}{c}{ \footnotesize SSIM$\uparrow$ }
\\
\hline

  HyperNeRF~\cite{park2021hypernerf}
  &$.592$
  &$13.74$
  &$.624$
  
  &$.632$
  &$11.94$
  &$.603$
  
  &$.585$
  &$14.97$
  &$.620$
  
  &$.487$
  &$15.04$
  &$.699$
  
  &$.462$
  &$13.52$
  &$.512$
  
  &$.531$
  &$14.00$
  &$.635$
  
  \\   D$^{2}$NeRF~\cite{d2nerf}
  &$.759$
  &$11.03$
  &$.578$
  
  &$.588$
  &$11.88$
  &$.638$

  &$.630$
  &$12.13$
  &$.599$
  
  &$.576$
  &$12.41$
  &$.652$
  
  &$.628$
  &$10.41$
  &$.453$
  
  &$.611$
  &$11.97$
  &$.608$

  \\ \hline
  HyperNeRF (w/ depth)
  & $.514$ & $14.86$ & $.635$
  & $.501$ & $13.25$ & $.664$
  & $.445$ & $15.58$ & $.665$
  & $.450$ & $15.01$ & $.704$
  & $.456$ & $14.40$ & $.535$
  & $.428$ & $15.80$ & $.667$
  
  \\  D$^{2}$NeRF (w/ depth)
  &$.730$
  &$11.08$
  &$.582$
  
  &$.585$
  &$12.14$
  &$.638$

  &$.609$
  &$12.11$
  &$.612$

  &$.608$
  &$12.30$
  &$.633$
  
  &$.645$
  &$10.51$
  &$.451$
  
  &$.599$
  &$12.02$
  &$.611$
  \\ \hline
  
  \textbf{Ours} (w/ depth)
  &$\textbf{.261}$
  &$\textbf{19.89}$
  &$\textbf{.734}$
  
  &$\textbf{.213}$
  &$\textbf{18.39}$
  &$\textbf{.778}$
  
  &$\textbf{.264}$
  &$\textbf{16.73}$
  &$\textbf{.712}$
  
  &$\textbf{.256}$
  &$\textbf{16.69}$
  &$\textbf{.756}$
  
  &$\textbf{.233}$
  &$\textbf{17.67}$
  &$\textbf{.630}$
  
  &$\textbf{.278}$
  &$\textbf{18.11}$
  &$\textbf{.724}$
  \\ \bottomrule

\end{tabu}
}

%% file: tables/comparisons_entireimg_depth_allseqs_part1.tex
\small{
\begin{tabu}{l||cc|cc|cc|cc|cc|cc}

\toprule

& \multicolumn{ 2 }{c}{
  \makecell{
  \textsc{\small Dog 1 (v1)}
\\\small(626 images)
  }
}
& \multicolumn{ 2 }{c}{
  \makecell{
  \textsc{\small Dog 1 (v2)}
\\\small(531 images)
  }
}
& \multicolumn{ 2 }{c}{
  \makecell{
  \textsc{\small Cat 1 (v1)}
\\\small(641 images)
  }
}
& \multicolumn{ 2 }{c}{
  \makecell{
  \textsc{\small Cat 1 (v2)}
\\\small(632 images)
  }
}
& \multicolumn{ 2 }{c}{
  \makecell{
  \textsc{\small Cat 2 (v1)}
\\\small(834 images)
  }
}
& \multicolumn{ 2 }{c}{
  \makecell{
  \textsc{\small Cat 2 (v2)}
\\\small(901 images)
  }
}
\\

& \multicolumn{1}{c}{ \footnotesize Acc@0.1m$\uparrow$ }
& \multicolumn{1}{c}{ $\epsilon_\textrm{depth}$$\downarrow$ }
& \multicolumn{1}{c}{ \footnotesize Acc@0.1m$\uparrow$ }
& \multicolumn{1}{c}{ $\epsilon_\textrm{depth}$$\downarrow$ }
& \multicolumn{1}{c}{ \footnotesize Acc@0.1m$\uparrow$ }
& \multicolumn{1}{c}{ $\epsilon_\textrm{depth}$$\downarrow$ }
& \multicolumn{1}{c}{ \footnotesize Acc@0.1m$\uparrow$ }
& \multicolumn{1}{c}{ $\epsilon_\textrm{depth}$$\downarrow$ }
& \multicolumn{1}{c}{ \footnotesize Acc@0.1m$\uparrow$ }
& \multicolumn{1}{c}{ $\epsilon_\textrm{depth}$$\downarrow$ }
& \multicolumn{1}{c}{ \footnotesize Acc@0.1m$\uparrow$ }
& \multicolumn{1}{c}{ $\epsilon_\textrm{depth}$$\downarrow$ }
\\
\hline

  HyperNeRF~\cite{park2021hypernerf}
  &$.107$ &$.687$
  
  &$.176$ &$.870$

  &$.316$ &$.476$

  &$.314$ &$.564$

  &$.277$ &$.765$

  &$.252$ &$.811$

  \\   D$^{2}$NeRF~\cite{d2nerf}
  &$.219$ &$.463$

  &$.220$ &$.456$

  &$.346$ &$.334$

  &$.403$ &$.314$
  
  &$.333$ &$.371$
  
  &$.339$ &$.361$
  
  \\ \hline
  HyperNeRF (w/ depth)
  & $.352$ & $.331$ 
  & $.357$ & $.338$ 
  & $.552$ & $.206$
  & $.596$ & $.209$ 
  & $.605$ & $.154$ 
  & $.612$ & $.170$

  \\  D$^{2}$NeRF (w/ depth)
  &$.338$ &$.423$

  &$.270$ &$.445$
  
  &$.510$ &$.325$

  &$.362$ &$.313$
  
  &$.438$ &$.298$

  &$.376$ &$.318$
  
  \\ \hline   \textbf{Ours} (w/ depth)
  &$\textbf{.841}$ &$\textbf{.165}$
  
  &$\textbf{.790}$ &$\textbf{.167}$
  
  &$\textbf{.889}$ &$\textbf{.184}$
  
  &$\textbf{.894}$ &$\textbf{.124}$
  
  &$\textbf{.967}$ &$\textbf{.050}$
  
  &$\textbf{.925}$ &$\textbf{.081}$
  \\ \bottomrule

\end{tabu}
}

%% file: tables/comparisons_entireimg_depth_allseqs_part2.tex
\small{
\begin{tabu}{l||cc|cc|cc|cc|cc|cc}

\specialrule{0pt}{1pt}{1pt}

& \multicolumn{ 2 }{c}{
  \makecell{
  \textsc{\small Cat 3}
\\\small(767 images)
  }
}
& \multicolumn{ 2 }{c}{
  \makecell{
  \textsc{\small Human 1}
\\\small(550 images)
  }
}
& \multicolumn{ 2 }{c}{
  \makecell{
  \textsc{\small Human 2}
\\\small(483 images)
  }
}
& \multicolumn{ 2 }{c}{
  \makecell{
  \textsc{\small Human - Dog}
\\\small(392 images)
  }
}
& \multicolumn{ 2 }{c|}{
  \makecell{
  \textsc{\small Human - Cat}
\\\small(431 images)
  }
}
& \multicolumn{ 2 }{c}{
  \makecell{
  \textsc{\small Mean }
  }
}
\\

& \multicolumn{1}{c}{ \footnotesize Acc@0.1m$\uparrow$ }
& \multicolumn{1}{c}{ $\epsilon_\textrm{depth}$$\downarrow$ }
& \multicolumn{1}{c}{ \footnotesize Acc@0.1m$\uparrow$ }
& \multicolumn{1}{c}{ $\epsilon_\textrm{depth}$$\downarrow$ }
& \multicolumn{1}{c}{ \footnotesize Acc@0.1m$\uparrow$ }
& \multicolumn{1}{c}{ $\epsilon_\textrm{depth}$$\downarrow$ }
& \multicolumn{1}{c}{ \footnotesize Acc@0.1m$\uparrow$ }
& \multicolumn{1}{c}{ $\epsilon_\textrm{depth}$$\downarrow$ }
& \multicolumn{1}{c}{ \footnotesize Acc@0.1m$\uparrow$ }
& \multicolumn{1}{c|}{ $\epsilon_\textrm{depth}$$\downarrow$ }
& \multicolumn{1}{c}{ \footnotesize Acc@0.1m$\uparrow$ }
& \multicolumn{1}{c}{ $\epsilon_\textrm{depth}$$\downarrow$ }
\\
\hline

  HyperNeRF~\cite{park2021hypernerf}
  &$.213$ &$.800$

  &$.053$ &$.821$

  &$.067$ &$1.665$

  &$.072$ &$.894$

  &$.162$ &$.862$

  &$.198$ &$.855$

  \\   D$^{2}$NeRF~\cite{d2nerf}
  &$.231$ &$.523$
  &$.066$ &$1.063$
  &$.128$ &$.890$
  &$.078$ &$.847$
  &$.126$ &$.880$
  &$.247$ &$.739$
  
  \\
  \hline
  HyperNeRF (w/ depth)
  & $.451$ & $.285$ 
  & $.211$ & $.591$ 
  & $.249$ & $.611$ 
  & $.283$ & $.565$ 
  & $.214$ & $.613$ 
  & $.439$ & $.374$ 
  
  \\  D$^{2}$NeRF (w/ depth)
  &$.243$ &$.496$
  &$.086$ &$.984$
  &$.131$ &$.813$
  &$.154$ &$.789$
  &$.176$ &$.757$
  &$.302$ &$.549$
  
  \\ \hline   \textbf{Ours} (w/ depth)
  &$\textbf{.949}$ &$\textbf{.066}$
  &$\textbf{.909}$ &$\textbf{.142}$
  &$\textbf{.849}$ &$\textbf{.142}$
  &$\textbf{.827}$ &$\textbf{.204}$
  &$\textbf{.914}$ &$\textbf{.104}$
  &$\textbf{.895}$ &$\textbf{.131}$
  
  \\ \bottomrule

\end{tabu}
}

%% file: tables/ablation_depth_with_acc.tex
\small{
\begin{tabular}{l||cc|cc|cc|cc|cc|cc|cc}

\toprule

& \multicolumn{ 2 }{c}{
  \makecell{
  \textsc{\small Dog 1}
\\\small(626 images)
  }
}
& \multicolumn{ 2 }{c}{
  \makecell{
  \textsc{\small Cat 1 }
\\\small(641 images)
  }
}
& \multicolumn{ 2 }{c}{
  \makecell{
  \textsc{\small Cat 2 }
\\\small(834 images)
  }
}
& \multicolumn{ 2 }{c}{
  \makecell{
  \textsc{\small Human 1 }
\\\small(550 images)
  }
}
& \multicolumn{ 2 }{c}{
  \makecell{
  \textsc{\small Human - Dog}
\\\small(392 images)
  }
}
& \multicolumn{ 2 }{c|}{
  \makecell{
  \textsc{\small Human - Cat}
\\\small(431 images)
  }
}
& \multicolumn{ 2 }{c}{
  \makecell{
  \textsc{\small Mean }
  }
}
\\

& \multicolumn{1}{c}{ \footnotesize LPIPS$\downarrow$ }
& \multicolumn{1}{c}{ \footnotesize Acc@0.1m$\uparrow$ }
& \multicolumn{1}{c}{ \footnotesize LPIPS$\downarrow$ }
& \multicolumn{1}{c}{ \footnotesize Acc@0.1m$\uparrow$ }
& \multicolumn{1}{c}{ \footnotesize LPIPS$\downarrow$ }
& \multicolumn{1}{c}{ \footnotesize Acc@0.1m$\uparrow$ }
& \multicolumn{1}{c}{ \footnotesize LPIPS$\downarrow$ }
& \multicolumn{1}{c}{ \footnotesize Acc@0.1m$\uparrow$ }
& \multicolumn{1}{c}{ \footnotesize LPIPS$\downarrow$ }
& \multicolumn{1}{c}{ \footnotesize Acc@0.1m$\uparrow$ }
& \multicolumn{1}{c}{ \footnotesize LPIPS$\downarrow$ }
& \multicolumn{1}{c|}{ \footnotesize Acc@0.1m$\uparrow$ }
& \multicolumn{1}{c}{ \footnotesize LPIPS$\downarrow$ }
& \multicolumn{1}{c}{ \footnotesize Acc@0.1m$\uparrow$ }
\\
\hline

  \small w/o depth
  &$.307$
  &$.296$
  
  &$.496$
  &$.051$

  &$.287$
  &$.193$
  
  &$.314$
  &$.125$
  
  &$.376$
  &$.206$
  
  &$.519$
  &$.017$
  
  &$.372$
  &$.154$
  
  \\ \small \textbf{Full} (w/ depth)
  &$\textbf{.271}$
  &$\textbf{.841}$
  
  &$\textbf{.382}$
  &$\textbf{.889}$

  &$\textbf{.237}$
  &$\textbf{.967}$
  
  &$\textbf{.213}$
  &$\textbf{.909}$
  
  &$\textbf{.256}$
  &$\textbf{.827}$
  
  &$\textbf{.233}$
  &$\textbf{.914}$
  
  &$\textbf{.268}$
  &$\textbf{.898}$
  \\ \bottomrule

\end{tabular}
}

%% file: tables/ablation_objmotion_extended_perseq_with_acc.tex
\small{
\begin{tabular}{l||ccccc|cccccccccccc|cc}
    \toprule
    \multirow{2}{*}{\footnotesize \makecell{Methods}} 
    & \multirow{2}{*}{\footnotesize \makecell{Optimizes\\Camera}}
    & \multirow{2}{*}{\footnotesize \makecell{Deformation\\Field}}
    & \multirow{2}{*}{\footnotesize \makecell{Deformable\\Objects}}
    & \multirow{2}{*}{\footnotesize \makecell{Root-Body\\Initialization}}
    & \multirow{2}{*}{\footnotesize \makecell{Root-Body\\Motion}}
    & \multicolumn{ 2 }{c}{
      \makecell{
      \textsc{\small Dog 1}
    \\\small(626 images)
      }
    }
    & \multicolumn{ 2 }{c}{
      \makecell{
      \textsc{\small Cat 1 }
    \\\small(641 images)
      }
    }
    & \multicolumn{ 2 }{c}{
      \makecell{
      \textsc{\small Cat 2 }
    \\\small(834 images)
      }
    }
    & \multicolumn{ 2 }{c}{
      \makecell{
      \textsc{\small Human 1 }
    \\\small(550 images)
      }
    }
    & \multicolumn{ 2 }{c}{
      \makecell{
      \textsc{\small Human - Dog}
    \\\small(392 images)
      }
    }
    & \multicolumn{ 2 }{c|}{
      \makecell{
      \textsc{\small Human - Cat}
    \\\small(431 images)
      }
    }
    & \multicolumn{ 2 }{c}{
      \makecell{
      \textsc{\small Mean }
      }
    }
    \\
    & 
    & 
    &
    & 
    &
    & \multicolumn{1}{c}{ \footnotesize LPIPS$\downarrow$ }
    & \multicolumn{1}{c}{ \footnotesize   Acc$\uparrow$ }
    & \multicolumn{1}{c}{ \footnotesize LPIPS$\downarrow$ }
    & \multicolumn{1}{c}{ \footnotesize Acc$\uparrow$ }
    & \multicolumn{1}{c}{ \footnotesize LPIPS$\downarrow$ }
    & \multicolumn{1}{c}{ \footnotesize Acc$\uparrow$ }
    & \multicolumn{1}{c}{ \footnotesize LPIPS$\downarrow$ }
    & \multicolumn{1}{c}{ \footnotesize Acc$\uparrow$ }
    & \multicolumn{1}{c}{ \footnotesize LPIPS$\downarrow$ }
    & \multicolumn{1}{c}{ \footnotesize Acc$\uparrow$ }
    & \multicolumn{1}{c}{ \footnotesize LPIPS$\downarrow$ }
    & \multicolumn{1}{c|}{ \footnotesize Acc$\uparrow$ }
    & \multicolumn{1}{c}{ \footnotesize LPIPS$\downarrow$ }
    & \multicolumn{1}{c}{ \footnotesize Acc$\uparrow$ }\\
    
    \specialrule{0.1pt}{2pt}{2pt}
    (1) \textbf{Full}
    & \makecell{\textcolor{MyGreen}{\checkmark}}
    & \makecell{\textcolor{DarkGreen}{NBS}}
    & \makecell{\textcolor{MyGreen}{\checkmark}}
    & \makecell{\textcolor{MyGreen}{\checkmark}}
    & \makecell{\textcolor{MyGreen}{\checkmark}}
    &$\textbf{.271}$
    &$\textbf{.841}$
    &$\textbf{.382}$
    &$.889$
    &$\textbf{.237}$
    &$\textbf{.967}$
    &$.213$
    &$.909$
    &$\textbf{.256}$
    &$.827$
    &$\textbf{.233}$
    &$\textbf{.914}$
    &$\textbf{.268}$
    &$\textbf{.898}$
    \\

    \specialrule{0pt}{2pt}{2pt} 
    (2) w/o cam. opt.
    & \xmark
    & \makecell{\textcolor{DarkGreen}{NBS}}
    & \makecell{\textcolor{MyGreen}{\checkmark}}
    & \makecell{\textcolor{MyGreen}{\checkmark}}
    & \makecell{\textcolor{MyGreen}{\checkmark}}
    &$.315$
    &$.801$
    &$.407$
    &$\textbf{.898}$
    &$.270$
    &$.959$
    &$\textbf{.202}$
    &$\textbf{.920}$
    &$.268$
    &$\textbf{.833}$
    &$.283$
    &$.851$
    &$.294$
    &$.885$
    \\

    \specialrule{0pt}{2pt}{2pt} 
    (3) w/ SE(3)-field
    & \makecell{\textcolor{MyGreen}{\checkmark}}
    & \makecell{\textcolor{DarkRed}{SE(3)-field}}
    & \makecell{\textcolor{MyGreen}{\checkmark}}
    & \makecell{\textcolor{MyGreen}{\checkmark}}
    & \makecell{\textcolor{MyGreen}{\checkmark}}
    &$.274$
    &$.833$
    &$.443$
    &$.786$
    &$.257$
    &$.930$
    &$.217$
    &$.893$
    &$.395$
    &$.619$
    &$.245$
    &$.898$
    &$.302$
    &$.841$
    \\

    \specialrule{0pt}{2pt}{2pt} 
    (4) w/o deform. field
    & \makecell{\textcolor{MyGreen}{\checkmark}}
    & None
    & \makecell{\xmark}
    & \makecell{\textcolor{MyGreen}{\checkmark}}
    & \makecell{\textcolor{MyGreen}{\checkmark}}
    &$.297$
    &$.833$
    &$.408$
    &$.872$
    &$.250$
    &$.940$
    &$.243$
    &$.862$
    &$.298$
    &$.798$
    &$.285$
    &$.833$
    &$.296$
    &$.867$
    \\

    \specialrule{0pt}{2pt}{2pt} 
    (5) w/o root-body init.
    & \makecell{\textcolor{MyGreen}{\checkmark}}
    & \makecell{\textcolor{DarkGreen}{NBS}}
    & \makecell{\textcolor{MyGreen}{\checkmark}}
    & \makecell{\xmark}
    & \makecell{\textcolor{MyGreen}{\checkmark}}
    &$.311$
    &$.821$
    &$.410$
    &$.848$
    &$.251$
    &$.951$
    &$.214$
    &$.892$
    &$.322$
    &$.747$
    &$.250$
    &$.899$
    &$.293$
    &$.870$
    \\

    \specialrule{0pt}{2pt}{2pt} 
    (6) w/o root-body
    & \makecell{\xmark $^\dagger$}
    & \makecell{\textcolor{DarkGreen}{NBS}}
    & \makecell{\textcolor{MyGreen}{\checkmark}}
    & \makecell{\xmark}
    & \makecell{\xmark}
    &\small{N/A}
    &\small{N/A}
    &\small{N/A}
    &\small{N/A}
    &\small{N/A}
    &\small{N/A}
    &\small{N/A}
    &\small{N/A}
    &\small{N/A}
    &\small{N/A}
    &\small{N/A}
    &\small{N/A}
    &\small{N/A}
    &\small{N/A}
    \\

    \specialrule{0pt}{2pt}{2pt} 
    (7) w/o root-body (SE3)
    & \makecell{\xmark}
    & \makecell{\textcolor{DarkRed}{SE(3)-field}}
    & \makecell{\textcolor{MyGreen}{\checkmark}}
    & \makecell{\xmark}
    & \makecell{\xmark}
    &$.373$
    &$.703$
    &$.437$
    &$.749$
    &$.311$
    &$.892$
    &$.376$
    &$.640$
    &$.326$
    &$.758$
    &$.328$
    &$.805$
    &$.360$
    &$.766$
    \\
    
    \bottomrule
\end{tabular}
}

%% file: tables/ablation_jointfinetuning_entireimg_with_acc.tex
\small{
\begin{tabular}{l||cc|cc|cc|cc|cc|cc|cc}

\toprule

& \multicolumn{ 2 }{c}{
  \makecell{
  \textsc{\small Dog 1}
\\\small(626 images)
  }
}
& \multicolumn{ 2 }{c}{
  \makecell{
  \textsc{\small Cat 1 }
\\\small(641 images)
  }
}
& \multicolumn{ 2 }{c}{
  \makecell{
  \textsc{\small Cat 2 }
\\\small(834 images)
  }
}
& \multicolumn{ 2 }{c}{
  \makecell{
  \textsc{\small Human 1 }
\\\small(550 images)
  }
}
& \multicolumn{ 2 }{c}{
  \makecell{
  \textsc{\small Human - Dog}
\\\small(392 images)
  }
}
& \multicolumn{ 2 }{c|}{
  \makecell{
  \textsc{\small Human - Cat}
\\\small(431 images)
  }
}
& \multicolumn{ 2 }{c}{
  \makecell{
  \textsc{\small Mean }
  }
}
\\

& \multicolumn{1}{c}{ \footnotesize LPIPS$\downarrow$ }
& \multicolumn{1}{c}{ \footnotesize Acc@0.1m$\uparrow$ }
& \multicolumn{1}{c}{ \footnotesize LPIPS$\downarrow$ }
& \multicolumn{1}{c}{ \footnotesize Acc@0.1m$\uparrow$ }
& \multicolumn{1}{c}{ \footnotesize LPIPS$\downarrow$ }
& \multicolumn{1}{c}{ \footnotesize Acc@0.1m$\uparrow$ }
& \multicolumn{1}{c}{ \footnotesize LPIPS$\downarrow$ }
& \multicolumn{1}{c}{ \footnotesize Acc@0.1m$\uparrow$ }
& \multicolumn{1}{c}{ \footnotesize LPIPS$\downarrow$ }
& \multicolumn{1}{c}{ \footnotesize Acc@0.1m$\uparrow$ }
& \multicolumn{1}{c}{ \footnotesize LPIPS$\downarrow$ }
& \multicolumn{1}{c|}{ \footnotesize Acc@0.1m$\uparrow$ }
& \multicolumn{1}{c}{ \footnotesize LPIPS$\downarrow$ }
& \multicolumn{1}{c}{ \footnotesize Acc@0.1m$\uparrow$ }
\\
\hline

  \small w/o joint-ft
  &$.273$
  &$\textbf{.859}$
  
  &$\textbf{.379}$
  &$\textbf{.889}$
  
  &$.239$
  &$.963$

  &$\textbf{.207}$
  &$\textbf{.916}$
  
  &$.259$
  &$\textbf{.840}$
  
  &$.241$
  &$.901$
  
  &$.268$
  &$\textbf{.902}$
  
  \\ \small \textbf{Full} (w/ joint-ft)
  &$\textbf{.271}$
  &$.841$
  
  &$.382$
  &$.889$

  &$\textbf{.237}$
  &$\textbf{.967}$  
  
  &$.213$
  &$.909$
  
  &$\textbf{.256}$
  &$.827$
  
  &$\textbf{.233}$
  &$\textbf{.914}$
  
  &$\textbf{.268}$
  &$.898$
  \\ \bottomrule

\end{tabular}
}